\documentclass[onefignum,onetabnum]{siamonline220329}
\newcommand{\PP}{\mathcal{P}}
\newcommand{\RR}{\mathbb{R}}

\renewcommand{\hat}{\widehat}

\DeclareMathAlphabet\mathbfcal{OMS}{cmsy}{b}{n}
\newcommand{\R}{\mathfrak{R}}
\newcommand{\N}{\mathfrak{N}}
\newcommand{\Hf}{\mathfrak{H}}

\newcommand{\A}{\mathcal{A}}

\newcommand{\C}{\mathfrak{C}}

\newcommand{\cc}{\mathfrak{c}}
\newcommand{\m}{\mathfrak{m}}

\newcommand{\U}{\mathcal{U}}
\newcommand{\V}{\mathcal{V}}
\newcommand{\Z}{\mathcal{Z}}

\newcommand{\Y}{\mathcal{Y}}

\newcommand{\mF}{\mathcal{F}}

\newcommand{\HH}{\mathbfcal{H}}
\newcommand{\FF}{\mathbfcal{T}}
\newcommand{\Expect}{\mathbb{E}}
\newcommand{\TT}{\mathbfcal{T}}

\newcommand{\T}{\mathsf{T}} 
\newcommand{\F}{\mathsf{F}}
\newcommand{\G}{\mathcal{G}}

\newcommand{\Hsf}{\mathsf{H}}
\renewcommand{\S}{\mathsf{S}}
\newcommand{\Th}{\widehat{\T}}

\newcommand{\E}{\mathcal{E}}

\newcommand{\CD}{\text{EMCMD$^2$}}
\newcommand{\MMD}{\text{MMD}}
\newcommand{\JMMD}{\text{JMMD}}

\newcommand{\dd}{{\rm d}}
\newcommand{\DD}{D_{\rm AC}}

\newcommand{\DDR}{D_{\rm RAC}}
\newcommand{\DG}{D_{G}}

\newcommand{\DGsil}{D_{G_{\sigma}^{\text{Leb}}}}
\newcommand{\DGsi}{D_{G_{\sigma}}}

\newcommand{\Gsi}{G_\sigma}
\newcommand{\Gsir}{G_\sigma^{\text{ren}}}
\newcommand{\Gsil}{G_\sigma^{\text{Leb}}}

\newcommand{\pipe}{\:\Vert\:}

\newcommand{\ThR}{\Th_{\rm{R}}}

\renewcommand{\SS}{\mathcal{S}}
\newcommand{\KK}{\mathcal{K}}
\newcommand{\M}{M}

\usepackage{lipsum}
\usepackage{amsfonts}
\usepackage{graphicx}
\usepackage{caption,subcaption}
\usepackage[textsize=tiny]{todonotes}

\usepackage[percent]{overpic}

\usepackage{tikz}

\usepackage{epstopdf}
\usepackage{algorithm}
\usepackage{algpseudocode}
\usepackage{pict2e}
\usepackage{amsmath}
\usepackage{amssymb}
\usepackage{mathtools}
\usepackage{centernot}
\ifpdf
\DeclareMathOperator*{\argmin}{argmin}
\DeclareGraphicsExtensions{.eps,.pdf,.png,.jpg}
\else
  \DeclareGraphicsExtensions{.eps}
\fi

\renewcommand{\t}{\triangle}
\usepackage{enumitem}
\setlist[enumerate]{leftmargin=.5in}
\setlist[itemize]{leftmargin=.5in}

\usepackage{amssymb}
\usepackage{dsfont}
\DeclareMathOperator*{\essinf}{ess\,inf}

\newsiamremark{remark}{Remark}
\newsiamremark{hypothesis}{Hypothesis}
\newsiamremark{assumption}{Assumption}
\newsiamthm{problem}{Problem}
\crefname{hypothesis}{Hypothesis}{Hypotheses}
\newsiamthm{claim}{Claim}

\headers{On the Error Analysis of Triangular Transport Maps for Conditional Generation via MMD-based Objectives}{M. Martino, B. Hosseini, A. Sagiv and A. Taghvaei}

\title{
On the Error Analysis of Triangular Transport Maps for Conditional Generation via MMD-based Objectives
\thanks{
\funding{MM and BH were supported by NSF grant DMS-2337678, AT was funded by AFOSR grant FA9550261B219. AS was funded by NSF grant DMS-2508811.}}}

\author{Michele Martino
\thanks{Department of Applied Mathematics, University of Washington, Seattle, WA, USA 
  (\email{michem29@uw.edu},\email{bamdadh.uw.edu} ).}
\and Bamdad Hosseini \footnotemark[2]
\and Amir Sagiv\thanks{Department of Mathematical Sciences, New Jersey Institute of Technology, Newark, NJ, USA 
  (\email{amir.sagiv@njit.edu}).}
\and Amirhossein Taghvaei\thanks{Department of Aeronautics and Astronautics, University of Washington, Seattle, WA, USA 
  (\email{amirtag.uw.edu}).}}

\usepackage{amsopn}

\DeclareMathOperator*{\esssup}{ess\,sup}

\usepackage{etoolbox}
\newtheorem{example}{Example}

\ifpdf
\hypersetup{
  pdftitle={
  On the Error Analysis of Triangular Transport Maps for Conditional Generation via MMD-based objectives 
  },
  pdfauthor={M. Martino, B. Hosseini, A. Sagiv and A. Taghvaei}
}
\fi

\begin{document}
\maketitle
\begin{abstract}
Conditional generative models parameterized by triangular transport maps are widely used for conditional sampling, yet theoretical guarantees on the quality of the conditional measures they produce remain scarce. We address this question for minimum divergence estimators of triangular maps. We use MMD-based objectives as a tractable setting in which the gap between joint and conditional errors can be analyzed precisely, and we measure performance through the average conditional estimation error, defined as the squared maximum mean discrepancy (MMD) between the estimated and the true conditionals, averaged over the conditioning variable. We first show that the common practice of minimizing a joint MMD may fail to recover the conditionals. Writing the joint and the conditional criteria as quadratic energies of the same conditional mean embedding residual, we prove that, for parametric classes, joint minimizers are generically inconsistent for the average conditional estimation error in the misspecified setting. We then analyze an estimator that directly minimizes the conditional criterion through a surrogate loss computable from joint target samples and reference samples, obtained by parameterizing conditional mean embedding regression with triangular maps. We prove a non-asymptotic error bound that decomposes the average conditional estimation error into approximation, estimation, and reference sampling errors, with fast rates that are independent of the ambient dimension in the parametric setting. Finally, for joint MMDs built from concentrating Gaussian kernels on the conditioning variable, we show that the joint criterion approximates a density-reweighted version of the conditional one, and we derive an error bound for the joint minimizer whose rate suffers from a curse of dimensionality in the conditioning variable. Beyond MMD minimization for static measure transport, our results suggest that training criteria and design choices for conditional generative models should target the conditional rather than the joint error, a principle that extends to other divergences as well as the dynamic transport counterpart.
\end{abstract}



\section{Introduction}
Conditional transport models \cite{marzouk2017sampling, baptista2024conditional} and generative models \cite{mirza2014conditional} are widely used in machine learning (ML) and offer 
a paradigm shift in scientific ML applications such as inverse problems \cite{orozco2025aspire}, 
stochastic simulation \cite{yang2020physics}, and surrogate modeling \cite{zhang2025probabilistic}.
The wide adoption and development of such conditional models demands the 
development of rigorous theoretical foundations for  generative models. The goal of this 
article is precisely to address this question for the particular case of minimum 
divergence conditional generative models parameterized as transport maps. In particular, 
we provide an answer to the following question: 
{\it How good are the conditional measures produced by a conditional generative model?}

To be more precise, consider a parameter space $\U$ and a data/observation space $\Y$ 
along with a target probability measure $\nu \in \PP(\Y \times \U)$
and an empirical data set of independent and identically distributed (i.i.d.) 
samples $\{(y_i,u_i) \}_{i=1}^m \overset{\textnormal{i.i.d.}}{\sim} \nu$; henceforth represented as 
$\nu^m$. In standard generative modeling, our goal is to obtain an 
approximation $\hat{\nu} \approx \nu$ while in conditional generative modeling
we further ask for a "good" approximation of the conditional measures $\nu(\cdot \mid y)$, i.e., 
we ask for $\hat{\nu}(\cdot \mid y) \approx \nu(\cdot \mid y)$. This is indeed a non-trivial task 
given that the empirical data consists of multiple samples from the joint measure $\nu$ 
and not multiple samples from individual conditionals: for absolutely continuous distributions, at best we have a single sample 
$u_i \sim \nu( \cdot \mid y_i)$ but we want to approximate the conditionals for 
unseen choices of $y$. 
Many existing approaches for this task aim to learn the conditionals by training a 
constrained generative model that matches the joint empirical measure $\nu^m$ 
\cite{baptista2024conditional}. However, such models can be inconsistent 
as shown in the following example where two  measures $\nu$ and $\hat{\nu}$ 
are close with respect to a divergence while their conditionals remain far apart. 

\begin{example}\label{ex:counter-CMMD-counter-example}
    Consider $Y \sim N(0,1)$ and set $U = \alpha Y$, for some $\alpha > 0$, and define $\nu = \text{Law}\; (Y,U)$. 
    Similarly define  $\hat{\nu} = \text{Law}\; (Y,-U)$. Consider the RBF kernel 
    $K(u, u') = \exp( - \frac{1}{2} \| u - u' \|^2)$ and let $D_{\text{Cond.}}: \PP(\U) \times \PP(\U) \to \RR_{\ge 0}$ 
    be the squared maximum mean discrepancy (MMD) defined via this kernel (see~\eqref{eq: MMD def} below). 
    Then a direct calculation, outlined in \Cref{app:CMMD-counter-example}, shows that for a fixed $y\neq 0$ we have
    \begin{equation}\label{eq: first part example}
D_{\text{Cond.}}\left(\nu(\cdot \mid y\right), \hat{\nu}(\cdot \mid y) ) = 
        2 \left( 1 - \exp \left( - 2\alpha^2 y^2  \right) \right) \xrightarrow[\text{as }\alpha \to \infty]{} 2
    \end{equation}
    At the same time, taking $D_{\text{Joint}}: \PP(\Y \times \U) \times \PP(\Y \times \U) \to \RR_{\ge 0}$
    to be the squared MMD defined by the tensor product kernel $K( (y,u) ,(y',u' ) ) 
    =K(y,y') K(u, u')$, we can show that
    \begin{equation}\label{eq: second part example}
        D_{\text{Joint}}(\nu, \hat{\nu}) = 
        \frac{2}{\sqrt{3 + 2\alpha^2}} -  \frac{2}{\sqrt{3(1 + 2\alpha^2)}} 
        \xrightarrow[\text{as }\alpha \to \infty]{} 0.
    \end{equation}
    In other words, for large $\alpha$ the conditionals remain different even though the joint measures 
    become closer. We show a schematic depiction of this problem in \Cref{fig:CMMDvsMMD}. 

    \begin{figure}[h]
    \centering
    \begin{subfigure}{0.3\textwidth}
        \includegraphics[width=\linewidth]{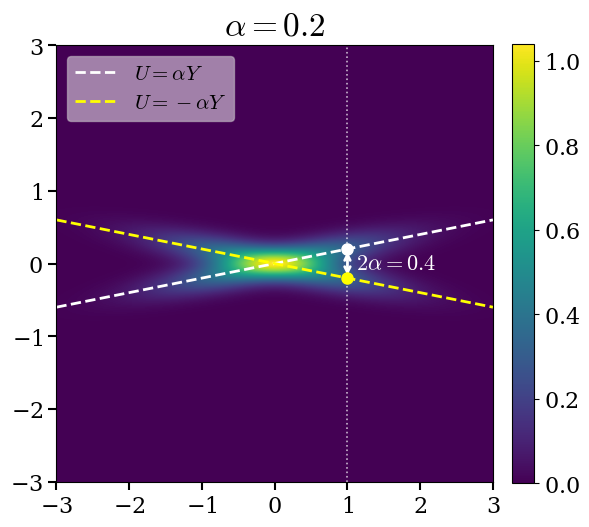}
    \end{subfigure}
    \hfill
    \begin{subfigure}{0.3\textwidth}
        \includegraphics[width=\linewidth]{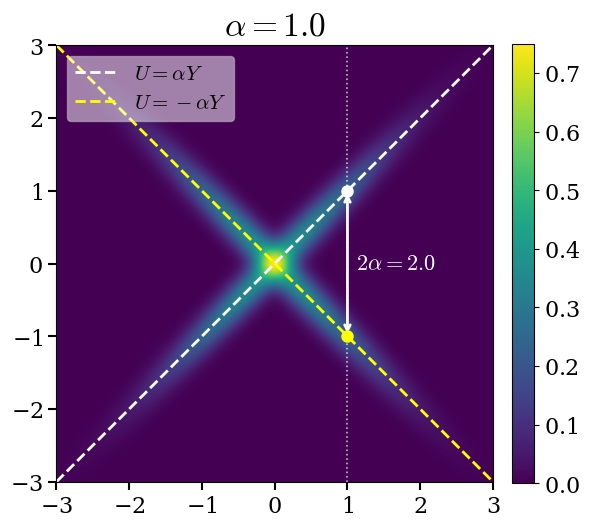}
    \end{subfigure}
    \hfill
    \begin{subfigure}{0.3\textwidth}
        \includegraphics[width=\linewidth]{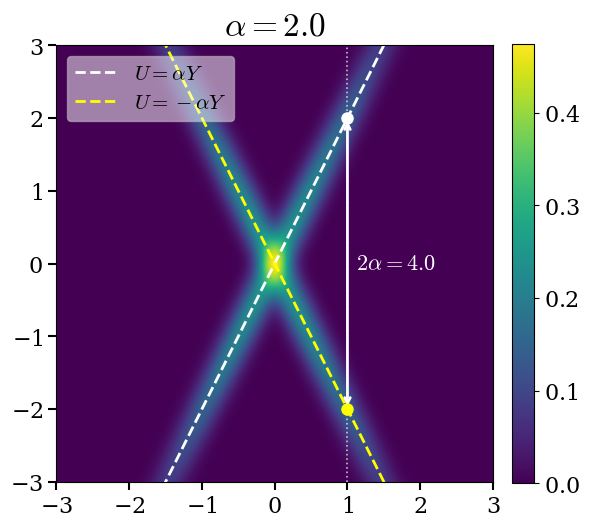}
    \end{subfigure}
    \caption{Joint density of $(Y, U)$ for two distributions supported on the lines $U = \alpha Y$ (white) and $U = -\alpha Y$ (yellow), with $Y \sim \mathcal{N}(0,1)$. The distributions are visualized as Gaussians with small perpendicular bandwidth $\varepsilon = 0.15$. The vertical dotted line marks $Y = 1$; the double-headed arrow shows the separation $2\alpha$ between the two modes at that slice.}
    \label{fig:CMMDvsMMD}
\end{figure}
    
\end{example}

Our primary goal in this manuscript is to understand and study this inconsistency and 
find a remedy. While we focus on static triangular maps \cite{marzouk2017sampling,hosseini2023conditional}, we note that the above 
inconsistency is not specific to this class and is pertinent for other conditional generative models  such as dynamic transport (flow matching \cite{lipman2022, albergo2025stochastic} and diffusion models~\cite{yang2023diffusion}) with triangular velocity fields, a perspective we discuss further in~\Cref{rem: joint vs conditional beyond static} and~\Cref{rem: dynamic transport}.

In light of the above example we want to know 
{\it under what conditions and for what choices of 
$D_{\text{Joint}}$ can we show that a generative model trained 
by minimizing $D_{\text{Joint}}$ has good conditional performance with respect to $D_{\text{Cond.}}$.}
To obtain a positive answer we need to work with appropriate divergences 
based on kernel mean embeddings
as well as model classes 
for our generative models that allow us to exploit the "smooth variations" 
of the conditional measures 
$\nu(\cdot \mid y)$ as a function of $y$. At an intuitive level 
this "smoothness" will allow us to share very limited information among 
individual conditionals to infer the entire family.
Henceforth, we will focus our attention on minimum divergence triangular transport models. 

\subsection{Minimum divergence estimation of triangular maps}
\label{subsec:min-div-triangular-maps}
Take $\V,\U,\Y$ and $\Z$ to be Hilbert spaces. Following \cite{baptista2024conditional,hosseini2023conditional} we define the class of triangular maps as follows:

\begin{definition}[Triangular maps]\label{def: block-triangular map}
We say that a map $\T_\t^\F:\Z\times\V\mapsto\Y\times\U$, is (block-) triangular\footnote{The name triangular 
derives from the fact that the gradient of such a map is block-triangular. These maps also serve 
as natural extensions of Knothe-Rosenblatt maps \cite{marzouk2017sampling} to abstract settings.} if it has the form 
$\T_\t^\F( z, v) = \left( \F(z), \T(\F(z), v) \right),$
with components $\F: \Z \to \Y$ and $\T: \Y \times \V \to \U$. 
\end{definition}

Triangular maps have a natural connection to conditional measures by \cite[Thm.~2.4]{baptista2024conditional} and its more general extension which we present in~\Cref{prop:triangular-transport-conditioning}. 
Consider two measures $\eta\in\PP(\Z\times\V)$ and $\nu \in \PP(\Y \times \U)$, where we think of $\eta$ as a {\it reference measure} 
and $\nu$ as {\it the target measure}. 
If $\T^\F_\t \# \eta = \nu$ \footnote{We write $\T \# \eta := \text{Law} \{ T(z) \mid z \sim \eta \}$ 
to denote the pushforward of $\eta$ under $\T$.} then, the conditionals $\nu(\cdot\mid y)$ admit the mixture representation $\int \T( y, \cdot) \# \eta(\cdot \mid z)\widetilde{\eta}( \dd z\mid \F
(z)=y)$ in terms of 
the weight measure $\widetilde{\eta}( \dd z\mid \F
(z)=y)$ supported on the pre-image set $\F^{-1}(\{y\})$ induced by the disintegration $\eta_
\Z(\dd z) = \int \widetilde{\eta}(\dd z\mid \F(z) = y)(\F\#\eta_\Z)(\dd y) $. 
In most conditional generative modeling tasks, this mixture representation of the target conditionals can be reduced to a more implementable push-forward representation without the need for a non-trivial $\F$ mapping by choosing an appropriate reference $\eta$. Indeed, setting 
$\eta = \nu_\Y \otimes \eta_\V$, i.e., the independent coupling between $\nu_\Y$ and a $\V$-reference measure 
$\eta_\V \in \PP(\V)$, we can simply take $\F = \text{Id}$ and obtain the 
simpler model $\T_\t^{\textnormal{Id}}(y, v) 
= (y, \T(y, v))$ alongside the conditioning identity 
$\T(y, \cdot) \# \eta_\V = \nu(\cdot \mid y)$. To this end, triangular maps are a
natural model class for conditional generative models by parameterizing the component $\T$.

\subsection{Overview of main results}
While various strategies for computing triangular transport exist 
\cite{hosseini2023conditional, wang2025efficient, baptista2025entropicconditional, baptista2024conditional, cui2023scalable, marzouk2017sampling} here we consider the class of minimum divergence estimators. Let 
$\nu^m$ be an empirical measure approximating $\nu$ with $m$ i.i.d. samples 
and similarly $\eta_\V^k$ be an empirical approximation to $\eta_\V$ with $k$ 
samples. Note that, due to the reference structure $\eta = \nu_\Y \otimes \eta_\V$, concatenating $m$ samples from the ones yielding $\eta_\V^k$ with 
the $\Y$-marginal part of the samples giving $\nu^m$ (when $m\leq k$), yields also the joint empirical reference $\eta^m$ if needed. Then we 
consider estimators of the form
\begin{equation}\label{eq:empirical-T-estimator}
    \hat{\T}^{m, k} \in \argmin_{\T \in \FF} 
    D_{\text{Joint}}^{m,k}\left(\nu, \T_\t \# \eta \right),
\end{equation}
where $\FF$ is a model hypothesis class, $D_{\text{Joint}}$
can be any divergence/loss defined on $\PP(\Y \times \U)$, and $ D_{\text{Joint}}^{m,k}$ is a corresponding empirical approximation or surrogate using $\nu^m$ and $\eta_\V^k$ in place of $\nu$ and $\eta_\V$.

\paragraph{Theoretical guarantees for conditional generation} In this work, we evaluate the estimators' capability of recovering the target conditionals $\nu(\cdot\mid y)$ via $\widehat{\nu}(\cdot\mid y) =\Th^{m,k}(y,\cdot)\#\eta_\V$ by the notion of \emph{average conditional estimation error}, namely, 
\begin{equation}\label{eq: quantity of interest for conditional generation}
\begin{aligned}
    \DD(\nu\pipe\widehat\nu)\coloneqq\int_\Y   D_{\text{Cond.}}\left(\nu(\cdot \mid y) , \widehat{\nu}(\cdot \mid y)\right) \nu_\Y(\dd y), 
\end{aligned}
\end{equation}
where we later set $D_{\text{Cond.}}$ to be the squared $\MMD_K$ defined in terms of a prescribed kernel $K: \U \times \U \to \RR$. 

As shown in~\Cref{prop: fixed sampled conditional control}, controlling~\eqref{eq: quantity of interest for conditional generation} translates into a high probability control of $D_{\text{Cond.}}(\nu(\cdot\mid y), \widehat \nu(\cdot\mid y))$ for an independent new sampled conditional value $y\sim \nu_\Y$. Nevertheless, we stress that controlling $\DD(\nu\pipe\widehat\nu)$ does not yield theoretical guarantees as strong as controlling directly $\sup_{y\in\Y}D_{\text{Cond.}}\left(\nu(\cdot \mid y) , \widehat{\nu}(\cdot \mid y)\right)$. However, the training data consist of joint samples, typically with a single observation for each value of $y$, so the conditional discrepancy at a fixed $y$ cannot be estimated without additional localization in $y$. Moreover, as pointed out in~\Cref{rem: limitations in controlling the averaged error}, the control of the average estimation error suffices to derive adequate theoretical guarantees on the pointwise conditional estimation error for sufficiently expressive classes. This motivates 
our focus on $\DD(\nu\pipe\widehat\nu)$.

 \paragraph{Inconsistency of estimators trained only at the joint level} 
Let us take $D_{\text{Joint}}$ to be the 
\emph{Joint Squared Maximum Mean discrepancy} 
\begin{align}\label{eq: Joint MMD squared}
\JMMD_G^2(\rho,\rho') \coloneqq \left\| I_{K\otimes G}(\rho) - I_{K\otimes G}(\rho')\right\|^2,\qquad \forall \rho,\rho' \in \PP(\Y \times \U)
\end{align}
where $K\otimes G:(\Y\times\U)\times (\Y\times\U)\mapsto \mathbb{R}$ is a tensor product kernel 
defined on 
$\Y \times \U$ 
\begin{align}\label{eq: joint product kernel}
K\otimes G((y,u),(y',u'))\coloneqq G(y,y') K(u,u')\, ,\qquad \forall(y,u),(y',u')\in\Y\times \U\, ,
\end{align}
based on two symmetric 
positive semidefinite (PSD) kernels $K$ and $G$ 
and $I_{K \otimes G}(\rho) := 
\int K(u, \cdot) G(y, \cdot) \rho(\dd y, \dd u) $ 
denotes the kernel mean 
embedding of $\rho$.
In this setup, the joint empirical estimator $D_{\text{Joint}}^{m,k}$ requires data $\nu^m$ and the construction of the empirical joint reference measure $\eta^m$ using $k = m $ samples from $\eta_\V.$ We then condense notation using $\widehat \T_G^{m}$ to denote $\widehat \T ^{m,k= m}$ from~\eqref{eq:empirical-T-estimator}. We introduce the additional subscript in order to denote its dependence on the choice of $G$. 

As mentioned earlier, we argue that the  approach is not ideal for the purpose of conditional generation. Indeed, denoting $\E_{\text{Bias}}^2\coloneqq \inf_{\T\in\TT}\DD(\nu\pipe \T_\t\#\eta)$, we show in~\Cref{thm: inconsistency} that, in the parametric setting, under generic conditions and for any practical choice of $G$,
\begin{equation*}
    \liminf_{m\to\infty}\DD(\nu\pipe \Th^m_{G\t}\#\eta)\geq \E_{\text{Bias}}^2+\mathfrak{d}\qquad\text{a.s.},
\end{equation*}
for an irreducible error $\mathfrak{d}>0$, even when $\Th^m_G$ is consistent with respect to the $\JMMD_G^2$. Since $\mathfrak{d}$ only depends on population minimizers, this holds for both biased and unbiased empirical approximations of $\JMMD_G^2$.

\paragraph{Error analysis of triangular maps for conditional generation} 
The previous paragraph motivates the use of an alternative optimization problem which instead
minimizes the \emph{Expected Squared Maximum Conditional Mean Discrepancy} 
introduced in \cite{park2020measure}:
\begin{equation}\label{def:EMCMD2}
    \CD(\rho \: \Vert \: \rho') := \int_\Y \left\| I_K \rho(\cdot \mid y) - I_K \rho'(\cdot \mid y) \right\|_\KK^2  \; \rho_\Y( \dd y),
    \qquad \forall \rho,\rho' \in \PP(\Y \times \U). 
\end{equation}
Note that this divergence is not symmetric unless $\rho$ and $\rho'$ share the same $\Y$-marginal and 
that it can also be viewed as the averaged squared $\MMD_K$ between the 
conditionals $\rho(\cdot \mid y)$ and $\rho'(\cdot \mid y)$. Essentially, the resulting estimator directly minimizes~\eqref{eq: quantity of interest for conditional generation} since $\DD(\rho\pipe\rho') = \CD(\rho\pipe\rho')$ choosing $D_{\text{Cond.}} = \MMD_K^2$. As we shall see in this alternative scenario, an empirical surrogate $D_{\text{Joint}}^{m,k}$ can be constructed using data $\nu^m$ and accessible marginal reference samples $\eta_\V^{k}$ for $k = mn$ and $n\geq 2$, without the need to build the empirical joint reference $\eta^m$. Slightly abusing notation, in this case we use $\widehat \T^{m,n}$ to denote $\widehat \T ^{m,k=mn}$ from~\eqref{eq:empirical-T-estimator}. 

We remark that this approach has already been formulated in~\cite{huang2022evaluating} but without a complete error analysis from the perspective of triangular measure transport for conditioning. It also includes engression~\cite{shen2025engression} as the special case of the energy-distance kernel. Recent error analyses of engression-type estimators were developed independently of the present work in~\cite{huang2026engression,chen2026engression,chai2026neural}; we compare these works with ours in~\Cref{subsec: related literature}.
In~\Cref{thm: main error bound 2}, we provide a complete error analysis of the more amenable estimator $\Th^{m,n}$ based on~\eqref{def:EMCMD2}. Particularly, we prove a non-asymptotic error bound on the average conditional estimation error~\eqref{eq: quantity of interest for conditional generation} of the estimator $\Th^{m,n}$ in terms of both sample sizes $m$ and $n$, under the control of the complexity of the Hilbert-valued function class induced by $\TT$. 
The bound decomposes the error into the bias term $\E_{\text{Bias}}^2$ plus the sum of a statistical generalization error and a Monte Carlo reference-sampling error. In the parametric setting, this sum reduces to $\mathcal{O}\left(\frac{\log(m)}{m}+ \frac{1}{\sqrt{mn}}\right)$ and is ambient-dimension-free.

Furthermore, in~\Cref{Error bound of joint minimizer} we also provide an upper bound on the average conditional estimation error of the $\JMMD_G^2$-based estimator $\Th^m_G$ when $G$ is taken to be an RBF kernel with length-scale $\sigma>0$ which we select optimally. In contrast to  $\Th^{m,n}$, the upper bound on $\Th_{\Gsi}^m$ suffers a curse of dimensionality which we conjecture is inevitable and caused by a  smoothing mechanism 
in the $y$ coordinate in the definition of $\JMMD_G^2$; see 
~\Cref{rem: high frequency blindness}.

\paragraph{Connection between $\CD$ and $\JMMD_G^2$} As a further contribution, we characterize the connection between $\CD(\rho\pipe \rho')$ and $\JMMD_G^2(\rho,\rho')$ towards
resolving the above inconsistencies. In~\Cref{prop: energy rep}, we  study the energy representations of both divergences in terms of the conditional mean embedding residuals between $\rho$ and $\rho'$ and the kernel integral operator associated to $G$. 
This is effectively the mechanism that allows us to 
prove the inconsistency of $\JMMD_G^2$-based estimators.
Moreover, we use Fourier analysis to show that  $\JMMD_G^2$ 
approaches a reweighted version of $\CD$ from below in the particular case where $G$ is selected from a family of renormalized  RBF kernels of length-scale $\sigma\downarrow 0$.
In other words, a $\nu_\Y$-density-reweighted $\CD$ is obtained as a limit of $\JMMD_G^2$.
Our proof technique reveals that 
this connection can be extended to any appropriately rescaled translation invariant kernel $G$.

Finally, this analysis uncovers
the mechanism behind~\Cref{ex:counter-CMMD-counter-example}, 
i.e., if the conditionals of $\rho, \rho'$ vary rapidly
then their difference is nearly indistinguishable to 
$\JMMD_G^2$ due to the presence of a low-pass filter 
while they remain detectable with respect to $\CD$.

Finally, our results support a general principle for conditional generative modeling: training criteria and design choices should be tailored to the conditional error rather than joint divergences. In other settings such as dynamic measure transport via flow matching with triangular velocity fields, this principle suggests choosing the conditioning paths according to their conditional, rather than joint, regularity, as we discuss in~\Cref{rem: dynamic transport}.

\subsection{Outline}
\begin{itemize}
\item In~\Cref{sec:preliminaries}, we provide a self-contained review on Reproducing Kernel Hilbert Spaces (RKHS), Maximum Mean Discrepancies (MMD), and their conditional analogues as well as a quick introduction to triangular measure transport for conditional generation.

\item In~\Cref{sec: setup}, we formally introduce the notion of average conditional divergence of~\eqref{eq: quantity of interest for conditional generation} and its theoretical guarantees for assessing conditional generation capabilities of triangular transport maps. Moreover, we present in details both the $\JMMD_G^2$ and the $\CD$ minimization setups for training triangular maps using finite samples from the target distribution $\nu$ and the reference source $\eta_\V$. 

\item In~\Cref{sec:error-analysis}, we present the complete error analysis of the $\CD$-based estimator $\Th^{m,n}$ with respect to~\eqref{eq: quantity of interest for conditional generation}.

\item In~\Cref{sec: DD vs DG minimization}, we focus on the relation between the criteria $\JMMD_G^2$ and $\CD$ by analysing their energy and Fourier representations. As a byproduct of these, we derive the aforementioned inconsistency statement for 
$\JMMD^2_G$-based estimators and the error analysis of $\Th_{\Gsi}^m$ with respect to~\eqref{eq: quantity of interest for conditional generation}.

\end{itemize}

\subsection{Related literature}\label{subsec: related literature}

Generative modeling can be cast as learning a transport map that pushes a simple reference measure forward to a complex target measure. This viewpoint encompasses generative adversarial networks (GANs)~\cite{goodfellow2014} and normalizing flows~\cite{rezende2015,papamakarios2021,kobyzev2021}, which can both be understood as minimum divergence estimators~\cite{pardo2018}. It also encompasses diffusion and flow-based models~\cite{ho2020,lipman2022}, which transport the reference through learned dynamics. Conditional variants of these models, such as conditional GANs~\cite{mirza2014conditional}, conditional normalizing flows~\cite{trippe2018,winkler2019} and conditional diffusion models~\cite{batzolis2021,saharia2022}, target the conditionals $\nu(\cdot\mid y)$ through architecture-specific mechanisms. In general, these mechanisms do not come with structural guarantees that the learned model recovers the conditionals.

Triangular maps provide a principled alternative, since their structure ensures that a map pushing the reference to the joint target also recovers the conditionals. Triangular rearrangements trace back to Knothe~\cite{knothe1957} and Rosenblatt~\cite{rosenblatt1952}, and were introduced in the computational setting for sampling and Bayesian inference in~\cite{elmoselhy2012,marzouk2017sampling}. Their connection to optimal transport (OT) was established in~\cite{carlier2010}, which recovers the Knothe--Rosenblatt (KR) rearrangement as a limit of Brenier maps, while OT characterizations of conditional quantiles were developed in~\cite{carlier2016}. The conditioning property of block-triangular maps was formalized in~\cite[Thm.~2.4]{baptista2024conditional}, where it underpins the monotone GAN approach to conditional sampling. It was further developed into a theory of conditional OT on function spaces in~\cite{hosseini2023conditional}, and efficient neural approaches were proposed in~\cite{wang2025efficient}. Minimum divergence estimators of triangular maps, including joint MMD-based ones, were studied from an approximation-theoretic perspective in~\cite{baptista2023approximation,pandey2024diffeomorphic}.

A parallel line of work provides rigorous foundations for triangular transport. Existence of triangular maps in general spaces was studied in~\cite{bogachev2005,bogachev2006}. The Knothe--Rosenblatt rearrangement was shown to admit sparse polynomial and deep ReLU approximations in~\cite{zech2022a,zech2022b}. Statistical consistency and rates for triangular flows were analyzed in~\cite{irons2022}, minimax rates for transport-based density estimation in~\cite{wang2022minimax}, and the tail behavior of Lipschitz triangular maps in~\cite{jaini2020}. Regularity of conditional OT maps in the conditioning variable was investigated in~\cite{hosseini2023conditional,gonzalezsanz2025}, and deep conditional samplers trained with $f$-divergence objectives were studied in~\cite{zhou2023}. However, most of these works control the approximation of the joint measure. For integral probability metrics such as MMD and Wasserstein distances, this does not translate into control of the conditionals: two measures can be close in such a joint divergence while their conditionals remain far apart (see~\Cref{ex:counter-CMMD-counter-example}). The same gap affects dynamic transport models. Conditional versions of flow matching and diffusion models rely on triangular velocity fields that keep the conditioning variable fixed along the flow~\cite{kerrigan2024dynamic}. Error bounds for flow matching control the Wasserstein error of the generated samples by the training error, amplified by the time-integrated Lipschitz constant of the velocity field~\cite{benton2023error}, so that the choice of path affects accuracy and not only computational cost. However, these error analyses target the joint error, and their conditional counterpart remains largely unexplored.

This work addresses this gap by quantifying the quality of the estimated conditionals through the average conditional estimation error~\eqref{eq: quantity of interest for conditional generation}. We focus on MMD-based divergence minimization, with the aim of identifying principles that may extend to more general divergences and setups. A related question was recently addressed in~\cite{aljarrah2025error} for triangular OT maps in the context of conditional OT and filtering. There, the conditional estimation error is bounded through an upper bound on the error of the estimated map, which is in turn controlled by stability results for OT maps. That analysis is thus specific to conditional OT maps, which are characterized by a min-max (averaged semidual) problem and enjoy uniqueness properties not shared by generic minimum divergence estimators.

We instead consider an empirical estimator that directly minimizes the quantity of interest, and we provide a complete error analysis for it. Our analysis relies on the squared MMD~\cite{gretton2012} and the theory of kernel mean embeddings~\cite{muandet2017kernel}. Conditional mean embeddings, first formulated in operator-theoretic terms in~\cite{song2009}, were given a measure-theoretic foundation together with a regression interpretation in~\cite{park2020measure}, where the conditional analogue of the squared MMD coincides with $\CD$ in~\eqref{def:EMCMD2}. Our estimator $\Th^{m,n}$ parameterizes this regression problem through triangular maps. The same estimator was considered in~\cite{huang2022evaluating} for evaluating aleatoric uncertainty, without an error analysis of the conditional estimation error. It is also closely related to engression~\cite{shen2025engression}, which fits a conditional generator $g(x,\varepsilon)$, with noise independent of the covariates, by minimizing the expected energy score. Such generators are triangular maps with reference $\nu_\Y\otimes\eta_\V$. Since the energy distance is the squared MMD of the distance-induced kernel $K(u,u')=\|u\|+\|u'\|-\|u-u'\|$~\cite{sejdinovic2013equivalence}, a direct computation shows that, for this kernel and measures with finite first moments, $M(\nu,\T)$ is twice the expected energy-score loss of $\T$, and $M^n(\nu^m,\T)$ in~\eqref{eq: M unbiased estimate with n} is twice the empirical engression loss with $n$ noise draws per observation. Engression is thus $\CD$ minimization for this kernel, and its objective coincides with $\CD$ up to a factor $2$ and an additive constant. This kernel is unbounded on unbounded domains and its feature map is only $1/2$-H\"older continuous, so it is not covered by~\Cref{Assumptions}. Bounds in expectation for engression with deep ReLU networks and multivariate responses were recently derived in~\cite{huang2026engression,chen2026engression}, with excess-risk rates of order $m^{-s/(2s+d_{\mathrm{eff}})}$, up to logarithmic factors, for $s$-smooth generators with effective input dimension $d_{\mathrm{eff}}$; \cite{huang2026engression} also analyzes the multi-step reverse Markov extension of engression~\cite{shen2025reverse}. Closest to our work is~\cite{chai2026neural}, which derives high-probability localized oracle inequalities for stochastic deep ReLU networks with bounded covariates and scalar, bounded responses, first for the energy loss, interpreted as an average Cram\'er distance between conditional CDFs, and then, in an appendix, for $\CD$ minimization with bounded kernels whose feature maps are $1/2$-H\"older continuous. Under a hierarchical compositional structure of the conditional quantile function, they obtain the faster rate $m^{-2s/(2s+d_{\mathrm{eff}})}$, relying, among other conditions, on Lipschitz conditional CDFs. Finally,~\cite{chatterjee2025} controls the average conditional estimation error by combining joint MMD minimization with a nearest-neighbor partitioning of the conditioning variable; the resulting rates are slow, since the analysis does not localize the generalization error and inherits the curse of dimensionality of nearest-neighbor methods, whereas $\Th^{m,n}$ avoids localization in $y$ altogether.

Our results were obtained independently of the analyses of engression above, which appeared while the present manuscript was in preparation, and their results closest to ours (specifically~\cite[Thm.~D.3]{chai2026neural}, an oracle inequality of the same form as~\Cref{thm: main error bound 2} for scalar responses and neural network classes) are obtained through different proof techniques. This is perhaps not surprising, since $\CD$ minimization amounts to a least-squares regression problem for the conditional mean embedding (\Cref{subsec: high-level strategy}; see also~\cite[Lemma~D.1]{chai2026neural}), for which several standard tools are available. We control the generalization error through one-shot localization~\cite{van1987new,van2002m,divol2025optimal-sm} rather than local Rademacher complexities and peeling, and the reference-sampling error, of order $1/\sqrt{mn}$ in both analyses, through a decomposition into a linear term and a degenerate $U$-statistic in the RKHS (\Cref{lem: app reference decomposition}) rather than Hoeffding-type bounds conditional on the data. Beyond these technical differences, \Cref{thm: main error bound 2} holds with explicit constants, for general output spaces $\U$ and general classes of triangular maps characterized through their metric entropy, and requires boundedness of the kernel rather than of the covariates and responses (although~\cite{chai2026neural} note that sub-Gaussian tails can be accommodated at the price of logarithmic factors). Moreover, for Lipschitz kernels, it bounds $\E_{\text{Bias}}^2$ quadratically in the distance between maps without regularity assumptions on the conditionals, whereas for the energy kernel the available bias bounds are linear in this distance~\cite{chen2026engression}, unless the conditional CDFs are Lipschitz as in~\cite{chai2026neural}. Complementary to the downstream guarantees of~\cite{chai2026neural} (e.g., for conditional moments and quantiles), \Cref{prop: fixed sampled conditional control} shows that a high-probability control of the average conditional estimation error, obtained by any estimator, yields guarantees for the conditional at a single new observation $y\sim\nu_\Y$. We also recall that none of the works above contrasts $\CD$ with the joint criterion $\JMMD_G^2$. In~\Cref{sec: DD vs DG minimization}, we relate the two criteria through energy and Fourier representations, show that $\JMMD_G^2$ minimization is generically inconsistent for the conditional estimation error in the misspecified setting (\Cref{thm: inconsistency}), and quantify the price, in terms of rates, of recovering conditional guarantees from joint minimization with concentrating RBF kernels (\Cref{Error bound of joint minimizer}), thereby connecting the joint MMD approaches of~\cite{huang2022evaluating,pandey2024diffeomorphic} with the conditional framework we advocate.
% Concerning the energy kernel itself, its $1/2$-H\"older continuity doubles the exponent in the metric entropy of the lifted class of parameterized conditional mean embeddings, which leaves parametric classes unaffected up to constants, and makes the available bound on $\E_{\text{Bias}}^2$ linear rather than quadratic in the distance between maps; \cite{chai2026neural} handle the first issue for neural network classes and the second through Lipschitz conditional CDFs. On unbounded domains, \Cref{thm: one-shot localization} allows envelopes with Orlicz-type tails, which suggests a route to this kernel without truncation, at least for finite-dimensional $\Y$, $\U$, and $\V$ (see~\Cref{rem:validity-of-population-assumptions}). We leave these extensions to future work.

\section{Preliminaries}\label{sec:preliminaries}

We dedicate this section to a review of preliminary notations, 
definitions, and results from the literature. 
\Cref{subsec: notation-and-defs} summarizes some notation 
and basic definitions. 
\Cref{subseq: MMD RKHS} contains a brief review of 
RKHSs and MMD. Finally, we briefly introduce 
triangular transport maps and recall some of their key properties in 
\Cref{sec:triangular transport review}.

\subsection{Basic notation and definitions}\label{subsec: notation-and-defs}
Throughout, all Hilbert spaces are separable, all kernels and maps are measurable, and we assume the measurability conditions needed for the quantities considered below to be well defined. These quantities include Bochner integrals, (conditional) mean embeddings, and suprema over function classes, which can then be taken over countable dense subsets. Given a separable Hilbert space $\U$, we write $\PP(\U)$ to denote the space of Borel probability measures on $\U$. For a pair of separable Hilbert spaces $\Y,\U$, we write $\Y\times\U$ to denote their Cartesian product, endowed with the product Hilbert structure. For a measure $\rho\in\PP(\Y\times\U)$, we define the marginals $\rho_\Y\coloneqq P_\Y\#\rho$ and $\rho_\U\coloneqq P_\U\#\rho$, where $P_\Y:(y,u)\mapsto y$ and $P_\U:(y,u)\mapsto u$. Moreover, following~\cite[Ch.~10]{bogachev2007measure-volII}, we define the (regular) conditional measures $\rho(\cdot\mid y)$ as the class of probability transition kernels satisfying: (a) $\rho(\cdot\mid y)\in\PP(\U)$ for $\rho_\Y$-almost every (a.e.) $y$; (b) for all Borel sets $A\subseteq\U$, the map $y\mapsto\rho(A\mid y)$ is measurable; and (c) for all Borel sets $B\subseteq\Y\times\U$, we have the identity $\rho(B)=\int_\Y\rho(B_y\mid y)\,\rho_\Y(\dd y)$, where $B_y\coloneqq\{u\in\U\mid(y,u)\in B\}$.

Given a $\sigma$-finite measure space $(\Omega,\mathcal A,\mu)$, a separable Hilbert space $\mathcal X$ and $p\in[1,\infty)$, we denote by $L^p(\Omega,\mathcal X;\mu)$ the Bochner space of (equivalence classes of $\mu$-a.e.\ equal) strongly measurable maps $f:\Omega\to\mathcal{X}$ with
\begin{align*}
\|f\|_{L^p(\Omega,\mathcal{X};\mu)}\coloneqq\left(\int_\Omega\|f(\omega)\|_{\mathcal{X}}^p\,\mu(\dd\omega)\right)^{1/p}<\infty,
\end{align*}
and by $L^\infty(\Omega,\mathcal{X};\mu)$ the space of those with $\|f\|_{L^\infty(\Omega,\mathcal{X};\mu)}\coloneqq\operatorname*{ess\,sup}_{\omega\sim\mu}\|f(\omega)\|_{\mathcal{X}}<\infty$. The space $L^2(\Omega,\mathcal{X};\mu)$ is Hilbert with inner product $\langle f,g\rangle_{L^2(\Omega,\mathcal{X};\mu)}\coloneqq\int_\Omega\langle f(\omega),g(\omega)\rangle_{\mathcal{X}}\,\mu(\dd\omega)$. Finally, for $a,b\in\RR$, we write $a\vee b\coloneqq\max\{a,b\}$ and $a\wedge b\coloneqq\min\{a,b\}$, and $a\lesssim b$ means that $a\leq cb$ for a numerical constant $c>0$ independent of all quantities of interest.

\subsection{Scalar-valued RKHSs, mean embeddings, and MMD}\label{subseq: MMD RKHS}
The following is a brief review of RKHSs as they pertain to 
the article. For a more comprehensive treatment, see \cite{berlinet2011reproducing, owhadi2019operator,muandet2017kernel,park2020measure}. 
\begin{itemize}[leftmargin=*]
\item {\bf PSD and Mercer kernels.} Given a set $\Z$, a symmetric and positive semidefinite (PSD) kernel on $\Z$ is a function $K:\Z\times\Z\mapsto\RR$ such that $K(z,z')=K(z',z)$ for all $z,z'\in\Z$ and for any collection of points $Z:=\{z_i\}_{i=1}^m \subseteq \Z$ the kernel matrix $K(Z,Z)\in\RR^{m\times m}$, with entries $K(Z,Z)_{ij}=K(z_i,z_j)$, is positive semidefinite. We will sometimes use the term kernel to refer to PSD kernels when there is no risk of confusion. If additionally a kernel $K$ is continuous, we say that $K$ is Mercer.  We further 
introduce the compact notation 
$K(Z,z):=(K(z_1,z),...,K(z_m,z))^\top$
as a column vector field on $\Z$, 
and similarly $K(z,Z)=K(Z,z)^\top$ for the row vector field.
\item {\bf RKHSs and the reproducing property.} For each kernel $K$, there exists a unique Hilbert space $\KK$, called the RKHS of $K$, which can be identified as the completion of the pre-Hilbert space
\begin{align*}
\widetilde{\KK}:=\{z\mapsto c^\top K(Z,z)\quad |\quad c\in\RR^m, \; Z\subset \Z, \;| Z| = m, \; m\in\mathbb{N}\}\,,
\end{align*} 
with respect to the inner product 
\begin{align}\label{eq: inner product scalar RKHS}
\langle f,f'\rangle_{\widetilde{\KK}}=c^\top K(Z,Z')c'\,,
\end{align}
for functions $f = c^\top K(Z, \cdot)$ and $f' = c'^\top K(Z', \cdot)$ and matrix $K(Z,Z')_{ij} = K(z_{i}, z'_{j})$ 
following our 
usual notation. We use $\langle \cdot, \cdot \rangle_{\KK}$
and $\| \cdot \|_{\KK}$ to denote the RKHS inner product and 
induced norm on $\KK$ and note the reproducing property
\begin{align}\label{eq: reproducing property}
f(z)=\langle f,K(z,\cdot)\rangle_\KK\, ,\qquad \forall f\in\KK\, ,\forall z \in\Z\, ,
\end{align}
as a consequence of the definition which gives
$\| f \|_\KK^2 = c^\top K(Z, Z) c$ 
for functions in $\widetilde{\KK}$.

\item {\bf The feature map perspective}
A map $\phi: \Z \to \mF$, where $\mF$ is a Hilbert space, is a feature map for the kernel $K$ if 
$K(z, z') = \langle \phi(z), \phi(z') \rangle_\mF$ for all $z,z' \in \Z$. Then the RKHS 
$\KK$ has an equivalent characterization as $\KK = \{ f: \Z \to \RR \mid f(z) = \langle c , \phi(z) \rangle_\mF, \; \| c\|_\mF < + \infty \}$ with $c$ denoting 
the minimal-norm coefficients defining $f$. The RKHS inner product and the 
induced norm then take the form $\langle f, f' \rangle_\KK = 
\langle c, c' \rangle_\mF$ and $\|f \|_\KK = \| c \|_\mF$.

\item {\bf Kernel mean embeddings.} 
For a PSD kernel $K$ we define $\PP_K(\Z)\subseteq\PP(\Z)$ to be the space of Borel probability measures satisfying the 
integrability condition $\int_\Z \sqrt{K(z,z)}\rho(dz)<\infty\,$. Subsequently, we define the kernel mean embedding 
operator $I_K: \PP_K(\Z)\mapsto \KK$
over this class as the mapping
$I_K(\rho):=\int_{\Z}K(z,\cdot)\rho(dz)\,.$

\item {\bf Maximum mean discrepancies.}
Finally, we define the MMD as the RKHS norm difference between the mean embeddings of pairs of measures in $\PP_K(\Z)$.
More precisely, for any pair of measures $\rho, \rho'\in \PP(\Z)$ we define  
\begin{align}\label{eq: MMD def}
\MMD_K(\rho,\rho')&:= \begin{cases}
    \left\|I_K(\rho)-I_K(\rho')\right\|_{\KK}\, & {\text{if}} \quad \rho,\rho' \in \PP_K(\Z) \\ 
    + \infty & \text{Otherwise}.
\end{cases} 
\end{align}
Note that MMDs are not necessarily divergences, i.e., 
$\MMD_K(\rho,\rho') = 0 \not\Rightarrow \rho = \rho'$. If the 
divergence property holds  we then say that the 
underlying kernel $K$ is {\it characteristic}. 
Many practical kernels, such as the Mat{\'e}rn family, are known to be characteristic \cite{berlinet2011reproducing,gretton2012, minh2010some}. Moreover, in light of this definition, we have that 
\begin{align*}
\CD(\rho\pipe \rho') = \int \MMD_K^2(\rho(\cdot\mid y), \rho'(\cdot\mid y))\rho_\Y(\dd y), \ \JMMD_G^2(\rho,\rho') = \MMD_{K\otimes G}^2(\rho,\rho').
\end{align*}
\end{itemize}

\subsection{Triangular transport maps}\label{sec:triangular transport review}
Recall the definition of a triangular map (\Cref{def: block-triangular map}) with 
individual components $\T, \F$. Then the fundamental result relating triangular 
transport to conditioning, which is the cornerstone of transport based 
conditional simulation algorithms, is \cite[Thm.~2.4]{baptista2024conditional}. For completeness, we present an extension of this result to full generality which also provides a correction to the claim of~\cite[Prop. 2.3]{hosseini2023conditional} (holding only when $\F$ is invertible).

\begin{proposition}\label{prop:triangular-transport-conditioning}
Consider a reference  measure $\eta \in \PP(\Z \times \V)$
and target $\nu \in \PP(\Y \times \U)$, where $\Z, \V, \Y, \U$ are 
Hilbert spaces. Further consider a triangular map 
$\T_\t^\F(z, v) = (\F(z), \T(\F(z), v) )$ 
with components $\F: \Z \to \Y$ and $\T: \Y \times \V \to \U$. If 
 $\T_\t^\F \# \eta = \nu$, then it holds that $\F \# \eta_\Z = \nu_\Y$ and
\begin{align*}
\int_{\F^{-1}(\{y\})}\T(y,\cdot)\#\eta(\cdot\mid z)\widetilde{\eta}(dz \mid \F(z) = y) = \nu(\cdot\mid y)\, , \qquad \textnormal{for }\nu_\Y\textnormal{-almost every } y\in\Y\, .
\end{align*}
in terms of the conditional measure $\widetilde{\eta}(dz \mid \F(z) = y)$ induced by the disintegration~\footnote{ On conditional slices of non-zero measure, $\widetilde{\eta}(\dd z \mid \F(z)=y) = \frac{\mathds{1}\{\F(z) = y\}\eta_\Z(\dd z)}{\int \mathds{1}\{\F(z') = y\}\eta_\Z(\dd z')}$.}
\begin{align*}
\eta_{\Z}(\dd z) = \int_\Y \widetilde{\eta}(\dd z\mid \F(z) = y)(\F\#\eta_\Z)(\dd y)\, .
\end{align*}
Thus, if $\F$ is invertible,
\begin{align*}
\T(y,\cdot)\#\eta(\cdot\mid \F^{-1}(y)) = \nu(\cdot\mid y)\, ,\qquad \textnormal{for }\nu_\Y\textnormal{-almost every } y\in\Y\, .
\end{align*}
Moreover, if the reference is an independent coupling $\eta = \eta_\Z \otimes \eta_\V$, even if $\F$ is not invertible, 
\begin{align*}
\T(y,\cdot)\#\eta_\V = \nu(\cdot\mid y)\, ,\qquad \textnormal{for }\nu_\Y\textnormal{-almost every } y\in\Y\, .
\end{align*}
recovering the result of~\cite[Thm.~2.4]{baptista2024conditional}.

\end{proposition}

In practical algorithms we primarily work with the setting where $\Z = \Y$
and the reference measure has independent $z$ and $v$ components. Additionally, we can 
further simplify our setup by assuming that $\eta$ and $\nu$ have matching marginals. 
We collect these in the next assumption which we henceforth assume holds:

\begin{assumption}\label{assumption:independent-reference}
 The reference measure $\eta \in \PP(\Y \times \V)$ is an independent coupling of $\nu_\Y$ 
 and an arbitrary reference $\eta_\V \in \PP(\V)$, that is  $\eta=\nu_\Y\otimes\eta_\V$.\label{Independent coupling}
\end{assumption}

Under this assumption we can simply take $\F = \text{Id}$ which simplifies 
\Cref{prop:triangular-transport-conditioning} to the following 
statement: 

\begin{corollary}\label{cor: conditional transport independent reference}
Suppose \Cref{assumption:independent-reference} holds and 
consider 
\begin{equation}\label{eq:triangular-map-form}
\T_\t(y, v) = (y, \T(y, v)).    
\end{equation}
Then if $\T_\t \# \eta =\nu$ 
then $\T(y, \cdot) \# \eta_\V = \nu(\cdot \mid y)$ for $\nu_\Y$ a.e. $y$.
\end{corollary}

The existence of such triangular maps is not in general guaranteed 
although various characterizations exist, see for example \cite{hosseini2023conditional} for the OT existence result 
or \cite{bogachev2005} for a generalization of KR maps to abstract spaces. 
Since our focus is on the approximation of triangular maps 
we simply assume the existence of such a ground truth map. 

\begin{assumption}\label{assumption:triangular-map-existence}
    There exists a ground truth triangular map $\T^\dagger_\t(y, v) = (y, \T^\dagger(y, v) )$ such that $\T^\dagger_\t \# \eta = \nu$ so 
    that $\T^\dagger(y, \cdot) \# \eta_\V  = \nu(\cdot \mid y)$.
\end{assumption}

With these assumptions in place it is clear that the main object of 
interest in practice is the (bottom) component $\T$ in \eqref{eq:triangular-map-form}
and we will mainly study the approximation of this map in later sections. 
In this light it is helpful for us to define function classes of triangular 
maps induced by the $\T$ components. For a class $\TT$
of maps from $\Y \times \V \to \U$ we define 
its induced triangular class $\TT_\t:= \{ \T_\t(y, v) = (y, \T(y, v)) \mid \T \in \TT \}$.

\section{Problem setup}\label{sec: setup}

We consider learning an approximate triangular map $\Th:\Y\times\V\mapsto \U$ 
given finitely many i.i.d. samples from $\nu$ and $\eta$. That is, we would 
like $\Th(y, \cdot) \# \eta_\V \approx \nu(\cdot \mid y)$ for a.e. $y$. In this paper, how well this approximation is achieved is quantified by the notion of average conditional estimation error provided by the {\it averaged conditional divergence}:

\begin{definition}\label{def: averaged cond div}
Consider a statistical divergence 
$D_{\text{Cond.}}: \PP(\U) \times \PP(\U) \mapsto \RR_{\ge 0}$, then we define the averaged conditional divergence (ACD) 
associated to $D_{\text{Cond.}}$  as 
\begin{equation*}
    \DD: \PP(\Y \times \U) \times \PP(\Y \times \U) \mapsto \RR_{\ge 0}, 
    \quad \DD(\rho , \rho') = 
    \int_\Y D_{\text{Cond.}}( \rho(\cdot \mid y) \pipe \rho'(\cdot \mid y) ) \rho_\Y(\dd y).
\end{equation*}
\end{definition} 
The goal is then to derive theoretical guarantees on the error $\DD(\nu\pipe \Th_\t \#\eta)$. Henceforth we take 
$D_{\text{Cond.}}$ to be the squared MMD induced by a kernel $K$, i.e., 
\begin{equation}\label{ACD-MMD-2}
\begin{aligned}
    \DD(\rho \pipe \rho') 
    &= \int_\Y  \MMD_K^2 \left(\rho(\cdot \mid y) , \rho'(\cdot \mid y)\right) \rho_\Y(\dd y). 
\end{aligned}
\end{equation}
Notice that this ACD is precisely the $\CD$ divergence introduced in \eqref{def:EMCMD2} simply written in a more compact notation. As briefly anticipated before, the next proposition justifies the need to control $\DD(\nu\pipe \Th_\t\#\eta )$ in order to grant theoretical guarantees on the conditional generation capability of $\Th$.

\begin{proposition}\label{prop: fixed sampled conditional control}
Suppose there exists an algorithm yielding a random estimator $\Th\in\TT$, trained with target samples $\{(y_i,u_i)\}_{i=1}^m\overset{\text{i.i.d.}}{\sim}\nu$, together with $\mathfrak{r}>0$ and a nondecreasing $g:[0,\infty)\to[0,\infty)$, such that for any $t>0$, with probability at least $1-e^{-t}$,
\begin{align*}
\DD\left(\nu\pipe\Th_\t\#\eta\right)\leq\E_{\text{Bias}}^2+g(t)m^{-1/\mathfrak{r}}.
\end{align*}
Define the pointwise bias and the pointwise-to-averaged bias gap
\begin{align*}
\E_{\text{Bias}}^2(y)\coloneqq\inf_{\T\in\TT}D_{\text{Cond.}}\left(\nu(\cdot\mid y),\T(y,\cdot)\#\eta_\V\right),\qquad\Gamma_\TT\coloneqq\E^2_{\text{Bias}}-\int_\Y\E^2_{\text{Bias}}(y)\,\nu_\Y(\dd y)\geq0.
\end{align*}
Then, for $y\sim\nu_\Y$ independent of the training samples, for any $t>0$ and $\delta>0$, it holds with probability at least $1-2e^{-t}$ that
\begin{align*}
m\geq\left(\frac{e^{t}g(t)}{\delta}\right)^{\mathfrak{r}}\qquad\Longrightarrow\qquad D_{\text{Cond.}}\left(\nu(\cdot\mid y),\Th(y,\cdot)\#\eta_\V\right)\leq\E_{\text{Bias}}^2(y)+e^t\,\Gamma_\TT+\delta.
\end{align*}
\end{proposition}

\begin{remark}[Controlling $\DD$ vs. pointwise control and decomposable classes]\label{rem: limitations in controlling the averaged error}
From inspecting~\Cref{prop: fixed sampled conditional control}, it becomes immediately clear that the term $e^{t} \Gamma_\TT$ is the natural price that we have to pay in obtaining a high probability control over the point-wise error $D_{\text{Cond.}}\left(\nu(\cdot\mid y),\Th(y,\cdot)\#\eta_\V\right)$ by not minimizing it directly and controlling only the averaged error $\DD(\nu \pipe \Th_\t\#\eta)$. However, such price can be mitigated (or completely eliminated) when the pointwise-to-averaged bias gap $\Gamma_\TT$ is small (or totally vanishes). To see when this is possible, abbreviating $\E_\T^2(y)\coloneqq D_{\text{Cond.}}\left(\nu(\cdot\mid y),\T(y,\cdot)\#\eta_\V\right)$, we can re-express
\begin{align*}
0\leq \Gamma_\TT = \inf_{\T\in\TT}\int\left(\E_\T^2(y)-\E_{\text{Bias}}^2(y)\right)\nu_\Y(\dd y)\leq \E_{\text{Bias}}^2.
\end{align*}
when $\E_\T^2(y)$ is non-negative and finite for any $\T\in\TT$ (e.g., $D_{\text{Cond.}} =\MMD_K^2$  for bounded $K$). From this consideration, in the well-specified case $\E_{\text{Bias}} = 0$, we immediately obtain $\E_{\Th}^2(y)\leq \delta$ with high probability. Furthermore, even in the misspecified case $\E_{\text{Bias}} > 0$, $\Gamma_\TT =0$ if and only if the interchangeability condition
\begin{align}\label{eq: interchangeability condition}
\inf_{\T\in\TT}\int \E_\T^2(y) \nu_\Y(\dd y) = \int \inf_{\T\in\TT}\E_\T^2(y) \nu_\Y(\dd y),
\end{align}
holds, in which case $\E_{\Th}^2(y)\leq \E_{\text{Bias}}^2(y)+ \delta$ with high probability. General sufficient conditions for~\eqref{eq: interchangeability condition} to hold can be found in~\cite{giner2009necessary}. In particular, a prominent sufficient requirement is when $\TT$ is \emph{decomposable/closed under gluing}~\cite{hiai1977integrals}, i.e., for all $\T,\S\in\TT$ and every measurable  $A\subseteq \Y$, $\T_A(y,v) = \mathbf{1}_A(y)\T(y,v)+\mathbf{1}_{A^c}(y)\S(y,v)\in\TT$. Since $\E_\T^2(y)$ depends on $\T$ only through $\T(y,\cdot)$, choosing $A=\{\E_{\T}^2\leq \E_{\S}^2\}$ gives $\E_{\T_A}^2=\min\{\E_{\T}^2,\E_{\S}^2\}$. A decomposable class can thus retain the best available map at each $y$, which is the mechanism behind the interchange. For instance, given any set $\mathcal S$ of maps from $\V$ to $\U$, the class of all measurable $\T$ such that $\T(y,\cdot)\in\mathcal S$ for $\nu_\Y$-a.e. $y$ is decomposable, since gluing preserves this slice-wise constraint. Finite-dimensional parametric classes are typically not decomposable, because the dependence of $\T(y,\cdot)$ on $y$ is tied together by shared parameters. For instance, gluing the maps $\T(y,v)=y+v$ and $\S(y,v)=-y+v$ of the affine class $\{(y,v)\mapsto ay+b+v\}$ along $A=\{y>0\}$ yields $\T_A(y,v)=|y|+v$, which is no longer affine in $y$. As a consequence, improving the fit at some values of $y$ may deteriorate it at others, and $\Gamma_\TT>0$ in general. Nevertheless, for classes that are rich enough to be dense in a decomposable class, such as neural networks by universal approximation results~\cite{hornik1991approximation} for example, $\Gamma_\TT$ can be made arbitrarily small.
\end{remark}

Next, we present in more detail the two different setups of the previously introduced problems towards the estimation of triangular maps which we intend to analyze: namely $\CD$ and $\JMMD_G^2$ minimization, as defined respectively in~\eqref{def:EMCMD2} and ~\eqref{eq: Joint MMD squared}.

\subsection{Estimating triangular maps by minimizing average conditional MMDs}
\label{sec: triangular maps and cond. div. min.}

The first approach considers estimators directly minimizing the notion of average conditional estimation error we seek to control towards the assessment of the estimated triangular maps, namely estimators of the form 
\begin{equation}\label{eq:population-map-est}
    \Th  \in \argmin_{\T \in \TT} \DD\left(\nu \pipe \T_\t \# \eta\right),
\end{equation}
where $\TT$ is an appropriate class of transport maps. Note that if 
$\T^\dagger \in \TT$ then this problem has a solution that achieves 
zero loss.
This estimator is natural for conditioning problems as it minimizes the averaged divergence between the conditionals of $\nu$ and those generated by the map $\T$; akin to a 
minimum mean-squared-error (MSE) estimator. 

Following \cite{park2020measure}, we substitute the expression for $\MMD^2$ 
and use Jensen's inequality on the integrand to obtain the bound
\begin{equation}\label{eq:M-derivation}
\begin{aligned}
    \DD(\rho \pipe \rho')  
    &=  \int_\Y \left\| I_K(\rho(\cdot \mid y)) - I_K(\rho'(\cdot \mid y))  \right\|_\KK^2 \rho_\Y(\dd y) \\
    &=  \int_\Y \left\| \int_\U K(u, \cdot) \rho(\dd u \mid y) 
     - I_K(\rho'(\cdot \mid y)) \right\|_\KK^2 \rho_\Y (\dd y) \\
     & \le \int_{\Y \times \U}  \left\| K(u, \cdot) - I_K(\rho'(\cdot \mid y)) \right\|_\KK^2 \rho(\dd u, \dd y) =: M( \rho \pipe \rho').
\end{aligned}
\end{equation}
The quantity $M: \PP(\Y \times \U) \times \PP(\Y \times \U) \to \RR_{\ge 0}$
is not a proper divergence and here we refer to it 
as the {\it surrogate loss or divergence}. The reason for this terminology 
is the observation that, by direct calculation (see~\cite[Sec. 4.1]{park2020measure}), we can compute the gap  
\begin{equation}\label{disp:jensen-gap}
    M(\rho \pipe \rho') - \DD(\rho \pipe \rho') 
    = \int_\U \| K(u, \cdot) \|_\KK^2 \rho_\U(\dd u) 
    - \int_\Y \left\| \int_\U K(u, \cdot) \rho(\dd u \mid y) \right\|_\KK^2 
    \rho_\Y (\dd y).
\end{equation}
Since this gap is only a function of  $\rho$ and not $\rho'$ 
we infer that 
\begin{equation*}
    \Th \in \argmin_{\T \in \TT} \DD\left( \nu \pipe \T_\t \# \eta\right) 
    =\argmin_{\T \in \TT} M\left( \nu \pipe \T_\t \# \eta\right). 
\end{equation*}
In other words, $M$ can serve as a surrogate loss for $\DD$ in the training of our maps.
    To see why we need the  $M$ at all, recall our generative modeling 
    setup where we only have access to joint samples $(y_i, u_i) \sim 
    \nu$. In this setting accurate evaluation of $\DD$ is not possible
    since we (often) have only one $u_i$ sample for each value of $y_i$
    and so the mean embedding $I_K(\nu(\cdot \mid y))$ cannot be 
    computed to high accuracy. On the other hand, the quantity $M$ only requires 
    $I_K( \T(y, \cdot) \# \eta_\V)$ which can be simulated 
    with arbitrary samples since $\eta_\V$ and $\T$ are both prescribed 
    in the algorithm. We also emphasize that due to the gap \eqref{disp:jensen-gap},  the value of 
$M$ will in general not be zero at the minimizer. 

\begin{remark}[Probabilistic interpretation of $M$]\label{rem: cond. var. remark}
Notice that the gap~\eqref{disp:jensen-gap} is a variance-type quantity: it equals $\mathbb{E}_{(Y,U)\sim \rho}\left\|K(U,\cdot) - \mathbb{E}[K(U,\cdot)\mid Y]\right\|_\mathcal{K}^2$, the average conditional variance of $K(U,\cdot)$ under $\rho$. It vanishes when $U$ is $\rho$-a.s. a function of $Y$, and is maximal, among couplings with the same $\U$-marginal, when $\rho = \rho_\Y\otimes \rho_\U$. 

Introducing the random variables $(Y,U)\sim\rho$ and the $Y$-measurable estimator $g_{\rho'}(Y):=I_K(\rho'(\cdot\mid Y)) \in \mathcal K$ (parameterized 
by $\rho'$), the surrogate loss may be expressed as 
\begin{align*}
    M(\rho \pipe \rho') = \Expect_{(Y,U)} \|K(U,\cdot) - g_{\rho'}(Y)\|^2_{\mathcal K} =: M(g_{\rho'})
\end{align*}
which is equal to the mean-squared-error for estimating $K(U,\cdot)$ in terms of $g(Y)$. The minimizer is 
unique $\rho_\Y$-a.e and it equals the conditional expectation $g^\star(Y) = \Expect[K(U,\cdot)|Y]$, with  the conditionals $\rho'(\cdot\mid y)$ being also unique if $K$ is characteristic. Then, $M(g^\star)$ is precisely 
the average conditional variance of $K(U,\cdot)$ measured in the 
RKHS norm; see \cite{park2020measure}. Taking $\rho=\nu$ and $\rho'=\T_\t\#\eta$, the estimator is expressed in terms of the map $\T$ according to 
$g_{\rho'}(y) = I_K(\T(y,\cdot)\#\eta_\V) = \Expect_{V\sim \eta_\V}K(\T(y,V),\cdot)$,
and the optimal estimator $g^\star(y)$ is achieved when the transport map $\T$ satisfies $\T(y,\cdot)\#\eta_\V = \nu(\cdot|y)$, for $\nu_\Y$-a.e. $y\in \Y$.
\end{remark}
We are now ready to define a series of approximation problems 
that successively approach a practical algorithm but each of these plays an important role in our theoretical analysis. 
Since $\eta_\V$ is chosen by the user and is fixed, we slightly 
abuse notation and write $M(\nu, \T) \coloneqq M(\nu \pipe \T_\t \# \eta)$. Similarly, we let $\DD(\nu, \T)\coloneqq \DD(\nu \pipe \T_\t \#\eta)$. This modification is convenient since we think of $\nu$ or its 
empirical approximations as training data and $\T$ as the 
parameter that we are trying to fit.

\begin{problem}[Population level $\CD$ minimization]
\label{prob: population level constrained minimization}
Fix kernel $K$ and consider the approximation class $\TT$.
Find 
\begin{equation*}
    \Th :\in \footnote{We assume minimum attainment. However, all the arguments presented can be trivially extended letting $\Th$ satisfy $\M(\nu, \Th)\leq \inf_{\T\in \TT} \M(\nu, \T) + \xi$ and letting $\xi \to 0^{+}$. The same applies to the following minimization problems throughout the paper.} \argmin_{\T \in \TT} M( \nu, \T ). 
\end{equation*}
\end{problem}

Since we are interested in conditional generation we further 
assume that $\nu$ is not accessible but rather an empirical approximation
$\nu^m$ is given as a training data set. 
\begin{problem}[$\CD$ idealized empirical minimization]
\label{prob: empirical minimization}
Consider the empirical measure $\nu^m = \frac{1}{m} \sum_{i=1}^m \delta_{(y_i, u_i)}$ with $(y_i, u_i) \overset{\textnormal{i.i.d.}}{\sim} \nu$  and fix the 
kernel $K$ and the class $\TT$. Find, 
\begin{align*}\label{eq:constrained-min}
\Th^{m}:\in\argmin_{\T\in\TT}\M(\nu^m, \T)\,.
\end{align*}
\end{problem}
Finally, we also replace the reference measure $\eta_\V$ with an 
empirical approximation to obtain the following practical problem in terms of an unbiased estimator of $\M$: 
\begin{problem}[$\CD$ fully empirical minimization]\label{prob: empirical minimization 2}
    Consider the i.i.d. samples $v_{i,j} \sim \eta_\V$ 
    for $i =1, \dots, m$ and $j = 1, \dots, n$ and define the unbiased
    empirical loss approximation
    \begin{align}\label{eq: M unbiased estimate with n}
    \begin{split}
    M^n(\nu^m, \T)\coloneqq \frac{1}{m}\sum_{i=1}^m \Bigg(K(u_i,u_i)&-2\frac{1}{n}\sum_{j=1}^n K(\T(y_i, v_{i,j}), u_i) \\
    &+\frac{1}{n(n-1)}\sum_{j=1}^n\sum_{\substack{k=1\\
    k\neq j}}^n K(\T(y_i, v_{i,j}), \T(y_i, v_{i,k})) \Bigg)\, .
    \end{split}
    \end{align}
    Then our goal is to find 
    \begin{equation}\label{eq:empirical-opt}
        \Th^{m, n} :\in \argmin_{\T \in \TT} 
        M^n (\nu^m, \T).
    \end{equation}
\end{problem}
With this setup, our theoretical analysis in~\Cref{sec:error-analysis} 
provides non-asymptotic error bounds on the quantity $M(\nu, \Th^{m,n})$ in terms of both sample sizes $m$ and $n$
and, thanks to \eqref{disp:jensen-gap}, the original 
divergence $\DD(\nu, \Th^{m, n})$ which shares the 
same convergence rate.

\subsection{Estimating triangular maps by minimizing joint MMDs}\label{sec: triangular maps and joint div. min.}
The second approach we consider bypasses the direct minimization of the average conditional estimation error we ultimately seek to control. Instead, the estimator is found by minimizing the squared $\JMMD_G^2$ introduced in~\eqref{eq: Joint MMD squared}, which we abbreviate for notational convenience to $\DG$. Also in this case, since we select and fix $\eta_\V$ and the remaining marginal $\eta_\Y = \nu_\Y$, with a slight abuse of notation, we denote $\DG(\nu, \T)\coloneqq \DG(\nu, \T_\t\#\eta)$.

\begin{problem}[Population level $\JMMD_G^2$ minimization]
\label{prob: population level joint minimization}
Together with the kernel $K$, fix an additional kernel $G:\Y\times\Y\to\mathbb{R}$ and consider the approximation class $\TT$. Find
\begin{equation*}
    \Th_G :\in \argmin_{\T \in \TT} \DG( \nu, \T). 
\end{equation*}
where $\DG$ is just a compact notation for $\JMMD_G^2$ defined in~\eqref{eq: Joint MMD squared} via the tensor product kernel $K\otimes G$ of~\eqref{eq: joint product kernel}
\end{problem}

Analogously to $\CD$ minimization, in practice, the minimization is carried out using samples since the underlying distributions are assumed to be not accessible. Thus, we state the following problem, where $\nu$ and $\eta$ are replaced by the empirical measures $\nu^m$ and $\eta^m$.

\begin{problem}[$\JMMD_G^2$ empirical minimization]
\label{prob: empirical level joint minimization}
Together with the kernel $K$ and approximation class $\TT$, fix an additional kernel $G:\Y\times\Y\to\mathbb{R}$. Consider the i.i.d.\ samples $v_j\sim\eta_\V$ and $(y_i,u_i)\sim\nu$ for $i,j=1,\dots,m$, let $\eta^m\coloneqq\frac1m\sum_{j=1}^m\delta_{(y_j,v_j)}$, and define the empirical approximation
\begin{align}\label{eq: empirical joint MMD loss}
\begin{split}
\DG^m(\nu^m,\T)\coloneqq\DG(\nu^m,\T_\t\#\eta^m)&= \frac{1}{m^2}\sum_{i=1}^m\sum_{i'=1}^m K(u_i,u_{i'})G(y_i,y_{i'}) \\
& \quad -\frac{2}{m^2}\sum_{i=1}^m \sum_{j=1}^m K\left(u_i, \T(y_j,v_j)\right)G(y_i,y_j)\\
& \quad + \frac{1}{m^2}\sum_{j=1}^m \sum_{j'=1}^m K\left(\T(y_j, v_j),\T(y_{j'}, v_{j'})\right)G(y_j, y_{j'}).
\end{split}
\end{align}
Then the goal is to find
\begin{equation}\label{eq: joint minimizer from samples}
    \Th_G^m :\in \argmin_{\T \in \TT} \DG^m(\nu^m, \T). 
\end{equation}
\end{problem}

\begin{remark}\label{rem: biased and unbiased estimators}
We use the V-statistic~\eqref{eq: empirical joint MMD loss}, i.e., the squared $\MMD_{K\otimes G}$ between the empirical measures $\nu^m$ and $\T_\t\#\eta^m$, as implemented in~\cite{pandey2024diffeomorphic}. It is a biased estimator of $\DG(\nu,\T)$, with bias of order $1/m$ uniformly in $\T$, but its square root is a pseudometric between empirical measures, which keeps the generalization analysis of~\Cref{sec: DD vs DG minimization} elementary. U-statistic variants, including an unbiased one, lead to the same results; see~\Cref{rem: other joint estimators}.
\end{remark}

Optimizing $\DG = \JMMD_G^2$, does not directly minimize the notion of average conditional estimation error provided by $\DD$ in~\Cref{def: averaged cond div}. However, in principle, this approach can yield perfect recovery of the target conditionals with respect to $\DD$. Indeed, for characteristic $K\otimes G$, thanks to the independent coupling structure of the reference measure $\eta = \nu_\Y\otimes\eta_\V$ and the structure of triangular transport maps, 
\begin{align*}
\DG(\nu, \T) = 0\, ,\qquad \Longrightarrow \qquad \T(y,\cdot)\#\eta_\V = \nu(\cdot\mid y)\, ,\quad \textnormal{for }\nu_{\Y}\textnormal{-almost every }y\in \Y, 
\end{align*}
as stated by~\Cref{cor: conditional transport independent reference}.
Nevertheless,~\Cref{ex:counter-CMMD-counter-example} points out that, in general,
\begin{align*}
\DG\left(\nu, \T\right) \leq \epsilon\, ,\qquad \centernot\Longrightarrow\qquad \DD\left(\nu, \T\right)\lesssim\epsilon\, . 
\end{align*}
Thus, when~\Cref{prob: empirical level joint minimization} is not carried out to reach $\DG(\nu, \Th^m_{G})=0$ for reasons such as the bias of the hypothesis class $\TT$, error due to finite samples, there is no guarantee that the triangular transport map learned achieves a controllable average conditional estimation error.

In~\Cref{sec: DD vs DG minimization} we formalize this consideration showing that minimizing $\DG$ yields inconsistent estimators and we verify when (and at what price) a control of the average conditional estimation error can be obtained imposing further structure on the joint divergence $\DG$ and the regularity of the conditioning variable's marginal distribution $\nu_\Y$. 

\begin{remark}[Joint and conditional errors beyond static transport]\label{rem: joint vs conditional beyond static}
The failure of the implication above is not specific to static triangular maps or to MMD-based training criteria. It arises whenever the error is measured by a divergence that does not decompose over the conditionals. For the Kullback--Leibler divergence, the chain rule shows that the joint divergence between two measures sharing the same $\Y$-marginal equals the average conditional divergence, so that a small joint error does imply a small average conditional error. For integral probability metrics such as MMDs and Wasserstein distances, instead, the two notions can be far apart, as~\Cref{ex:counter-CMMD-counter-example} shows. Consequently, guarantees on the conditionals produced by other transport models, such as the dynamic formulation of flow matching with triangular velocity fields, also require an analysis tailored to the average conditional estimation error, rather than to the joint one; we return to this point in~\Cref{rem: dynamic transport}.
\end{remark}

\section{Error analysis of the $\CD$ minimizer}\label{sec:error-analysis}
In this section we present a master theorem that controls the error of the 
empirical map $\Th^{m, n}$, solution of~\Cref{prob: empirical minimization 2}. In particular, we show that 
\begin{equation*}
    \DD\left( \nu, \Th^{m, n}\right) \lesssim \E_{\text{Bias}}^2+ \text{generalization error}, 
\end{equation*}
where the bias term $\E_{\text{Bias}}^2\coloneq \inf_{\T\in\TT} \DD(\nu, \T) = \DD(\nu, \Th)$ quantifies the approximation error of the model class $\TT$, and the generalization error controls the 
stochastic error due to the random training data. Our analysis follows classic ideas 
from empirical process theory \cite{wainwright2019high, vanderVaartWellner2023} based on the observation we made earlier 
that $M^n(\nu^m, \T)$  is an empirical approximation to a regression loss for the mean embedding 
of $\nu(\cdot \mid y)$; this can also be seen by observing that
\eqref{eq: M unbiased estimate with n} is a $U$-statistic. Thanks to \eqref{disp:jensen-gap} we can then 
bound $M(\nu, \Th^{ m,n})$ to obtain a bound on $\DD(\nu, \Th^{m, n})$.
However, challenges arise in bounding the generalization error: one has to obtain a uniform law of large numbers over the randomness of both the training data $\{(y_i,u_i)\}_{i=1}^m$ and the independent samples approximating the conditional mean embeddings $\{v_{i,j}\}_{i=1}^m ,_{j=1}^n$ in terms of the complexity of the Hilbert-valued class of functions $y\to \T(y,\cdot)$ induced by $\TT$.

\subsection{Assumptions and setup}\label{sec:error-analysis-assumptions-and-setup}
We collect some notation and assumptions before presenting our theoretical results.
We let $\HH$ be a Hilbert space of functions $\Hsf: \V \mapsto \U$ such that $T(y,\cdot)\in \HH$ for $\nu_\Y$-almost every $y\in\Y$. Throughout the paper, we identify with a slight abuse of notation the hypothesis class $\TT: \Y\times \V\to \U$ with its isomorphic class of maps from $\Y$ to $\HH$, $\{y\to \T(y,\cdot) \mid \T\in\TT\}$. Moreover, we introduce the shorthand notation for the transport map-parameterized embeddings 
\begin{align}\label{eq: short embeddings}
g_\T(y)\coloneqq I_K(\T(y,\cdot)\#\eta_\V)
\end{align}
and let $\mathcal{G}_\TT\coloneqq \{g_\T \mid \T\in\TT\}$ be the lifted embedding class induced by $\TT$. Next, following \cite{wainwright2019high}, we define the covering number of $\TT$ with respect to a general metric $\|\cdot\|$ equipping $\TT$ as
    \begin{equation*}
        \N\left (\epsilon, \TT, \| \cdot \|\right) := \min \left\{ | C | \mid
        C \text{ is an $\epsilon$-cover of } \TT \text{ w.r.t. }\|\cdot\| \right\}, 
    \end{equation*}
    The metric entropy of $\TT$ is then simply the logarithm of the covering number,
    \begin{align}\label{eq: eps-entropy}
    \Hf(\epsilon, \TT, \|\cdot\|):= 
    \log \N\left(\epsilon, \TT, \|\cdot\| \right) \, .
    \end{align}
    As stated below, we assume a control of the metric entropy of $\TT$ with respect to the uniform Bochner norm $\|\T\|_{\infty} \coloneqq \|\T\|_{L^\infty(\Y, \HH; \nu_\Y)} = \esssup_{y\sim\nu_\Y}\|\T(y,\cdot)\|_{\HH}$. Analogously, we also consider the metric entropy of the lifted embedding class $\mathcal{G}_\TT$, but with respect to the uniform norm defined by $\|g_\T\|_{\infty} \coloneqq \|g_\T\|_{L^\infty(\Y, \mathcal{K}; \nu_\Y)} = \esssup_{y\sim\nu_\Y}\|g_\T(y)\|_{\mathcal{K}}$.

With this setup we are now ready to state our first set of technical assumptions which yield a desired upper bound on the average conditional estimation error of the estimator $\Th^{m}$, i.e., when $n = \infty$.

\begin{assumption}\label{Assumptions} 
It holds that:
\begin{enumerate}
\item \label{Boundedness kernel}\emph{(Boundedness of $K$)} There exists  $B_K\geq 1$ such that
$\sup_{u \in \U} K(u,u) \le B_K^2$ \footnote{Notice that the reproducing property implies that this condition is 
equivalent to $\sup_{u\in \U}\|K(u,\cdot)\|_\mathcal{K}\leq B_K$.}.
\item \label{lipshitzness} \emph{(Lipschitzness of $K$)}
There exists a constant $L_K\geq 1$ such that 
\begin{align*}
\|K(u,\cdot)-K(u',\cdot)\|_{\mathcal{K}}\leq L_K|u-u'\|_\U\, \qquad \forall u,u'\in\U\, .
\end{align*}

\item \label{embedding} \emph{(Embedding of $\HH$)} 
$\HH\hookrightarrow L^1(\V, \U; \eta_\V)$, i.e., there exists a constant $C_{1,\eta_\V} \geq 1$ such that 
\begin{align*}
\|h\|_{L^{1}(\V,\U; \eta_\V)}\leq C_{1,\eta_\V} \|h\|_{\HH},\qquad \forall h \in \HH.
\end{align*}

\item \label{Entropy lifted} \emph{(Entropy of the lifted class $\mathcal{G}_\TT$)} There exist constants $C_{\mathcal{G}_\TT}\geq 1$ and $\gamma\in [0,1)$ such that
\begin{align*}
\Hf(\delta, \mathcal{G}_\TT, \|\cdot\|_\infty) \leq C_{\mathcal{G}_\TT}\delta^{-\gamma}\log(1 + \delta^{-1}),\qquad \forall \delta>0.
\end{align*}
 
\end{enumerate}
\end{assumption}

\begin{remark}\label{rem:validity-of-population-assumptions}
The boundedness and Lipschitz conditions in~\Cref{Boundedness kernel} and~\Cref{lipshitzness} of~\Cref{Assumptions} are imposed mainly for simplicity of presentation. In the proof of~\Cref{thm: main error bound 2}, which we outline in~\Cref{subsec: high-level strategy} and treat fully in~\Cref{app: sec4}, they serve two purposes. First, they make the loss uniformly bounded, so that a constant envelope can be used in the one-shot localization argument we employ to control the estimators' generalization error (see~\Cref{thm: one-shot localization}). Second, they allow the entropy of the relevant function classes considered in our proof to be controlled in the uniform norm, therefore enabling simple Dudley-type bounds (see~\Cref{prop: dudley general}) yielding the necessary conditions towards the application of the one-shot localization. Both tools are stated in~\Cref{app: sec4} in a more general form than we need: the localization theorem allows envelopes with Orlicz-type tails, and the Dudley-type bound allows weighted uniform norms (see~\Cref{rem: on the localization statement} and~\Cref{rem: on the dudley statement}). This makes it possible to relax both conditions, e.g., to kernels or transport maps that are unbounded or only locally Lipschitz with controlled growth, provided the target and reference measures have sufficiently light tails, at the price of additional logarithmic factors in the resulting rates. Since the weighted form of~\Cref{prop: dudley general} is stated in~\cite{divol2025optimal-sm} for finite-dimensional data, this relaxation applies as stated when $\Y$, $\U$ and $\V$ are finite-dimensional.

The embedding condition in~\Cref{embedding} is used to control the bias term $\E_{\text{Bias}}$ in terms of the distance of $\TT$ to $\T^\dagger$ (see~\Cref{lem: app bias}). Finally, the restriction $\gamma<1$ in~\Cref{Entropy lifted} only simplifies the evaluation of the entropy integrals in the Dudley-type bounds, which otherwise diverge at small scales. For $\gamma\geq 1$, these integrals can be truncated at a positive scale using~\Cref{prop: dudley general}, which leads, up to logarithmic factors, to rates of order $(\log(m)/m)^{\frac{2}{2+\gamma}}\vee m^{-1/\gamma}$ (see~\cite{divol2025optimal-sm} for an example).
\end{remark}

We further need the following assumptions regarding the empirical 
approximations to the reference measure $\eta_\V$, i.e., the conditions yielding a bound on the average conditional estimation error of $\Th^{m,n}$ for $n<\infty$.

\begin{assumption}\label{assum: strengthened assum}
\begin{enumerate}
\item \label{inf embedding} \emph{(Strengthened embedding of $\HH$)} $\HH\hookrightarrow L^\infty(\V, \U; \eta_\V)$, i.e., there exists a constant $C_{\infty,\eta_\V} \geq 1$ such that 
\begin{align*}
\|h\|_{L^{\infty}(\V,\U; \eta_\V)}\leq C_{\infty,\eta_\V} \|h\|_{\HH},\qquad \forall h \in \HH.
\end{align*}
\item \label{entropy hypothesis} \emph{(Entropy of the hypothesis class $\TT$)} There exist constants $C_{\TT}\geq 1$ and $\gamma\in [0,1)$ such that
\begin{align*}
\Hf(\delta, \TT, \|\cdot\|_\infty) \leq C_{\TT}\delta^{-\gamma}\log(1 + \delta^{-1}),\qquad \forall \delta>0.
\end{align*}
\end{enumerate}
\end{assumption}
\begin{remark}
The conditions in~\Cref{assum: strengthened assum} are only needed to control the additional error due to approximating the conditional mean embeddings with the reference samples $v_{i,j}\sim\eta_\V$, i.e., for the bound on $\Th^{m,n}$ in~\Cref{thm: main error bound 2}. Since $M^n(\nu^m,\T)$ evaluates the feature map at individual points $\T(y_i,v_{i,j})$, the increments of the corresponding error process in $\T$ must be controlled for almost every reference sample, and not only on average over $\eta_\V$. This is guaranteed by the embedding $\HH\hookrightarrow L^\infty(\V,\U;\eta_\V)$ in~\Cref{inf embedding}, see~\Cref{lem: app reference decomposition}. The entropy condition~\Cref{entropy hypothesis} on $\TT$ is then transferred to the class of reference-sampling errors in~\Cref{prop: app reference error}.

Moreover, under~\Cref{lipshitzness} and~\Cref{embedding} of~\Cref{Assumptions}, the entropy condition~\Cref{entropy hypothesis} on $\TT$ implies the entropy condition~\Cref{Entropy lifted} on the lifted class $\mathcal{G}_\TT$, with the same exponent $\gamma$. Indeed, by~\Cref{lem: app bias}, $\|g_\T-g_\S\|_\infty\leq\lambda\|\T-\S\|_\infty$ with $\lambda\coloneqq L_KC_{1,\eta_\V}$, so that the image of a $(\delta/\lambda)$-net of $\TT$ is a $\delta$-net of $\mathcal{G}_\TT$. Thus,
\begin{align*}
\Hf(\delta,\mathcal{G}_\TT,\|\cdot\|_\infty)\leq\Hf\left(\frac\delta\lambda,\TT,\|\cdot\|_\infty\right)&\leq C_\TT\,\lambda^\gamma\delta^{-\gamma}\log\left(1+\lambda\delta^{-1}\right)\\
&\leq C_\TT\,\lambda^\gamma\max\{1,\lambda\}\,\delta^{-\gamma}\log\left(1+\delta^{-1}\right),
\end{align*}
where the last step uses $\log(1+\lambda x)\leq\max\{1,\lambda\}\log(1+x)$ for $x\geq0$. Therefore,~\Cref{Entropy lifted} holds with $C_{\mathcal{G}_\TT}=\max\{1,C_\TT\lambda^\gamma\max\{1,\lambda\}\}$, and \Cref{entropy hypothesis} in \Cref{assum: strengthened assum} subsumes \Cref{Entropy lifted} in \Cref{Assumptions}. Examples of classes of Hilbert-valued functions satisfying entropy conditions of this type, such as classes of smooth maps of the conditioning variable, are provided in~\cite{park2023towards}.
\end{remark}

\subsection{Average conditional estimation error bound of $\Th^m$ and $\Th^{m, n}$}\label{subsec:main-error-bound}
We are now ready to state our first main theoretical result giving an error bound for $\Th^m$ and the practical estimator $\Th^{m,n}$.

\begin{theorem}[Error bound for $\widehat{\T}^{m}$ and $\widehat{\T}^{m,n}$]\label{thm: main error bound 2}
Suppose that~\Cref{Assumptions} holds and let $m\geq3$. Then, for any $t>0$, it holds with probability at least $1-e^{-t}$ that
\begin{align*}
\DD\left(\nu, \Th^m \right) \lesssim \E_{\text{Bias}}^{2}+ \left(1 + t\right)^2 \left(\left(\frac{\C_1\log (m)}{m}\right)^{\frac{2}{2+\gamma}} +  \frac{\C_2\log(m)}{m}\right),
\end{align*}
where ``$\lesssim$'' absorbs a universal constant,
\begin{align*}
\C_1\coloneqq C_\mathcal{L}B_K^{2-\gamma},\qquad \C_2\coloneqq \frac{C_\mathcal{L}B_K^2}{(1-\gamma)^2},\qquad C_\mathcal{L}\coloneqq 2C_{\mathcal{G}_\TT}(8B_K)^\gamma\log(e+8B_K),
\end{align*}
and, with $\T^\dagger$ as in~\Cref{assumption:triangular-map-existence} satisfying $\T^\dagger(y,\cdot)\in\HH$ for $\nu_\Y$-a.e.\ $y$,
\begin{align*}
\E_{\text{Bias}} \leq L_K C_{1, \eta_\V}\, \textnormal{dist}_{L^\infty(\Y, \HH; \nu_\Y)}\left(\TT, \T^\dagger\right).
\end{align*}
Moreover, further supposing~\Cref{assum: strengthened assum}, for any $t>0$ and $n\geq2$, it holds with probability at least $1-2e^{-t}$ that
\begin{align*}
\DD\left(\nu, \Th^{m,n} \right) \lesssim \E_{\text{Bias}}^{2}+ \underbrace{\left(1 + t\right)^2 \left(\left(\frac{\C_1\log (m)}{m}\right)^{\frac{2}{2+\gamma}} + \frac{\C_2\log(m)}{m}\right)}_{\text{Sampling}} + \underbrace{\frac{\C_3}{m}+\frac{\C_4 + B_K^2\sqrt{t}}{\sqrt{mn}}}_{\text{Embedding Approx.}},
\end{align*}
where, with $\Lambda_\infty\coloneqq L_KC_{\infty,\eta_\V}$, $R_\TT\coloneqq\textnormal{diam}_{\|\cdot\|_\infty}(\TT)$, $\cc_1\coloneqq\min\{R_\TT,2B_K/\Lambda_\infty\}$ and $\cc_2\coloneqq\min\{R_\TT,6B_K/(5\Lambda_\infty)\}$,
\begin{align*}
\C_3 \coloneqq \frac{B_K\Lambda_\infty C_\TT\,\cc_2^{1-\gamma}\log(e+\cc_2^{-1})}{(1-\gamma)^2},\qquad \C_4 \coloneqq B_K\Lambda_\infty\sqrt{C_\TT}\,\cc_1^{1-\frac\gamma2}\sqrt{\log(e+\cc_1^{-1})}.
\end{align*}
\end{theorem}

\begin{remark}\label{n remark}
The additional embedding approximation term in the bound for $\Th^{m,n}$ in~\Cref{thm: main error bound 2} does not deteriorate the rate obtained for $\Th^m$ as long as the number of reference samples $n$ grows suitably with $m$. Indeed, the term $\C_3/m$ is of lower order in $m$ than the sampling term. Moreover, since $\frac{2}{2+\gamma}\leq1$, the sampling term is of order $(\log(m)/m)^{\frac{2}{2+\gamma}}$. Thus, the remaining term, of order $1/\sqrt{mn}$, is negligible provided
\begin{align*}
\frac{1}{\sqrt{mn}}\lesssim\left(\frac{\log(m)}{m}\right)^{\frac{2}{2+\gamma}}\quad\Longleftrightarrow\quad n\gtrsim\frac{m^{\frac{2-\gamma}{2+\gamma}}}{\log(m)^{\frac{4}{2+\gamma}}},
\end{align*}
up to constants depending on $\C_1$, $\C_4$, $B_K$ and $t$. Since $\gamma\in[0,1)$, the exponent $\frac{2-\gamma}{2+\gamma}$ lies in $(\frac13,1]$. Taking $n$ proportional to $m$ therefore always suffices to retain the fast rate, and, for $\gamma>0$, a sublinear number of reference samples per training sample is already enough. In particular, the total number $mn$ of evaluations of $\T$ needed to compute $M^n(\nu^m,\T)$ can be kept of order $m^{\frac{4}{2+\gamma}}$, up to logarithmic factors.
\end{remark}

\subsection{High level strategy of \Cref{thm: main error bound 2}}\label{subsec: high-level strategy}

We briefly outline the proof of~\Cref{thm: main error bound 2}. The first step is to lift the minimization of $M(\nu^m,\cdot)$ over $\TT$ to the space of conditional mean embeddings $\G_\TT$, where it becomes a classical least-squares regression problem. Define
\begin{align*}
\phi\big(g,(y,u)\big)\coloneqq\left\|K(u,\cdot)-g(y)\right\|^2_\mathcal{K},\qquad R(g)\coloneqq\int\phi(g,\xi)\,\nu(\dd\xi),\qquad R^m(g)\coloneqq\frac1m\sum_{i=1}^m\phi(g,\xi_i),
\end{align*}
where $\xi_i\coloneqq(y_i,u_i)\overset{\textnormal{i.i.d.}}{\sim}\nu$ are the target training data. With $g^\dag(y)\coloneqq I_K(\nu(\cdot\mid y))$ and $g_\T$ as in~\eqref{eq: short embeddings},~\Cref{lem: app lifted problem} shows that
\begin{align*}
M(\nu,\T)&=R(g_\T),\qquad M(\nu^m,\T)=R^m(g_\T),\\
\DD(\nu,\T)&=\|g_\T-g^\dag\|_2^2=R(g_\T)-R(g^\dag).
\end{align*}
Thus, $\Th^m$ corresponds to an empirical risk minimizer over the class $\mathcal{G}_\TT$, and its excess risk is exactly $\DD(\nu,\Th^m)$.

To obtain fast rates on this regression problem, we apply the \emph{one-shot localization technique} (attributed to van de Geer \cite{van1987new,van2002m}) in the form presented by~\cite[Thm.~A.1]{divol2025optimal-sm}. In order to apply it, we obtain the local fluctuation bound of~\Cref{prop: app local fluctuation} from the Dudley-type bounds of~\Cref{prop: dudley general}. In~\Cref{thm: one-shot localization}, we slightly generalize the one-shot localization theorem to also hold uniformly over approximate minimizers. Doing so, in~\Cref{prop: app oracle inequality} we obtain that, with probability at least $1-e^{-t}$, every $\T^*\in\TT$ such that
\begin{align*}
M(\nu^m,\T^*)\leq\inf_{\T\in\TT}M(\nu^m,\T)+\varrho
\end{align*}
satisfies
\begin{align*}
\DD(\nu,\T^*)\leq c\left(\E^2_{\text{Bias}}+\varrho+(1+t)^2\left(\left(\frac{\C_1\log(e+m)}{m}\right)^{\frac{2}{2+\gamma}}+\frac{\C_2\log(e+m)}{m}\right)\right).
\end{align*}
Taking $\T^*=\Th^m$ and $\varrho=0$ gives the bound for $\Th^m$.

For $\Th^{m,n}$, the integrals defining $g_\T$ are approximated with the reference samples from $\eta_\V$. This introduces the reference-sampling error
\begin{align*}
\E_n\coloneqq\sup_{\T\in\TT}\left|\delta_n(\T)-\delta_n(\Th)\right|,\qquad\delta_n(\T)\coloneqq\M^n(\nu^m,\T)-\M(\nu^m,\T).
\end{align*}
Since $\Th^{m,n}$ minimizes $\M^n(\nu^m,\cdot)$, it is an approximate minimizer of $\M(\nu^m,\cdot)$ with tolerance $2\E_n$ (see~\eqref{eq: decoupling errors}). Using the decomposition of~\Cref{lem: app reference decomposition}, $\E_n$ is the supremum of a centered empirical process whose summands have variance is of order $1/n$. Thus, \Cref{prop: app reference error} bounds $2\E_n$ by a deterministic $\varrho$ with high probability by combining the Dudley-type bound of~\Cref{prop: dudley general} with the concentration inequality of~\Cref{lem: bousquet}. By union bound over the two high-probability events, we then show that, with probability at least $1-2e^{-t}$,
\begin{align*}
\M(\nu^m,\Th^{m,n})&\leq\inf_{\T\in\TT}\M(\nu^m,\T)+\varrho,\qquad\varrho\coloneqq c\left(\frac{\C_4+B_K^2\sqrt t}{\sqrt{mn}}+\frac{\C_3+B_K^2t}{m}\right)\geq 2\E_n,\\
\DD(\nu,\Th^{m,n})&\leq c\left(\E^2_{\text{Bias}}+\varrho+(1+t)^2\left(\left(\frac{\C_1\log(e+m)}{m}\right)^{\frac{2}{2+\gamma}}+\frac{\C_2\log(e+m)}{m}\right)\right).
\end{align*}

The bias term is controlled separately in~\Cref{lem: app bias} to yield
\begin{align*}
\E_{\text{Bias}}\leq L_KC_{1,\eta_\V}\,\textnormal{dist}_{L^\infty(\Y,\HH;\nu_\Y)}\left(\TT,\T^\dagger\right).
\end{align*}

\section{Error analysis of the $\JMMD_G^2$ minimizer}\label{sec: DD vs DG minimization}
This section instead concerns the analysis of the $\DG$ minimization approach towards training triangular transport maps for conditional generation formulated in~\Cref{prob: empirical level joint minimization}. In~\Cref{sec: CD vs JMMD 1} we first introduce the energy representation of the two divergences $\DD$ and $\DG$. Afterwards, in~\Cref{sec: inconsistency} we show that the estimator $\Th^m_G$ introduced in~\eqref{eq: joint minimizer from samples} is inconsistent for the average conditional estimation error in~\eqref{eq: quantity of interest for conditional generation} in the misspecified case (i.e., $\E_{\text{Bias}}>0$). Ultimately, in~\Cref{sec: joint minimization conditional MMD rate} we prove that selecting $G$ among a family of concentrating RBF kernels allows for a quantifiable upper bound of the estimator's misspecified error which however suffers a curse of dimensionality arising from the Gaussian smoothing induced by $G$ in concentrating around its diagonal.

\subsection{Energy representation of $\DD$ and $\DG$}\label{sec: CD vs JMMD 1}

For measures $\rho,\rho'\in\mathcal{P}(\Y\times \U)$, we introduce the conditional kernel mean embedding difference function $\Delta_{\rho,\rho'}: \Y\mapsto\mathcal{K}$, defined by
\begin{align}\label{eq: compact notation CME difference}
\Delta_{\rho,\rho'}(y) \coloneqq I_{K}(\rho(\cdot\mid y))-I_{K}(\rho'(\cdot\mid y))\, .
\end{align}

Furthermore, we define the transport map-parameterized differences between the target and the pushforwards of the reference under the elements of $\TT$ by
\begin{align}\label{eq: compact notation CME difference with map}
\Delta_{\T}(y)\coloneqq \Delta_{\nu,\T_\t\#\eta}(y).
\end{align}

As a preliminary observation, we note that both $\DD(\nu, \T)$ and $\DG(\nu, \T)$ are energies of $\Delta_\T$. They differ only in the operator that defines the energy. To this end, for bounded measurable kernel $G:\Y\times\Y\to\mathbb{R}$, we introduce the integral operator $A_
G: L^2(\Y,\mathcal{K};\rho_\Y)\to L^2(\Y,\mathcal{K}; \rho_\Y)$ defined by 
\begin{align}\label{eq: kernel integral operator in y}
A_G f(y)\coloneqq \int_{\Y} f(y')G(y,y')\rho_\Y(\dd y').
\end{align}

\begin{proposition}[Energy representation]\label{prop: energy rep}
Let $K$ be bounded as in~\Cref{Boundedness kernel} of~\Cref{Assumptions} and let $G:\Y\times\Y\to\mathbb{R}$ be a bounded measurable PSD kernel. Then, for any $\rho,\rho'\in\mathcal{P}(\Y\times \U)$ such that $\rho_\Y = \rho'_\Y$,
\begin{align}\label{eq: energy representaton DD and DG}
\DD(\rho\pipe\rho')=\left\|\Delta_{\rho,\rho'}\right\|_{L^2(\Y,\mathcal{K};\rho_\Y)}^2, \qquad \DG(\rho,\rho')=\left\langle\Delta_{\rho,\rho'}, A_G\Delta_{\rho,\rho'}\right\rangle_{L^2(\Y,\mathcal{K};\rho_\Y)}.
\end{align}
Moreover, $A_G$ is bounded, self-adjoint and positive, and it acts as $A_G^{\mathbb{R}}\otimes \textnormal{Id}_{\mathcal{K}}$, where $A_G^{\mathbb{R}}$ denotes the integral operator~\eqref{eq: kernel integral operator in y} acting on $L^2(\Y,\mathbb{R};\rho_\Y)$, which is self-adjoint, positive and Hilbert-Schmidt (hence compact). If $G$ is integrally strictly positive definite with respect to $\rho_\Y$\footnote{That is, $\iint G(y,y')g(y)g(y')\rho_\Y(\dd y)\rho_\Y(\dd y')>0$ for every nonzero $g\in L^2(\Y,\mathbb{R};\rho_\Y)$. This is the case of translation invariant bounded continuous kernels with spectral measure from the Bochner theorem with full support, e.g., characteristic kernels~\cite[Thm. 9]{sriperumbudur2010hilbert} such as RBF, Matérn, and Laplace.}, then $A_G$ and $A_G^{\mathbb{R}}$ are injective and the eigenvalues of $A_G^{\mathbb{R}}$ satisfy $\lambda_1\geq\lambda_2\geq\dots>0$, with $\lambda_k\to 0$ when $L^2(\Y,\mathbb{R};\rho_\Y)$ is infinite dimensional (e.g., $\rho_\Y$ is not supported on finitely many atoms).
\end{proposition}

Thus, when $G$ is integrally strictly positive definite with respect to $\nu_\Y$, for $\Delta_\T=\sum_k c_k(\T)\phi_k$ with an eigen-basis $\{\phi_k\}$ of $A_G^\mathbb{R}$ (see~\eqref{eq: spectral rep}), 
\begin{align*}
\DD(\nu,\T) = \sum_k\left\|c_k(\T)\right\|^2_\mathcal{K}, \qquad \DG(\nu,\T) = \sum_k \lambda_k\left\|c_k(\T)\right\|^2_\mathcal{K}\qquad \lambda_k \downarrow0.
\end{align*}
From this, we can then observe that $\DG$ amounts to a version of $\DD$ that discounts the tail of the spectrum of $A_G$. Furthermore, it is immediate that for whatever choice of bounded $G$, $\DG$ and $\DD$ cannot be made proportional when $L^2(\Y,\mathbb{R};\nu_\Y)$ is infinite dimensional.

Next, let us introduce the space spanned by the conditional mean embedding differences induced by $\TT$
\begin{align*}
\mathcal{D}\coloneqq \overline{\textnormal{span}}\left\{\Delta_\T \mid \T\in\TT\right\}\subseteq L^2(\Y,\mathcal{K};\nu_\Y),
\end{align*}
and denote by $\lambda_G^{-}$ and $\lambda_G^{+}$, respectively, the infimum and supremum of the Rayleigh quotient $\langle f, A_G f\rangle_{L^2(\Y,\mathcal{K};\nu_\Y)}/\|f\|^2_{L^2(\Y,\mathcal{K};\nu_\Y)}$ over $f\in\mathcal{D}\setminus\{0\}$. Then, we denote the \emph{distortion} of $G$ on the class $\TT$
\begin{align}\label{eq: distorsion}
\kappa_G \coloneqq \frac{\lambda_G^+}{\lambda_G^-}\in[1,\infty],\qquad\text{with }\kappa_G\coloneqq\infty\text{ if }\lambda_G^-=0,
\end{align}
In this setup, we can then derive the following result as a corollary of~\Cref{prop: energy rep}.

\begin{proposition}\label{prop: comparison}
Let $K$ and $G$ be as in~\Cref{prop: energy rep}, with $A_G$ taken with respect to $\nu_\Y$. Then, for any $\T\in\TT$,
\begin{align*}
\lambda_G^-\,\DD(\nu,\T)\leq \DG(\nu,\T)\leq \lambda_G^+\,\DD(\nu,\T),
\end{align*}
and consequently
\begin{align}\label{eq: sandwitch bound}
\E_{\text{Bias}}^2\leq \DD(\nu, \Th_G)\leq \kappa_G\, \E_{\text{Bias}}^2,
\end{align}
with the convention $\kappa_G\,\E_{\text{Bias}}^2\coloneqq+\infty$ whenever $\kappa_G=\infty$. Moreover, if $A_G$ is injective and $\DD(\nu,\T^\star)=0$ for some $\T^\star\in\TT$, then $\DD(\nu,\Th_G)=0$.
\end{proposition}

\begin{remark}\label{rem: distorsion control}
The sandwich bound~\eqref{eq: sandwitch bound} behaves very differently in the well-specified and misspecified regimes.

\emph{Well-specified case.} If $\E_{\text{Bias}}=0$ is attained in $\TT$ and $A_G$ is injective, then $\DD(\nu,\Th_G)=0$, even when $\kappa_G=\infty$ by the last part of~\Cref{prop: comparison}. Hence $\DG$-minimization is consistent for the average conditional estimation error~\eqref{eq: quantity of interest for conditional generation} in the well-specified case.

\emph{Misspecified case.} When $\E_{\text{Bias}}>0$, the upper bound in~\eqref{eq: sandwitch bound} is informative only if $\lambda_G^->0$. For $f\in L^2(\Y,\mathcal{K};\nu_\Y)$, let $c_k(f)\coloneqq\int_\Y f\phi_k\,\dd\nu_\Y\in\mathcal{K}$ be its coefficients along the eigenfunctions of $A_G^{\RR}$, so that $\langle f,A_Gf\rangle_{L^2(\Y,\mathcal{K};\nu_\Y)}=\sum_k\lambda_k\|c_k(f)\|_\mathcal{K}^2$. The Rayleigh quotient is therefore small on elements of $\mathcal{D}$ whose $y$-dependence concentrates on high-index eigenfunctions. Precisely, suppose $\mathcal{D}$ contains unit-norm elements $f_j$ with $c_k(f_j)\to0$ as $j\to\infty$ for every $k$. By Bessel's inequality, for every $k_0$,
\begin{align*}
\left\langle f_j,A_Gf_j\right\rangle_{L^2(\Y,\mathcal{K};\nu_\Y)}\leq\sum_{k\leq k_0}\lambda_k\left\|c_k(f_j)\right\|_\mathcal{K}^2+\lambda_{k_0+1},
\end{align*}
so that $\lambda_G^-=0$ and $\kappa_G=\infty$\footnote{Since $A_G$ is not compact when $\dim\mathcal{K}=\infty$, infinite dimensionality of $\mathcal{D}$ alone does not force $\lambda_G^-=0$. For instance, $\lambda_G^-=\lambda_G^+=\lambda_1$ if every $\Delta_\T$ is of the form $\phi_1 e_\T$ with $e_\T\in\mathcal{K}$.}. This is the typical situation for classes with rich $y$-dependence which oscillate quickly. We discuss this phenomenon in~\Cref{rem: high frequency blindness}.

Conversely, $\kappa_G<\infty$ whenever $\mathcal{D}$ is finite dimensional and $A_G$ is injective. This is the case for models that are affine in their parameters, i.e., $\theta\mapsto I_K(\T_\theta(y,\cdot)\#\eta_\V)$ affine. For nonlinear parametric models, however, $\mathcal{D}$ is the closed span of the whole family $\{\Delta_{\T_\theta}\}_{\theta\in\Theta}$ and is in general infinite dimensional. Moreover, $\kappa_G$ is nondecreasing under enlargement of $\TT$, since $\mathcal{D}$ can only grow.

Even though~\Cref{rem: distorsion control} highlights the limitations of the bound provided by~\Cref{prop: comparison}, we present a bound on the average conditional estimation error of $\Th_G^m$ in the particular case, but not general by any means, where $\lambda_G^->0$. To do so, we assume that a generalization bound at the $\DG$ level holds.
\end{remark}

\begin{assumption}[Generalization bound at the $\DG$ level]\label{assum: generalization of joint MMD estimator}
There exists $\R:\mathbb{Z}_{\geq 0}\times [0,\infty)\to[0,\infty)$, decreasing in the first argument and non-decreasing in the second argument and independent of $\sigma$, such that for any $t>0$, with probability at least $1-e^{-t}$,
\begin{align*}
\sqrt{\DG(\Th^m_{G})}\leq \sqrt{\DG(\Th_{G})} + \R(m,t).
\end{align*}
\end{assumption}
% In the literature, several works have already established the needed conditions for the $\DG$ generalization bound of~\Cref{assum: generalization of joint MMD estimator} to hold. In particular,~\cite{briol2019statistical} shows that in the parametric setting where $\TT = \TT_\Theta$ as in~\eqref{eq: param class} for compact $\Theta$, and $G$ bounded by $B_G$ (besides $K$ by $B_K$) as in~\Cref{Boundedness kernel} in~\Cref{Assumptions},
% \begin{align}\label{eq: parametric class joint MMD rate}
% \R(m,t) = 4\sqrt{\frac{2}{m}\sup_{(y,u)\in\Y\times\U}K\otimes G\left((y,u),(y,u)\right)}\left(2+\sqrt{t}\right) = 4 B_K B_G\sqrt{\frac{2}{m}}(2+\sqrt{t}).
% \end{align}
In the parametric setting where $\TT=\TT_\Theta$ as in~\eqref{eq: param class} for compact $\Theta\subseteq\RR^p$, with $G$ bounded by $B_G$ (besides $K$ by $B_K$) as in~\Cref{Boundedness kernel} in~\Cref{Assumptions},~\Cref{assum: generalization of joint MMD estimator} can be verified as in~\cite[Thm.~1]{briol2019statistical}. The argument only uses the concentration of the target and of the reference empirical measures separately, so their coupling through the $y_j$'s is immaterial, but it requires a bound uniform over $\Theta$: if $\theta\mapsto\T_\theta$ is $L_\TT$-Lipschitz into $L^\infty(\Y,\HH;\nu_\Y)$, and~\Cref{lipshitzness} in~\Cref{Assumptions} and~\Cref{inf embedding} in~\Cref{assum: strengthened assum} hold, a standard covering argument over $\Theta$ (see~\Cref{lem: app joint generalization}) yields~\Cref{assum: generalization of joint MMD estimator} with
\begin{align}\label{eq: parametric class joint MMD rate}
\R(m,t)= cB_KB_G\sqrt{\frac{1+t+p\log(e+L_\Theta m)}{m}},\qquad L_\Theta\coloneqq\frac{\textnormal{diam}(\Theta)L_KC_{\infty,\eta_\V}L_\TT}{B_K},
\end{align}
for a numerical constant $c>0$.

\begin{proposition}[Spectral bound for $\Th_G^m$]\label{prop: spectral bound joint estimator}
Let $K$ and $G$ be as in~\Cref{prop: energy rep}, and suppose that~\Cref{assum: generalization of joint MMD estimator} holds. If $\lambda_G^->0$, then for every $t>0$, with probability at least $1-e^{-t}$,
\begin{align*}
\DD\left(\nu,\Th^m_G\right)\leq\left(\sqrt{\kappa_G}\,\E_{\text{Bias}}+\frac{\R(m,t)}{\sqrt{\lambda_G^-}}\right)^2.
\end{align*}
In particular, in the well-specified case, $\DD(\nu,\Th^m_G)\leq\R(m,t)^2/\lambda_G^-$, which matches the assumed $\DG$ rate.
\end{proposition}

However, when $\lambda_G^- = 0$ as in most interesting cases,~\eqref{eq: sandwitch bound} gives no quantitative upper bound of $\DD(\nu,\Th_G)$ and lower bound beyond $\E^2_{\text{Bias}}$ in the misspecified regime. This motivates the strict lower bound of~\Cref{sec: inconsistency} and the alternative upper bound of~\Cref{sec: joint minimization conditional MMD rate}.

\subsection{Inconsistency of $\DG$ minimizer for the average conditional estimation error}\label{sec: inconsistency}
Next, we show that the left inequality of~\eqref{eq: sandwitch bound} is strict under generic conditions, therefore making $\DG$-minimization based estimators inconsistent for~\eqref{eq: quantity of interest for conditional generation}. We consider the parametric setting where
\begin{align}\label{eq: param class}
\TT =\TT_\Theta\coloneqq \left\{\T_\theta = \T(\theta) \mid \theta\in\Theta  \right\},
\end{align}
and $\Theta\subseteq \mathbb{R}^p$ is compact. Doing so, we then let 
\begin{align*}
\widehat\theta:\in \SS\coloneqq \argmin_{\theta\in\Theta} \DD(\nu, \T_\theta),\qquad \widehat\theta_G:\in \SS_G\coloneqq\argmin_{\theta\in\Theta} \DG(\nu, \T_\theta), 
\end{align*} 
and similarly $\widehat\theta_G^m:\in \SS_G^m\coloneqq\argmin_{\theta\in\Theta} \DG^{m}(\nu^m, \T_\theta)$. 
Moreover, we denote, for any $f\in L^2(\Y,\mathcal{K}; \nu_\Y)$ 
\begin{align}\label{eq: DG DD expression for general h}
Q_{f}(\theta)\coloneqq \left\|f-\mu_\theta\right\|_{L^2(\Y,\mathcal{K}; \nu_\Y)}^2,\qquad Q_{G,f}(\theta)\coloneq \langle f-\mu_\theta, A_G(f-\mu_\theta)\rangle_{L^2(\Y,\mathcal{K}; \nu_\Y)},
\end{align}
where we compactly define $\mu_\theta(y) \coloneqq I_K(\T_\theta(y,\cdot)\#\eta_\V)$. Note that when $f(y) = I_K(\nu(\cdot\mid y))$ we recover $Q_f(\theta) = \DD(\nu, \T_\theta)$ and $Q_{G,f}(\theta) = \DG(\nu, \T_\theta)$. We also let $E_\theta \coloneqq \textnormal{span}\{\partial_1 \mu_\theta, ..., \partial_p \mu_\theta\}$ be the model's tangent space at $\theta\in\Theta$ and $J_\theta \mu_\theta$ be the Jacobian of $\theta\to \mu_\theta$. 

We then set the following conditions yielding the inconsistency result.

\begin{assumption}[Inconsistency conditions]\label{assum: inconsistency assumptions}
It holds that 
\begin{enumerate}
\item \label{regularity} \emph{(Regularity)} $\theta\to \mu_\theta$ is $C^2$ on an open set $\Omega\supset \Theta$ and $J_\theta \mu_\theta$ is injective for every $\theta\in\Theta$.
\item \label{interiority} \emph{(Interiority)} The $\DD$ minimizers are interior, i.e., $\SS\subset \text{int} \Theta$.
\item \label{no Invariance} \emph{(No invariance)} For every $\theta\in\Theta$, $E_\theta$ is not $A_G$-invariant.
\end{enumerate}
\end{assumption}

\begin{remark}
The conditions of~\Cref{assum: inconsistency assumptions} are fairly general.
\Cref{regularity} holds for smooth parametrizations without redundant parameters. Moreover, $\theta\mapsto\mu_\theta$ continuous with locally Lipschitz derivatives would suffice.
\Cref{interiority} is standard in the analysis of misspecified M-estimators, and can be ensured by taking $\Theta$ large enough.
\Cref{no Invariance} fails only in exceptional situations. By~\Cref{prop: energy rep}, an $A_G$-invariant finite-dimensional subspace is spanned by eigenfunctions of $A_G^{\RR}$ with $\mathcal{K}$-valued coefficients, so the tangent spaces $E_\theta$ would have to be built from these specific eigenfunctions of the kernel integral operator.
\end{remark}

Next, we state the inconsistency result. To do so, we introduce
\begin{align}\label{eq: consistency targets}
\mathcal{B} \coloneqq \left\{f\in L^2(\Y,\mathcal{K};\nu_\Y) \mid \exists \theta\in \Theta \textnormal{ s.t. } \nabla Q_f(\theta) = 0 = \nabla Q_{G,f}(\theta)\right\},
\end{align}
which contains the set of potential target conditional mean embeddings for which $\SS\cap \SS_G \neq \varnothing$.

\begin{theorem}[Inconsistency of $\DG$ minimizers]\label{thm: inconsistency}
Suppose $K$ and $G$ are as in~\Cref{prop: energy rep} with $A_G$ taken with respect to $\nu_\Y$. Under~\Cref{assum: inconsistency assumptions}, it follows that $\mathcal{B}$ is closed, nowhere dense and Haar-null\footnote{We call a Borel set $\mathcal{A}\subseteq L^2(\Y,\mathcal{K};\nu_\Y)$ \emph{Haar-null} if there exists a Borel probability measure $\mu$ on $L^2(\Y,\mathcal{K};\nu_\Y)$ such that $\mu(\mathcal{A}+x)=0$ for every $x\in L^2(\Y,\mathcal{K};\nu_\Y)$. In finite dimensions, this notion coincides with being Lebesgue-null.}. Moreover, when $y\to I_K(\nu(\cdot\mid y))\notin\mathcal{B}$ it holds that:
\begin{enumerate}
\item No $\DG$ minimizer is a $\DD$ minimizer and vice versa, i.e., $\SS\cap\SS_G = \varnothing$.
\item The gap $\mathfrak{d}_G\coloneqq \min_{\theta_G\in\SS_G}\DD\left(\nu, \T_{\theta_G}\right) - \E_{\text{Bias}}^2>0$.
\item Consequently, $\DD\left(\nu,\T_{\theta_G}\right)\geq \E_{\text{Bias}}^2 + \mathfrak{d}_{G}$ for every $\theta_G\in\SS_G$.
\item Whenever $\sup_{\theta\in\Theta}\left|\DG^{m}\left(\nu^m, \T_\theta\right)- \DG\left(\nu, \T_\theta\right)\right|\to 0$ a.s., every limit point of $\widehat \theta_G^m$ is in $\SS_G$ and $\liminf_{m\to \infty}\DD\left(\nu, \T_{\widehat \theta_G^m}\right)\geq \E_{\text{Bias}}^2 + \mathfrak{d}_G$.
\end{enumerate}
\end{theorem}

\begin{remark}[Other empirical approximations of $\DG$]\label{rem: other joint estimators}
Replacing the within-sample sums in~\eqref{eq: empirical joint MMD loss} by their off-diagonal counterparts normalized by $1/(m(m-1))$, either only in the first and third terms as in~\cite{briol2019statistical} or also in the cross term (which gives an unbiased estimator of $\DG(\nu,\T)$), changes the objective by at most $4B_K^2B_G^2/m$, respectively $8B_K^2B_G^2/m$, uniformly in $\T\in\TT$ (cf.~\cite[Lem.~2]{briol2019statistical}). The corresponding minimizers are thus $\mathcal{O}(B_K^2B_G^2/m)$-approximate minimizers of $\DG^m(\nu^m,\cdot)$, so that~\Cref{assum: generalization of joint MMD estimator} holds with $\R(m,t)$ in~\eqref{eq: parametric class joint MMD rate} increased by at most $4B_KB_G/\sqrt m$, and the uniform convergence condition in~\Cref{thm: inconsistency} holds for one of these objectives if and only if it holds for the others. In particular, since $\mathfrak{d}_G$ is defined through population minimizers only, the inconsistency in~\Cref{thm: inconsistency} does not depend on whether the empirical objective is biased or unbiased.
\end{remark}

The inconsistency result shown in the parametric setting refines under generic conditions the lower bound of~\eqref{eq: sandwitch bound}. In the next section, for a suitable choice of $G$, we derive an alternative to the upper bound of~\eqref{eq: sandwitch bound} which is more informative to the goal of bounding the average conditional estimation error of $\Th^m_G$. 

\subsection{Average conditional estimation error bound of $\Th_{\Gsi}^m$ under Gaussian smoothing}\label{sec: joint minimization conditional MMD rate}
After having established that $\DG$ distorts $\DD$ via the additional kernel $G$ resulting in inconsistent estimators under generic conditions, we pivot our interest on how small this distortion can be made, and at what cost. For this part of the analysis, for simplicity, we consider the setting where $\Y$ is a Borel set of $\mathbb{R}^{d_\Y}$. We then select $G$ among the concentrating family of RBF kernels
\begin{align}\label{eq: RBF kernels family}
\Gsi(y,y')\coloneqq \exp(-\frac{\|y-y'\|^2}{2\sigma^2}),\qquad \Gsil\coloneqq c_\sigma \Gsi,\qquad c_\sigma\coloneqq \left(2\pi\sigma^2\right)^{-d_\Y/2}\, . 
\end{align}
We note that $\Gsil(y-y') = \phi_\sigma(y-y')$, where $\phi_\sigma$ denotes the density of the Gaussian $N(0, \sigma^2 I_{d_\Y})$. Moreover, since the objectives $\DGsi$ and $\DGsil$ differ only by a constant, their population minimizers $\Th_{\Gsi}$ and $\Th_{\Gsil}$ coincide.
Next, for probability distributions $\rho, \rho'\in\mathcal{P}(\Y\times \U)$ with $\rho_\Y = \rho'_\Y$ absolutely continuous with density $p_{\rho_\Y}$, we let  
\begin{align}\label{eq: product density}
f_{\rho,\rho'}\coloneqq \Delta_{\rho,\rho'}p_{\rho_{\Y}},\qquad f_{\T}\coloneqq f_{\nu, \T_\t\#\eta},
\end{align}
and we introduce the $p_{\rho_\Y}$-reweighted version of $\DD(\rho \pipe \rho')$, which we define as
\begin{align}\label{eq: reweighted CD}
\DDR(\rho \pipe \rho') \coloneqq \int_\Y\left\|\Delta_{\rho,\rho'}(y)\right\|_{\mathcal{K}}^2 p_{\rho_\Y}(y)\rho_\Y(\dd y) = \|f_{\rho,\rho'}\|_{L^2(\mathbb{R}^{d_\Y},\mathcal{K};\dd y)}^2,
\end{align} 
where $\dd y$ denotes the Lebesgue measure on $\mathbb{R}^{d_\Y}$. In the case $\Y$ is not all of $\mathbb{R}^{d_\Y}$, we extend any function $f$ defined on $\Y$ to be $0$ outside $\Y$.
We then relate $\DGsil$ and $\DDR$ via Fourier representations of $\DGsi$ and $\DGsil$.
\begin{lemma}[Fourier representation of $\DGsi$ and $\DGsil$]\label{lem: DDR vs DG}
Let $K$ bounded by $B_K$ and $\rho,\rho'\in\mathcal{P}(\Y\times\U)$ with $\rho_\Y = \rho'_\Y$, $\|p_{\rho_\Y}\|_\infty<\infty$. Then, the following hold:
\begin{enumerate}
\item $f_{\rho,\rho'}\in L^1\cap L^2(\mathbb{R}^{d_\Y},\mathcal{K}; \dd y)$ and $\left\|f_{\rho,\rho'}\right\|_{L^2(\mathbb{R}^{d_\Y},\mathcal{K};\dd y)}^2 = \DDR(\rho\pipe \rho')\leq 4B_K^2\|p_{\rho_\Y}\|_\infty$.

\item We can represent $\DDR(\rho\pipe\rho')=\int\|\widehat f_{\rho,\rho'}(\xi)\|_{\mathcal{K}}^2\dd\xi$ and, for any $\sigma>0$,
\begin{align}\label{eq: Fourier rep}
\DGsil\left(\rho,\rho'\right) = \langle f_{\rho,\rho'},\phi_{\sigma}*f_{\rho,\rho'}\rangle_{L^2(\mathbb{R}^{d_\Y},\mathcal{K}; \dd y)} = \int e^{-\sigma^2\|\xi\|^2/2}\left\|\widehat f_{\rho,\rho'}(\xi)\right\|_{\mathcal{K}}^2\dd\xi,
\end{align}
where $\widehat f$ denotes the unitary Fourier transform of $f$.
\item 
$0\leq \DGsil(\rho,\rho')\leq \DDR(\rho\pipe \rho')$ and $\DGsil(\rho,\rho')\uparrow\DDR(\rho\pipe\rho')$ as $\sigma\downarrow 0$.
\item The following increment identity holds:
\begin{align}\label{eq: increment identity}
\DDR(\rho\pipe\rho')-\DGsil(\rho,\rho') = \frac{1}{2}\int \phi_\sigma(z)\left\|f_{\rho,\rho'}(\cdot+z)-f_{\rho,\rho'}(\cdot)\right\|_{L^2(\mathbb{R}^{d_\Y},\mathcal{K};\dd y)}^2\dd z.
\end{align}
\end{enumerate}
All the above statements then hold when $\rho = \nu$, $\rho' = \T_\t\#\eta$ for any $\T\in \TT$ replacing $f_{\rho,\rho'}$ with $f_\T$. Moreover, since $\DGsil = c_\sigma \DGsi$, all the above statements also transfer to $\DGsi$ up to the multiplicative factor $c_\sigma$.
\end{lemma}

\begin{remark}[Generalization to translation invariant $G$]\label{rem: generalization to translation invariant}
We note that~\Cref{lem: DDR vs DG} can be extended beyond RBF kernels to the class of translation invariant kernels, which includes Matérn and Laplace types. Indeed, for $G(y,y')=\psi(y-y')$ with $\psi:\RR^{d_\Y}\to\RR$ continuous and positive definite, by Bochner's theorem (see~\cite[Thm. 3]{sriperumbudur2010hilbert}), $\psi(x)=\int e^{i\omega\cdot x}\Lambda(\dd\omega)$ for a finite nonnegative spectral measure $\Lambda$ on $\RR^{d_\Y}$. Thus, employing the same proof of~\Cref{lem: DDR vs DG},
\begin{align*}
\DG(\rho,\rho')=(2\pi)^{d_\Y}\int\left\|\widehat f_{\rho,\rho'}(\omega)\right\|^2_\mathcal{K}\Lambda(\dd\omega),
\end{align*}
and all the analogues of the conclusions of~\Cref{lem: DDR vs DG} can be derived with the appropriate book-keeping provided that the normalized spectral density is radially nonincreasing. In the specific RBF setting, $\Lambda$ is the Gaussian measure.
\end{remark}
\begin{remark}[Blindness of $\DGsi$ to fast variations in $y$]\label{rem: high frequency blindness}
By~\Cref{lem: DDR vs DG}, in Fourier variables $\DGsil$ and $\DDR$ differ only by the multiplier $e^{-\sigma^2\|\xi\|^2/2}$. Frequencies $\|\xi\|\ll 1/\sigma$ are weighted almost fully, whereas frequencies $\|\xi\|\gg 1/\sigma$ are exponentially discounted. Hence, the components of the reweighted discrepancy $f_{\rho,\rho'}=p_{\rho_\Y}\Delta_{\rho,\rho'}$ that vary in $y$ on scales much smaller than $\sigma$ are essentially invisible to $\DGsil$, while $\DDR$ accounts for them in full. For instance, if $f_{\rho,\rho'}(y)=a(y)\cos(\omega\cdot y)$ for a smooth $\mathcal{K}$-valued envelope $a$ and $\|\omega\|\gg1/\sigma$, then, up to the bandwidth of $a$, $\DGsil(\rho,\rho')$ is smaller than $\DDR(\rho\pipe\rho')$ by a factor of order $e^{-\sigma^2\|\omega\|^2/2}$. Two transport maps whose conditional mean embeddings differ by such a rapidly varying function of $y$ are therefore nearly indistinguishable by $\DGsil$, while being far apart in $\DDR$, and hence in $\DD$, since $\DDR\leq\|p_{\rho_\Y}\|_\infty\DD$. This is the mechanism behind~\Cref{ex:counter-CMMD-counter-example}: there, $\Delta_{\rho,\rho'}(y)=K(\alpha y,\cdot)-K(-\alpha y,\cdot)$ varies on the scale $1/\alpha$, which is much smaller than the unit length-scale of $G$ when $\alpha$ is large. It is also the mechanism behind $\lambda_G^-=0$ in~\Cref{rem: distorsion control}. By~\Cref{rem: generalization to translation invariant}, the same picture holds for any bounded continuous translation-invariant $G$, with $e^{-\sigma^2\|\xi\|^2/2}$ replaced by its spectral measure, which is finite and thus always discounts sufficiently high frequencies.
\end{remark}

\begin{remark}[A dynamic counterpart: conditional design of triangular flows]\label{rem: dynamic transport}
The contrast between $\CD$ and $\JMMD_G^2$ minimization has a natural counterpart in dynamic measure transport. In conditional flow matching, triangular velocity fields are typically constructed so that the conditioning variable stays fixed along the flow~\cite{kerrigan2024dynamic}. For conditioning, however, it only matters that the $y$-component of each trajectory returns to the observed value at the final time, and that trajectories started from different values of $y$ never merge. Paths that move away from $y$ and come back, as in~\cite{johnsonlocalizing}, as well as interpolation schedules depending on $y$, therefore yield equally valid conditional samplers.

Error bounds for flow matching amplify the velocity error exponentially in the time-integrated Lipschitz constant of the velocity field~\cite{benton2023error}. Adapting their stability argument to each fixed value of $y$, for a prescribed motion in $y$, bounds the average conditional Wasserstein error (i.e., $\DD$ with $D_{\text{Cond.}}$ being the Wasserstein distance) by the local training errors amplified by $y$-dependent time-integrated Lipschitz constants, averaged over $\nu_\Y$. Controlling a single joint Lipschitz constant instead amplifies the training error by the worst case over $y$, which is the dynamic analogue of targeting the joint error. Since training relies on joint samples, the local training errors can only be controlled on average over $\nu_\Y$, as in the static setting. The conditional bound then indicates which weighted average of them should be controlled, suggesting $y$-dependent weights in the training loss, and it improves on the worst case precisely when the conditionals are irregular only on regions of small probability. The same argument also clarifies the roles of the design choices. The motion in $y$ leaves the dynamics of each conditional unchanged and acts on the regularity of the velocity in the conditioning variable, hence on how accurately a single model can approximate it. Schedules adapted to $y$ act on the local constants and on the local training errors, by homogenizing the velocity across conditionals, consistently with the experimental findings of~\cite{johnsonlocalizing} on a two-dimensional funnel distribution with sharply varying conditionals. Designing paths, schedules, and training losses to control these averaged conditional quantities, rather than joint ones, would then be the principled choice for conditional generation, in the same spirit as minimizing $\CD$ rather than $\JMMD_G^2$ in the static setting. Developing this conditional error analysis for dynamic transport, its implications for the design of conditional flow matching schemes efficiently implementable from joint data samples, and the relations with respect to this work's analysis is a very appealing research direction we plan to pursue.
\end{remark}

\begin{remark}[Limitations of $y$-dependent renormalizations to target $\DD$ directly]\label{rem: renormalized kernel}
As the length-scale vanishes, the joint criterion $\DGsil$ built from the Gaussian kernel targets $\DDR$, a version of the average conditional estimation error that is reweighted by the density of the conditioning variable. In this case, regions where this density is large weigh more. One could remove this reweighting by renormalizing the kernel with respect to the marginal of the conditioning variable. Dividing by the integral of the kernel in only one of its arguments breaks symmetry and does not yield a kernel. Its symmetric counterpart
\begin{align*}
\Gsir(y,y')\coloneqq\frac{\Gsi(y,y')}{\sqrt{\int_\Y\Gsi(y,y'')\,\nu_\Y(\dd y'')\int_\Y\Gsi(y',y'')\,\nu_\Y(\dd y'')}},
\end{align*}
is instead a positive semidefinite kernel, and an argument analogous to the proof of~\Cref{lem: DDR vs DG} suggests that the associated joint criterion targets the average conditional estimation error $\DD$ itself as the length-scale vanishes. However, this kernel depends on the unknown marginal of the conditioning variable through its normalization. In practice, the normalization must be estimated from samples. This makes the criterion data-dependent, introduces an additional estimation error which our analysis does not cover, and gives estimators not very amenable to implementation.
\end{remark}

We therefore maintain our focus on the implementable choice $G=\Gsi$, and its equivalent $\Gsil$ up to the factor $c_\sigma$, for which the joint criterion targets $\DDR$ rather than $\DD$ by~\Cref{lem: DDR vs DG}. In this case, the procedure to obtain an error bound on $\Th_{\Gsi}^m$ then relies on the two main goals:
\begin{enumerate}
\item \label{conversion}\emph{(Conversion gap)} Relating the terms $\DD(\nu, \T)$ and $\DDR(\nu, \T)$ for any $\T\in\TT$.
\item \label{smoothing}\emph{(Smoothing gap)} Controlling, as $\sigma\downarrow 0$, the uniform smoothing gap
\begin{align}\label{eq: uniform smoothing gap}
\Xi_\sigma\coloneqq \sup_{\T\in\TT}\left(\DDR(\nu,\T)-\DGsil(\nu,\T)\right).
\end{align}
\end{enumerate}
To control the conversion and smoothing gap, we make the following assumptions. In presenting them, we let $p_{\nu_\Y}^{\min}\coloneqq \essinf_{y\in\Y}p_{\nu_\Y}(y)$ and $\m(\lambda)\coloneqq \nu_\Y\left(\left\{ y \mid 0< p_{\nu_\Y}(y)<\lambda \right\}\right)$.

\begin{assumption}[Conversion and smoothing gap conditions]\label{assum: smoothing and conversion gap} It holds that:
\begin{enumerate}
\item \label{margin condition} \emph{($p_{\nu_\Y}$ margin condition)} There exist $\beta\in (0,\infty]$ and $C_\beta \geq 1$ satisfying
\begin{align*}
\m(\lambda)\leq 
C_\beta \lambda^\beta,\qquad \forall \lambda>0,
\end{align*}
where $\beta = \infty$ means $p_{\nu_\Y}^{\min}>0$ (so that $\m(\lambda) = 0$ for $\lambda\leq p_{\nu_\Y}^{\min}$) and we set
\begin{align}\label{eq: beta constants}
\cc_\beta\coloneqq\frac{\beta}{1+\beta}\in(0,1],\qquad \C_\beta\coloneqq \begin{cases}\left(4B_K^2 C_\beta\right)^{\frac{1}{1+\beta}},\quad &\text{if }\beta<\infty,\\
\frac{1}{2}\left(p_{\nu_\Y}^{\min}\right)^{-1}, & \text{if }\beta=\infty.
\end{cases}
\end{align}

\item \label{L2-modulus assumption}\emph{($L^2$-modulus of the weighted discrepancy)} There exists $\zeta\in(0,1]$\footnote{$\zeta>1$ automatically forces $f_\T$ to be the $0$ constant function.} such that 
\begin{align}\label{eq: L2-modulus eq}
0<\footnotemark\mathcal{M}_\zeta\coloneqq \sup_{\T\in\TT} \sup_{z\neq 0}\frac{\left\|f_\T(\cdot+z)-f_\T(\cdot)\right\|_{L^2(\mathbb{R}^{d_\Y},\mathcal{K};\dd y)}}{\|z\|^\zeta}<\infty.
\end{align}
\end{enumerate}
\end{assumption}
\footnotetext{If $\mathcal{M}_\zeta=0$, then every $f_\T$ is invariant under translations and hence vanishes, i.e., every $\T\in\TT$ satisfies $\DD(\nu,\T)=0$.}

\begin{remark}
For bounded convex set $\Y\subset \mathbb{R}^{d_\Y}$\footnote{Working on bounded convex $\Y$ is just a simplification for this remark. However, the remark can be extended to unbounded domains under appropriate tail assumptions on $\nu_\Y$.} assuming the uniform H\"older condition 
\begin{align}\label{eq: Holder cond}
\sup_{\T\in\TT}\|\Delta_\T(y)-\Delta_\T(y')\|_\mathcal{K}\leq L_{\Delta}\|y-y'\|^\alpha,\qquad \forall y,y'\in\Y,
\end{align}
and the complementary $|p_{\nu_\Y}(y)-p_{\nu_\Y}(y')|\leq L_{p_{\nu_\Y}}\|y-y'\|^\alpha$ holding for any $y,y'\in\Y$ if $p_{\nu_\Y}^{\min}>0$ (or otherwise over all of $\mathbb{R}^{d_\Y}$ with $p_{\nu_\Y}$ extended to $0$ outside $\Y$), by the uniform upper bound for any $\T\in\TT$
\begin{align*}
\left\|f_\T(y') - f_\T(y)\right\|_\mathcal{K}\leq \left\|\Delta_\T(y)-\Delta_\T(y')\right\|_\mathcal{K}p(y)+\left\|\Delta_\T(y')\right\|_{\mathcal{K}}\left|p(y)-p(y')\right|\leq L_{f} \left\|y-y'\right\|^{\alpha},
\end{align*}
with $L_{f}\coloneqq L_{\Delta}\|p_{\nu_\Y}\|_\infty+2B_K L_{p_{\nu_\Y}}$,~\Cref{L2-modulus assumption} in~\Cref{assum: smoothing and conversion gap} is satisfied with $\zeta = 1/2\wedge \alpha$ if $p_{\nu_\Y}^{\min}>0$ and with $\zeta=\alpha$ otherwise. The condition of~\eqref{eq: Holder cond} can be verified for Lipschitz $K$ when $\T^\dag$ and all $\T\in\TT$ are H\"older regular in $y$ uniformly using the stability of $\MMD_K$ with Lipschitz $K$ (see~\cite[Thm. 3.2]{baptista2023approximation}). It can also be verified when (in place of $\T^\dag$) $\nu$ admits a H\"older regular likelihood potential $\Phi^\dag: \Y\times\U\to \mathbb{R}$ and prior probability measure $\mu\in\PP(\U)$ such that
\begin{align*}
\frac{\nu(\dd u\mid y)}{\mu(\dd u)} = \frac{1}{Z(y)}\exp\left(-\Phi^\dag(y,u)\right),\qquad Z(y)\coloneqq \int \exp\left(-\Phi^\dag(y,u)\right)\mu(\dd u).
\end{align*}
In this alternative scenario, the proof follows by coupling the stability of $\MMD_K$ and a slight variation\footnote{In~\cite{garbuno2023bayesian}, they consider integral probability metrics (IPM) defined via Lipschitz balls, which do not include the IPM $\MMD_K$. However, the same arguments go through also for $\MMD_K$ when $K$ is bounded.} of a result on Bayesian (data) perturbation analysis for integral probability metrics in~\cite{garbuno2023bayesian}
\end{remark}

In~\Cref{lem: conversion gap}, we show that~\Cref{margin condition} yields
\begin{align*}
\DD(\nu, \T)\leq 2\C_\beta \DDR(\nu, \T)^{\cc_\beta}, \ \forall \T\in\TT,\quad\Longrightarrow \quad \DD(\nu, \Th_{\Gsi}) \leq 2\C_\beta\left(\|p_{\nu_\Y}\|_\infty\E_{\text{Bias}}^2 + \Xi_\sigma\right)^{\cc_\beta},
\end{align*}
resolving the conversion gap between $\DD$ and $\DDR$ and providing a more informative upper bound compared to~\eqref{eq: sandwitch bound}. Indeed, in~\Cref{lem: smoothing gap}, we show that the uniform smoothing gap in~\eqref{eq: uniform smoothing gap} can be bounded by
\begin{align}\label{eq: smoothing gap bound and constant}
\Xi_\sigma \leq \C_{\Xi}\sigma^{2\zeta},\qquad \C_{\Xi}\coloneqq \frac{1}{2}d_\Y^\zeta \mathcal{M}_\zeta^2.
\end{align}

Using the assumptions and results outlined in this section so far, we are now ready to derive the average conditional estimation error of $\Th^m_{\Gsi}$, which we recall is not a direct minimizer of~\eqref{eq: quantity of interest for conditional generation}.

\begin{theorem}[Error bound of $\Th^m_{\Gsi}$]\label{Error bound of joint minimizer}
Let $K$ be bounded as in~\Cref{Boundedness kernel} of~\Cref{Assumptions}, let $\nu_\Y$ have Lebesgue density $p_{\nu_\Y}$ with $\|p_{\nu_\Y}\|_\infty<\infty$, suppose~\Cref{assum: smoothing and conversion gap}, and assume~\Cref{assum: generalization of joint MMD estimator} holds for $G_\sigma$. Define
\begin{align*}
\widetilde\C_1\coloneqq 2(2\pi)^{-d_\Y/4}\sqrt{\|p_{\nu_\Y}\|_\infty},\qquad \widetilde\C_2\coloneqq(2\pi)^{-d_\Y/2},
\end{align*}
and, for $\sigma>0$, $m\in\mathbb{N}$ and $t>0$,
\begin{align}\label{eq: master rho}
\E_\sigma(m,t)\coloneqq \|p_{\nu_\Y}\|_\infty\E_{\text{Bias}}^2+\C_\Xi\sigma^{2\zeta}+\widetilde\C_1\E_{\text{Bias}}\R(m,t)\,\sigma^{-d_\Y/2}+\widetilde\C_2\R(m,t)^2\sigma^{-d_\Y},
\end{align}
where $\cc_\beta,\C_\beta$ are as in~\eqref{eq: beta constants} and $\C_\Xi,\zeta$ as in~\eqref{eq: smoothing gap bound and constant}. Then, for every $t>0$, the following hold.
\begin{enumerate}[wide, labelindent=0pt]
\item\label{Fixed length-scale}(Fixed length-scale) For every $\sigma>0$, with probability at least $1-e^{-t}$, $\DD\left(\nu,\Th^m_{\Gsi}\right)\leq 2\C_\beta\,\E_\sigma(m,t)^{\cc_\beta}$ and consequently
\begin{align*}
\DD\left(\nu,\Th^m_{\Gsi}\right)\leq 2\C_\beta\Bigg(\underbrace{\left(\|p_{\nu_\Y}\|_\infty\E_{\text{Bias}}^2\right)^{\cc_\beta}}_{\text{Conversion}}&+\underbrace{\left(\C_\Xi\sigma^{2\zeta}\right)^{\cc_\beta}}_{\text{Smoothing}}\\
&\hspace{-1em}+\underbrace{\left(\widetilde\C_1\E_{\text{Bias}}\R(m,t)\,\sigma^{-d_\Y/2}\right)^{\cc_\beta}}_{\text{Mixed sampling, 1}}+\underbrace{\left(\widetilde\C_2\R(m,t)^2\sigma^{-d_\Y}\right)^{\cc_\beta}}_{\text{Pure sampling, 2}}\Bigg).
\end{align*}

\item\label{Bias-agnostic length-scale}(Bias-agnostic length-scale) Let
\begin{align}\label{eq: sigma 2 and B}
\sigma_2(m,t)\coloneqq\left(\frac{\widetilde\C_2\R(m,t)^2}{\C_\Xi}\right)^{\frac{1}{2\zeta+d_\Y}},\qquad \widetilde \R_2(m,t)\coloneqq\C_\Xi^{\frac{d_\Y}{2\zeta+d_\Y}}\left(\widetilde\C_2\R(m,t)^2\right)^{\frac{2\zeta}{2\zeta+d_\Y}}.
\end{align}
Then, with probability at least $1-e^{-t}$,
\begin{align*} \DD\left(\nu,\Th^m_{G_{\sigma_2(m,t)}}\right)\leq 2\C_\beta\left(\sqrt{\|p_{\nu_\Y}\|_\infty}\,\E_{\text{Bias}}+\sqrt{2\widetilde \R_2(m,t)}\right)^{2\cc_\beta}.
\end{align*}
\item\label{Oracle length-scale}(Oracle length-scale) Let
\begin{align}\label{eq: sigma 1 and A}
\sigma_1(m,t)\coloneqq\left(\frac{\widetilde\C_1\E_{\text{Bias}}\R(m,t)}{\C_\Xi}\right)^{\frac{2}{4\zeta+d_\Y}},\qquad \widetilde \R_1(m,t)\coloneqq\C_\Xi^{\frac{d_\Y}{4\zeta+d_\Y}}\left(\widetilde\C_1\E_{\text{Bias}}\R(m,t)\right)^{\frac{4\zeta}{4\zeta+d_\Y}},
\end{align}
so that $\sigma_1=\widetilde \R_1(m,t)=0$ if $\E_{\text{Bias}}=0$, and set $\sigma_{\textnormal{opt}}(m,t)\coloneqq\sigma_1(m,t)\vee\sigma_2(m,t)$. Then, with probability at least $1-e^{-t}$,
\begin{align*}
\DD\left(\nu,\Th^m_{G_{\sigma_{\textnormal{opt}}(m,t)}}\right)\leq 4\C_\beta\left(\left(\|p_{\nu_\Y}\|_\infty\E_{\text{Bias}}^2\right)^{\cc_\beta}+\widetilde \R_1(m,t)^{\cc_\beta}+\widetilde \R_2(m,t)^{\cc_\beta}\right).
\end{align*}
\end{enumerate}
\end{theorem}

\Cref{Fixed length-scale} holds for every $\E_{\text{Bias}}\geq0$, every $\beta\in(0,\infty]$ and every $\zeta\in(0,1]$. \Cref{Bias-agnostic length-scale} and~\Cref{Oracle length-scale} are obtained from it by choosing $\sigma$: $\sigma_2$ balances the smoothing term against the pure sampling term, while $\sigma_1$ balances it against the mixed sampling term. \Cref{Bias-agnostic length-scale} does not require knowledge of $\E_{\text{Bias}}$ and decomposes into the bias plus an estimation term in root scale. \Cref{Oracle length-scale} requires knowledge of $\E_{\text{Bias}}$ and is optimal over $\sigma$ up to a factor $4$, since the $\sigma$-dependent part of $\E_\sigma(m,t)$ is at least $\max\{\widetilde\R_1(m,t),\widetilde\R_2(m,t)\}$ for every $\sigma>0$. In particular, it is never worse than~\Cref{Bias-agnostic length-scale} up to constants. The difference lies in the mixed sampling term, which is of order $\E_{\text{Bias}}\sqrt{\widetilde\R_2(m,t)}$ in~\Cref{Bias-agnostic length-scale} and $\widetilde\R_1(m,t)$ in~\Cref{Oracle length-scale}. When $\E^2_{\text{Bias}}\lesssim\widetilde\R_2(m,t)$, both are dominated by $\widetilde\R_2(m,t)$. When $\E^2_{\text{Bias}}\gg\widetilde\R_2(m,t)$, e.g., for fixed $\E_{\text{Bias}}>0$ and $m\gtrsim B_K^2B_G^2\left(1+t+p\log(e+L_\Theta m)\right)\E_{\text{Bias}}^{-(2\zeta+d_\Y)/\zeta}$ when $\R(m,t)$ is as in~\eqref{eq: parametric class joint MMD rate}, one has $\widetilde\R_1(m,t)\ll\E_{\text{Bias}}\sqrt{\widetilde\R_2(m,t)}$, and the rate of the mixed term improves from $m^{-\zeta/(2\zeta+d_\Y)}$ to $m^{-2\zeta/(4\zeta+d_\Y)}$ up to logarithmic factors. In the well-specified case $\E_{\text{Bias}}=0$,~\Cref{Bias-agnostic length-scale} and~\Cref{Oracle length-scale} coincide up to constants and give $\DD(\nu,\Th^m_{\Gsi})\lesssim\C_\beta\widetilde \R_2(m,t)^{\cc_\beta}$. When $\R(m,t)$ is as in~\eqref{eq: parametric class joint MMD rate},
\begin{align*}
\widetilde \R_1(m,t)&\approx\left(\E_{\text{Bias}}B_KB_G\sqrt{1+t+p\log(e+L_\Theta m)}\right)^{\frac{4\zeta}{4\zeta+d_\Y}}m^{-\frac{2\zeta}{4\zeta+d_\Y}},\\
\widetilde \R_2(m,t)&\approx\left(B_KB_G\sqrt{1+t+p\log(e+L_\Theta m)}\right)^{\frac{4\zeta}{2\zeta+d_\Y}}m^{-\frac{2\zeta}{2\zeta+d_\Y}},
\end{align*}
up to constants depending only on $\C_\Xi$, $\|p_{\nu_\Y}\|_\infty$, $d_\Y$ and $\zeta$. Differently from the error bound of~\Cref{thm: main error bound 2} for $\Th^m$ and $\Th^{m,n}$, both rates have a $d_\Y$-dependent exponent, even in the well-specified case: joint-MMD rate $\R(m,t)^2$ controls $\DGsi(\nu,\Th^m_{\Gsi})$ but not $\DD(\nu,\Th^m_{\Gsi})$, due to the blindness of the joint criterion to fast variations in $y$ highlighted in~\Cref{rem: high frequency blindness}. We conjecture that this curse of dimensionality is unavoidable and leave a matching lower bound for future research.

\section{Conclusion}\label{sec: conclusions}
In this work, we studied the ability of minimum $\MMD$-based divergence estimators of triangular transport maps to recover conditional measures, as measured by the average conditional estimation error of~\eqref{eq: quantity of interest for conditional generation}. For estimators that directly minimize the conditional criterion $\CD$ through a surrogate loss computable from joint samples, we derived non-asymptotic error bounds that decompose the error into approximation, estimation, and reference sampling contributions, with fast rates in terms of the complexity of the hypothesis class. In contrast, we showed that joint $\JMMD_G^2$ minimization is generically inconsistent for the conditional error in the misspecified setting. Conditional guarantees can be recovered from joint minimization by letting the kernel on the conditioning variable concentrate, but only at the price of rates that deteriorate with the dimension of the conditioning variable.

Several directions remain open. First, the principled approach developed here, based on comparing joint and conditional criteria and on lifting conditional estimation to a regression problem, could be extended to other divergences, model classes, and conditional generative models, thereby providing theoretical guarantees in more general settings. A particularly promising instance is dynamic measure transport, such as conditional flow matching with triangular velocity fields~\cite{kerrigan2024dynamic}, where, as discussed in~\Cref{rem: dynamic transport}, the distinction between joint and conditional errors suggests designing the conditioning paths, schedules, and training losses so as to control conditional, rather than joint, error bounds~\cite{benton2023error,johnsonlocalizing}. Second, many conditional sampling tasks are organized hierarchically, for instance across scales, fidelity levels, or blocks of variables ordered according to a dependence structure, and are naturally addressed by composing conditional triangular maps~\cite{akbari2023causal,calle2026generative, blorstad2026learning, davis2026multifidelity}; extending our error analysis to such hierarchies of conditional transport problems is a natural next step. Third, it would be valuable to complement our analysis with dimension reduction techniques, such as autoencoders and variational autoencoders~\cite{kingma2013auto,tolstikhin2017wasserstein}, conditional autoencoders with triangular decoders~\cite{al2026conditional}, likelihood-informed subspaces~\cite{cui2014likelihood, cui2025subspace, cui2022unified}, and low-dimensional transport representations on function spaces~\cite{baptista2026approximating}, in order to understand how low-dimensional structure in the conditioned and conditioning variables affects the conditional estimation error.

\section*{Use of large language models}
The authors used large language models, particularly Claude Opus 5.5, to aid research, refine arguments, and present this article. The authors take full responsibility for the content and correctness of this article.

\bibliographystyle{siamplain}
\bibliography{references}

@incollection{marzouk2017sampling,
  title         = {Sampling via Measure Transport: An Introduction},
  author        = {Marzouk, Youssef and Moselhy, Tarek and Parno, Matthew and Spantini, Alessio},
  booktitle     = {Handbook of Uncertainty Quantification},
  editor        = {Ghanem, Roger and Higdon, David and Owhadi, Houman},
  publisher     = {Springer},
  address       = {Cham},
  pages         = {785--825},
  year          = {2017},
  doi           = {10.1007/978-3-319-12385-1_23}
}

@article{baptista2024conditional,
  title         = {Conditional Sampling with Monotone {GAN}s: From Generative Models to Likelihood-Free Inference},
  author        = {Baptista, Ricardo and Hosseini, Bamdad and Kovachki, Nikola B. and Marzouk, Youssef M.},
  journal       = {SIAM/ASA Journal on Uncertainty Quantification},
  volume        = {12},
  number        = {3},
  pages         = {868--900},
  year          = {2024},
  doi           = {10.1137/23M1581546}
}

@misc{mirza2014conditional,
  title         = {Conditional Generative Adversarial Nets},
  author        = {Mirza, Mehdi and Osindero, Simon},
  howpublished  = {arXiv preprint},
  year          = {2014},
  eprint        = {1411.1784},
  archiveprefix = {arXiv}
}

@article{orozco2025aspire,
  title         = {{ASPIRE}: Iterative Amortized Posterior Inference for {B}ayesian Inverse Problems},
  author        = {Orozco, Rafael and Siahkoohi, Ali and Louboutin, Mathias and Herrmann, Felix J.},
  journal       = {Inverse Problems},
  volume        = {41},
  number        = {4},
  eid           = {045001},
  year          = {2025},
  doi           = {10.1088/1361-6420/adba3d}
}

@article{yang2020physics,
  title         = {Physics-Informed Generative Adversarial Networks for Stochastic Differential Equations},
  author        = {Yang, Liu and Zhang, Dongkun and Karniadakis, George Em},
  journal       = {SIAM Journal on Scientific Computing},
  volume        = {42},
  number        = {1},
  pages         = {A292--A317},
  year          = {2020},
  doi           = {10.1137/18M1225409}
}

@misc{zhang2025probabilistic,
  title         = {Probabilistic Operator Learning: Generative Modeling and Uncertainty Quantification for Foundation Models of Differential Equations},
  author        = {Zhang, Benjamin J. and Liu, Siting and Osher, Stanley J. and Katsoulakis, Markos A.},
  howpublished  = {arXiv preprint},
  year          = {2025},
  eprint        = {2509.05186},
  archiveprefix = {arXiv}
}

@article{hosseini2023conditional,
  title         = {Conditional Optimal Transport on Function Spaces},
  author        = {Hosseini, Bamdad and Hsu, Alexander W. and Taghvaei, Amirhossein},
  journal       = {SIAM/ASA Journal on Uncertainty Quantification},
  volume        = {13},
  number        = {1},
  pages         = {304--338},
  year          = {2025},
  doi           = {10.1137/23M1618922}
}

@inproceedings{lipman2022,
  title         = {Flow Matching for Generative Modeling},
  author        = {Lipman, Yaron and Chen, Ricky T. Q. and Ben-Hamu, Heli and Nickel, Maximilian and Le, Matthew},
  booktitle     = {The Eleventh International Conference on Learning Representations (ICLR)},
  year          = {2023},
  eprint        = {2210.02747},
  archiveprefix = {arXiv}
}

@article{albergo2025stochastic,
  author  = {Michael Albergo and Nicholas M. Boffi and Eric Vanden-Eijnden},
  title   = {Stochastic Interpolants: A Unifying Framework for Flows and Diffusions},
  journal = {Journal of Machine Learning Research},
  year    = {2025},
  volume  = {26},
  number  = {209},
  pages   = {1--80},
  url     = {http://jmlr.org/papers/v26/23-1605.html}
  }

@article{yang2023diffusion,
  title         = {Diffusion Models: A Comprehensive Survey of Methods and Applications},
  author        = {Yang, Ling and Zhang, Zhilong and Song, Yang and Hong, Shenda and Xu, Runsheng and Zhao, Yue and Zhang, Wentao and Cui, Bin and Yang, Ming-Hsuan},
  journal       = {ACM Computing Surveys},
  volume        = {56},
  number        = {4},
  pages         = {1--39},
  year          = {2024},
  doi           = {10.1145/3626235}
}

@article{wang2025efficient,
  title         = {Efficient Neural Network Approaches for Conditional Optimal Transport with Applications in {B}ayesian Inference},
  author        = {Wang, Zheyu Oliver and Baptista, Ricardo and Marzouk, Youssef and Ruthotto, Lars and Verma, Deepanshu},
  journal       = {SIAM Journal on Scientific Computing},
  volume        = {47},
  number        = {4},
  pages         = {C979--C1005},
  year          = {2025},
  doi           = {10.1137/24M1678659}
}

@inproceedings{baptista2025entropicconditional,
  title         = {Conditional Simulation via Entropic Optimal Transport: Toward Non-Parametric Estimation of Conditional {B}renier Maps},
  author        = {Baptista, Ricardo and Pooladian, Aram-Alexandre and Brennan, Michael and Marzouk, Youssef and Niles-Weed, Jonathan},
  booktitle     = {Proceedings of the 28th International Conference on Artificial Intelligence and Statistics (AISTATS)},
  series        = {Proceedings of Machine Learning Research},
  volume        = {258},
  pages         = {4807--4815},
  publisher     = {PMLR},
  year          = {2025},
  eprint        = {2411.07154},
  archiveprefix = {arXiv}
}

@article{cui2023scalable,
  title         = {Scalable Conditional Deep Inverse {R}osenblatt Transports Using Tensor Trains and Gradient-Based Dimension Reduction},
  author        = {Cui, Tiangang and Dolgov, Sergey and Zahm, Olivier},
  journal       = {Journal of Computational Physics},
  volume        = {485},
  eid           = {112103},
  year          = {2023},
  doi           = {10.1016/j.jcp.2023.112103}
}

@inproceedings{park2020measure,
  title         = {A Measure-Theoretic Approach to Kernel Conditional Mean Embeddings},
  author        = {Park, Junhyung and Muandet, Krikamol},
  booktitle     = {Advances in Neural Information Processing Systems (NeurIPS)},
  volume        = {33},
  pages         = {21247--21259},
  year          = {2020},
  eprint        = {2002.03689},
  archiveprefix = {arXiv}
}

@misc{huang2022evaluating,
  title         = {Evaluating Aleatoric Uncertainty via Conditional Generative Models},
  author        = {Huang, Ziyi and Lam, Henry and Zhang, Haofeng},
  howpublished  = {arXiv preprint},
  year          = {2022},
  eprint        = {2206.04287},
  archiveprefix = {arXiv}
}

@article{goodfellow2014,
  title         = {Generative Adversarial Networks},
  author        = {Goodfellow, Ian and Pouget-Abadie, Jean and Mirza, Mehdi and Xu, Bing and Warde-Farley, David and Ozair, Sherjil and Courville, Aaron and Bengio, Yoshua},
  journal       = {Communications of the ACM},
  volume        = {63},
  number        = {11},
  pages         = {139--144},
  year          = {2020},
  doi           = {10.1145/3422622}
}

@inproceedings{rezende2015,
  title         = {Variational Inference with Normalizing Flows},
  author        = {Rezende, Danilo Jimenez and Mohamed, Shakir},
  booktitle     = {Proceedings of the 32nd International Conference on Machine Learning (ICML)},
  series        = {Proceedings of Machine Learning Research},
  volume        = {37},
  pages         = {1530--1538},
  publisher     = {PMLR},
  year          = {2015},
  eprint        = {1505.05770},
  archiveprefix = {arXiv}
}

@article{papamakarios2021,
  title         = {Normalizing Flows for Probabilistic Modeling and Inference},
  author        = {Papamakarios, George and Nalisnick, Eric and Rezende, Danilo Jimenez and Mohamed, Shakir and Lakshminarayanan, Balaji},
  journal       = {Journal of Machine Learning Research},
  volume        = {22},
  number        = {57},
  pages         = {1--64},
  year          = {2021},
  eprint        = {1912.02762},
  archiveprefix = {arXiv}
}

@article{kobyzev2021,
  title         = {Normalizing Flows: An Introduction and Review of Current Methods},
  author        = {Kobyzev, Ivan and Prince, Simon J. D. and Brubaker, Marcus A.},
  journal       = {IEEE Transactions on Pattern Analysis and Machine Intelligence},
  volume        = {43},
  number        = {11},
  pages         = {3964--3979},
  year          = {2021},
  doi           = {10.1109/TPAMI.2020.2992934}
}

@book{pardo2018,
  title         = {Statistical Inference Based on Divergence Measures},
  author        = {Pardo, Leandro},
  publisher     = {Chapman \& Hall/CRC},
  address       = {Boca Raton, FL},
  year          = {2006},
  doi           = {10.1201/9781420034813}
}

@inproceedings{ho2020,
  title         = {Denoising Diffusion Probabilistic Models},
  author        = {Ho, Jonathan and Jain, Ajay and Abbeel, Pieter},
  booktitle     = {Advances in Neural Information Processing Systems (NeurIPS)},
  volume        = {33},
  pages         = {6840--6851},
  year          = {2020},
  eprint        = {2006.11239},
  archiveprefix = {arXiv}
}

@misc{trippe2018,
  title         = {Conditional Density Estimation with {B}ayesian Normalising Flows},
  author        = {Trippe, Brian L. and Turner, Richard E.},
  howpublished  = {arXiv preprint},
  year          = {2018},
  eprint        = {1802.04908},
  archiveprefix = {arXiv}
}

@misc{winkler2019,
  title         = {Learning Likelihoods with Conditional Normalizing Flows},
  author        = {Winkler, Christina and Worrall, Daniel and Hoogeboom, Emiel and Welling, Max},
  howpublished  = {arXiv preprint},
  year          = {2019},
  eprint        = {1912.00042},
  archiveprefix = {arXiv}
}

@misc{batzolis2021,
  title         = {Conditional Image Generation with Score-Based Diffusion Models},
  author        = {Batzolis, Georgios and Stanczuk, Jan and Sch{\"o}nlieb, Carola-Bibiane and Etmann, Christian},
  howpublished  = {arXiv preprint},
  year          = {2021},
  eprint        = {2111.13606},
  archiveprefix = {arXiv}
}

@article{saharia2022,
  title         = {Image Super-Resolution via Iterative Refinement},
  author        = {Saharia, Chitwan and Ho, Jonathan and Chan, William and Salimans, Tim and Fleet, David J. and Norouzi, Mohammad},
  journal       = {IEEE Transactions on Pattern Analysis and Machine Intelligence},
  volume        = {45},
  number        = {4},
  pages         = {4713--4726},
  year          = {2023},
  doi           = {10.1109/TPAMI.2022.3204461}
}

@article{knothe1957,
  title         = {Contributions to the Theory of Convex Bodies},
  author        = {Knothe, Herbert},
  journal       = {Michigan Mathematical Journal},
  volume        = {4},
  number        = {1},
  pages         = {39--52},
  year          = {1957},
  doi           = {10.1307/mmj/1028990175}
}

@article{rosenblatt1952,
  title         = {Remarks on a Multivariate Transformation},
  author        = {Rosenblatt, Murray},
  journal       = {The Annals of Mathematical Statistics},
  volume        = {23},
  number        = {3},
  pages         = {470--472},
  year          = {1952},
  doi           = {10.1214/aoms/1177729394}
}

@article{elmoselhy2012,
  title         = {{B}ayesian Inference with Optimal Maps},
  author        = {El Moselhy, Tarek A. and Marzouk, Youssef M.},
  journal       = {Journal of Computational Physics},
  volume        = {231},
  number        = {23},
  pages         = {7815--7850},
  year          = {2012},
  doi           = {10.1016/j.jcp.2012.07.022}
}

@article{carlier2010,
  title         = {From {K}nothe's Transport to {B}renier's Map and a Continuation Method for Optimal Transport},
  author        = {Carlier, Guillaume and Galichon, Alfred and Santambrogio, Filippo},
  journal       = {SIAM Journal on Mathematical Analysis},
  volume        = {41},
  number        = {6},
  pages         = {2554--2576},
  year          = {2010},
  doi           = {10.1137/080740647}
}

@article{carlier2016,
  title         = {Vector Quantile Regression: An Optimal Transport Approach},
  author        = {Carlier, Guillaume and Chernozhukov, Victor and Galichon, Alfred},
  journal       = {The Annals of Statistics},
  volume        = {44},
  number        = {3},
  pages         = {1165--1192},
  year          = {2016},
  doi           = {10.1214/15-AOS1401}
}

@article{baptista2023approximation,
  title         = {An Approximation Theory Framework for Measure-Transport Sampling Algorithms},
  author        = {Baptista, Ricardo and Hosseini, Bamdad and Kovachki, Nikola B. and Marzouk, Youssef M. and Sagiv, Amir},
  journal       = {Mathematics of Computation},
  volume        = {94},
  number        = {354},
  pages         = {1863--1909},
  year          = {2025},
  doi           = {10.1090/mcom/4013}
}

@article{pandey2024diffeomorphic,
  title         = {Diffeomorphic Measure Matching with Kernels for Generative Modeling},
  author        = {Pandey, Biraj and Hosseini, Bamdad and Batlle, Pau and Owhadi, Houman},
  journal       = {SIAM Journal on Mathematics of Data Science},
  volume        = {7},
  number        = {3},
  pages         = {937--964},
  year          = {2025},
  doi           = {10.1137/24M1642202}
}

@article{bogachev2005,
  title         = {Triangular Transformations of Measures},
  author        = {Bogachev, Vladimir I. and Kolesnikov, Alexander V. and Medvedev, Kirill V.},
  journal       = {Sbornik: Mathematics},
  volume        = {196},
  number        = {3},
  pages         = {309--335},
  year          = {2005},
  doi           = {10.1070/SM2005v196n03ABEH000882}
}

@article{bogachev2006,
  title         = {Nonlinear Transformations of Convex Measures},
  author        = {Bogachev, Vladimir I. and Kolesnikov, Alexander V.},
  journal       = {Theory of Probability \& Its Applications},
  volume        = {50},
  number        = {1},
  pages         = {34--52},
  year          = {2006},
  doi           = {10.1137/S0040585X97981457}
}

@article{zech2022a,
  title         = {Sparse Approximation of Triangular Transports, Part {I}: The Finite-Dimensional Case},
  author        = {Zech, Jakob and Marzouk, Youssef},
  journal       = {Constructive Approximation},
  volume        = {55},
  number        = {3},
  pages         = {919--986},
  year          = {2022},
  doi           = {10.1007/s00365-022-09569-2}
}

@article{zech2022b,
  title         = {Sparse Approximation of Triangular Transports, Part {II}: The Infinite-Dimensional Case},
  author        = {Zech, Jakob and Marzouk, Youssef},
  journal       = {Constructive Approximation},
  volume        = {55},
  number        = {3},
  pages         = {987--1036},
  year          = {2022},
  doi           = {10.1007/s00365-022-09570-9}
}

@inproceedings{irons2022,
  title         = {Triangular Flows for Generative Modeling: Statistical Consistency, Smoothness Classes, and Fast Rates},
  author        = {Irons, Nicholas J. and Scetbon, Meyer and Pal, Soumik and Harchaoui, Zaid},
  booktitle     = {Proceedings of the 25th International Conference on Artificial Intelligence and Statistics (AISTATS)},
  series        = {Proceedings of Machine Learning Research},
  volume        = {151},
  pages         = {10161--10195},
  publisher     = {PMLR},
  year          = {2022},
  eprint        = {2112.15595},
  archiveprefix = {arXiv}
}

@misc{wang2022minimax,
  title         = {On Minimax Density Estimation via Measure Transport},
  author        = {Wang, Sven and Marzouk, Youssef},
  howpublished  = {arXiv preprint},
  year          = {2022},
  eprint        = {2207.10231},
  archiveprefix = {arXiv}
}

@inproceedings{jaini2020,
  title         = {Tails of {L}ipschitz Triangular Flows},
  author        = {Jaini, Priyank and Kobyzev, Ivan and Yu, Yaoliang and Brubaker, Marcus},
  booktitle     = {Proceedings of the 37th International Conference on Machine Learning (ICML)},
  series        = {Proceedings of Machine Learning Research},
  volume        = {119},
  pages         = {4673--4681},
  publisher     = {PMLR},
  year          = {2020},
  eprint        = {1907.04481},
  archiveprefix = {arXiv}
}

@misc{gonzalezsanz2025,
  title         = {Linearization of {M}onge--{A}mp{\`e}re Equations and Statistical Applications},
  author        = {Gonz{\'a}lez-Sanz, Alberto and Sheng, Shunan},
  howpublished  = {arXiv preprint},
  year          = {2024},
  eprint        = {2408.06534},
  archiveprefix = {arXiv}
}

@article{zhou2023,
  title         = {A Deep Generative Approach to Conditional Sampling},
  author        = {Zhou, Xingyu and Jiao, Yuling and Liu, Jin and Huang, Jian},
  journal       = {Journal of the American Statistical Association},
  volume        = {118},
  number        = {543},
  pages         = {1837--1848},
  year          = {2023},
  doi           = {10.1080/01621459.2021.2016424}
}

@inproceedings{kerrigan2024dynamic,
  title         = {Dynamic Conditional Optimal Transport through Simulation-Free Flows},
  author        = {Kerrigan, Gavin and Migliorini, Giosue and Smyth, Padhraic},
  booktitle     = {Advances in Neural Information Processing Systems (NeurIPS)},
  volume        = {37},
  pages         = {93602--93642},
  year          = {2024},
  doi           = {10.52202/079017-2968}
}

@article{benton2023error,
  title         = {Error Bounds for Flow Matching Methods},
  author        = {Benton, Joe and Deligiannidis, George and Doucet, Arnaud},
  journal       = {Transactions on Machine Learning Research},
  year          = {2024},
  eprint        = {2305.16860},
  archiveprefix = {arXiv}
}

@misc{aljarrah2025error,
  title         = {Error Analysis of Triangular Optimal Transport Maps for Filtering},
  author        = {Al-Jarrah, Mohammad and Hosseini, Bamdad and Jin, Niyizhen and Martino, Michele and Taghvaei, Amirhossein},
  howpublished  = {arXiv preprint},
  year          = {2025},
  eprint        = {2510.19283},
  archiveprefix = {arXiv}
}

@article{gretton2012,
  title         = {A Kernel Two-Sample Test},
  author        = {Gretton, Arthur and Borgwardt, Karsten M. and Rasch, Malte J. and Sch{\"o}lkopf, Bernhard and Smola, Alexander},
  journal       = {Journal of Machine Learning Research},
  volume        = {13},
  number        = {25},
  pages         = {723--773},
  year          = {2012},
  url           = {https://jmlr.org/papers/v13/gretton12a.html}
}

@article{muandet2017kernel,
  title         = {Kernel Mean Embedding of Distributions: A Review and Beyond},
  author        = {Muandet, Krikamol and Fukumizu, Kenji and Sriperumbudur, Bharath and Sch{\"o}lkopf, Bernhard},
  journal       = {Foundations and Trends{\textregistered} in Machine Learning},
  volume        = {10},
  number        = {1--2},
  pages         = {1--141},
  year          = {2017},
  doi           = {10.1561/2200000060}
}

@inproceedings{song2009,
  title         = {{H}ilbert Space Embeddings of Conditional Distributions with Applications to Dynamical Systems},
  author        = {Song, Le and Huang, Jonathan and Smola, Alex and Fukumizu, Kenji},
  booktitle     = {Proceedings of the 26th Annual International Conference on Machine Learning (ICML)},
  pages         = {961--968},
  publisher     = {ACM},
  year          = {2009},
  doi           = {10.1145/1553374.1553497}
}

@inproceedings{chatterjee2025,
  title         = {One-Shot Conditional Sampling: {MMD} Meets Nearest Neighbors},
  author        = {Chatterjee, Anirban and Choudhury, Sayantan and Hore, Rohan},
  booktitle     = {Proceedings of the 43rd International Conference on Machine Learning (ICML)},
  series        = {Proceedings of Machine Learning Research},
  volume        = {306},
  pages         = {13115--13163},
  publisher     = {PMLR},
  year          = {2026},
  eprint        = {2509.25507},
  archiveprefix = {arXiv}
}

@book{bogachev2007measure-volII,
  title         = {Measure Theory},
  author        = {Bogachev, Vladimir I.},
  volume        = {II},
  publisher     = {Springer},
  address       = {Berlin, Heidelberg},
  year          = {2007},
  doi           = {10.1007/978-3-540-34514-5}
}

@book{berlinet2011reproducing,
  title         = {Reproducing Kernel {H}ilbert Spaces in Probability and Statistics},
  author        = {Berlinet, Alain and Thomas-Agnan, Christine},
  publisher     = {Springer},
  address       = {New York},
  year          = {2004},
  doi           = {10.1007/978-1-4419-9096-9}
}

@book{owhadi2019operator,
  title         = {Operator-Adapted Wavelets, Fast Solvers, and Numerical Homogenization: From a Game Theoretic Approach to Numerical Approximation and Algorithm Design},
  author        = {Owhadi, Houman and Scovel, Clint},
  series        = {Cambridge Monographs on Applied and Computational Mathematics},
  volume        = {35},
  publisher     = {Cambridge University Press},
  address       = {Cambridge},
  year          = {2019},
  doi           = {10.1017/9781108594967}
}

@article{minh2010some,
  title         = {Some Properties of {G}aussian Reproducing Kernel {H}ilbert Spaces and Their Implications for Function Approximation and Learning Theory},
  author        = {Minh, Ha Quang},
  journal       = {Constructive Approximation},
  volume        = {32},
  number        = {2},
  pages         = {307--338},
  year          = {2010},
  doi           = {10.1007/s00365-009-9080-0}
}

@article{giner2009necessary,
  title         = {Necessary and Sufficient Conditions for the Interchange Between Infimum and the Symbol of Integration},
  author        = {Giner, Emmanuel},
  journal       = {Set-Valued and Variational Analysis},
  volume        = {17},
  number        = {4},
  pages         = {321--357},
  year          = {2009},
  doi           = {10.1007/s11228-009-0119-y}
}

@article{hiai1977integrals,
  title         = {Integrals, Conditional Expectations, and Martingales of Multivalued Functions},
  author        = {Hiai, Fumio and Umegaki, Hisaharu},
  journal       = {Journal of Multivariate Analysis},
  volume        = {7},
  number        = {1},
  pages         = {149--182},
  year          = {1977},
  doi           = {10.1016/0047-259X(77)90037-9}
}

@article{hornik1991approximation,
  title         = {Approximation Capabilities of Multilayer Feedforward Networks},
  author        = {Hornik, Kurt},
  journal       = {Neural Networks},
  volume        = {4},
  number        = {2},
  pages         = {251--257},
  year          = {1991},
  doi           = {10.1016/0893-6080(91)90009-T}
}

@book{wainwright2019high,
  title         = {High-Dimensional Statistics: A Non-Asymptotic Viewpoint},
  author        = {Wainwright, Martin J.},
  series        = {Cambridge Series in Statistical and Probabilistic Mathematics},
  volume        = {48},
  publisher     = {Cambridge University Press},
  address       = {Cambridge},
  year          = {2019},
  doi           = {10.1017/9781108627771}
}

@article{divol2025optimal-sm,
  title         = {Supplement to ``{O}ptimal Transport Map Estimation in General Function Spaces''},
  author        = {Divol, Vincent and Niles-Weed, Jonathan and Pooladian, Aram-Alexandre},
  journal       = {The Annals of Statistics},
  volume        = {53},
  number        = {3},
  year          = {2025},
  doi           = {10.1214/24-AOS2482SUPP},
  note          = {Supplementary material to the article in Ann. Statist. 53(3):963--988, \url{https://doi.org/10.1214/24-AOS2482}}
}

@inproceedings{park2023towards,
  title         = {Towards Empirical Process Theory for Vector-Valued Functions: Metric Entropy of Smooth Function Classes},
  author        = {Park, Junhyung and Muandet, Krikamol},
  booktitle     = {Proceedings of the 34th International Conference on Algorithmic Learning Theory (ALT)},
  series        = {Proceedings of Machine Learning Research},
  volume        = {201},
  pages         = {1216--1260},
  publisher     = {PMLR},
  year          = {2023},
  eprint        = {2202.04415},
  archiveprefix = {arXiv}
}

@article{van1987new,
  title         = {A New Approach to Least-Squares Estimation, with Applications},
  author        = {van de Geer, Sara},
  journal       = {The Annals of Statistics},
  volume        = {15},
  number        = {2},
  pages         = {587--602},
  year          = {1987},
  doi           = {10.1214/aos/1176350362}
}

@article{van2002m,
  title         = {{M}-Estimation Using Penalties or Sieves},
  author        = {van de Geer, Sara},
  journal       = {Journal of Statistical Planning and Inference},
  volume        = {108},
  number        = {1--2},
  pages         = {55--69},
  year          = {2002},
  doi           = {10.1016/S0378-3758(02)00270-7}
}

@article{sriperumbudur2010hilbert,
  title         = {{H}ilbert Space Embeddings and Metrics on Probability Measures},
  author        = {Sriperumbudur, Bharath K. and Gretton, Arthur and Fukumizu, Kenji and Sch{\"o}lkopf, Bernhard and Lanckriet, Gert R. G.},
  journal       = {Journal of Machine Learning Research},
  volume        = {11},
  number        = {50},
  pages         = {1517--1561},
  year          = {2010},
  eprint        = {0907.5309},
  archiveprefix = {arXiv}
}

@misc{briol2019statistical,
  title         = {Statistical Inference for Generative Models with Maximum Mean Discrepancy},
  author        = {Briol, Fran{\c{c}}ois-Xavier and Barp, Alessandro and Duncan, Andrew B. and Girolami, Mark},
  howpublished  = {arXiv preprint},
  year          = {2019},
  eprint        = {1906.05944},
  archiveprefix = {arXiv}
}

@inproceedings{johnsonlocalizing,
  title         = {Localizing Conditional Transport in Triangular Flows},
  author        = {Johnson-V{\'a}zquez, Alexander and Hosseini, Bamdad},
  booktitle     = {NeurIPS 2026 Workshop on AI for Stochastic Dynamics},
  year          = {2026},
  note          = {To appear}
}

@article{garbuno2023bayesian,
  title         = {{B}ayesian Posterior Perturbation Analysis with Integral Probability Metrics},
  author        = {Garbuno-I{\~n}igo, Alfredo and Helin, Tapio and Hoffmann, Franca and Hosseini, Bamdad},
  journal       = {Bayesian Analysis},
  year          = {2026},
  doi           = {10.1214/26-BA1608},
  note          = {Advance publication}
}

@inproceedings{akbari2023causal,
  title         = {Causal Discovery via Monotone Triangular Transport Maps},
  author        = {Akbari, Sina and Ganassali, Luca and Kiyavash, Negar},
  booktitle     = {NeurIPS 2023 Workshop on Optimal Transport and Machine Learning (OTML)},
  year          = {2023},
  url           = {https://openreview.net/forum?id=PMGGrKTIii},
  eprint        = {2305.18210},
  archiveprefix = {arXiv}
}

@article{calle2026generative,
  title         = {Generative Multi-Scale Modeling via Spatial Autoregressive Transport Maps},
  author        = {Calle-Saldarriaga, Alejandro and Wiemann, Paul F. V. and Katzfuss, Matthias},
  journal       = {Technometrics},
  volume        = {68},
  number        = {4},
  pages         = {836--849},
  year          = {2026},
  doi           = {10.1080/00401706.2026.2689970}
}

@misc{blorstad2026learning,
  title         = {Learning the Structure of Triangular Transport Maps},
  author        = {Bl{\o}rstad, Morten and Parviainen, Pekka and Lunde, Berent {\AA}nund Str{\o}mnes},
  howpublished  = {arXiv preprint},
  year          = {2026},
  eprint        = {2609.37122},
  archiveprefix = {arXiv}
}

@misc{davis2026multifidelity,
  title         = {Multifidelity Formulations for Triangular Transport},
  author        = {Davis, Owen and Sharp, Daniel and Marzouk, Youssef and Geraci, Gianluca},
  howpublished  = {arXiv preprint},
  year          = {2026},
  eprint        = {2610.00698},
  archiveprefix = {arXiv}
}

@inproceedings{kingma2013auto,
  title         = {Auto-Encoding Variational {B}ayes},
  author        = {Kingma, Diederik P. and Welling, Max},
  booktitle     = {2nd International Conference on Learning Representations (ICLR)},
  year          = {2014},
  eprint        = {1312.6114},
  archiveprefix = {arXiv}
}

@inproceedings{tolstikhin2017wasserstein,
  title         = {{W}asserstein Auto-Encoders},
  author        = {Tolstikhin, Ilya and Bousquet, Olivier and Gelly, Sylvain and Sch{\"o}lkopf, Bernhard},
  booktitle     = {6th International Conference on Learning Representations (ICLR)},
  year          = {2018},
  eprint        = {1711.01558},
  archiveprefix = {arXiv}
}

@misc{al2026conditional,
  title         = {Conditional Sampling via {W}asserstein Autoencoders and Triangular Transport},
  author        = {Al-Jarrah, Mohammad and Martino, Michele and Yim, Marcus and Hosseini, Bamdad and Taghvaei, Amirhossein},
  howpublished  = {arXiv preprint},
  year          = {2026},
  eprint        = {2604.02644},
  archiveprefix = {arXiv}
}

@article{cui2014likelihood,
  title         = {Likelihood-Informed Dimension Reduction for Nonlinear Inverse Problems},
  author        = {Cui, Tiangang and Martin, James and Marzouk, Youssef M. and Solonen, Antti and Spantini, Alessio},
  journal       = {Inverse Problems},
  volume        = {30},
  number        = {11},
  eid           = {114015},
  year          = {2014},
  doi           = {10.1088/0266-5611/30/11/114015}
}

@article{cui2025subspace,
  title         = {Subspace Accelerated Measure Transport Methods for Fast and Scalable Sequential Experimental Design, with Application to Photoacoustic Imaging},
  author        = {Cui, Tiangang and Koval, Karina and Herzog, Roland and Scheichl, Robert},
  journal       = {SIAM/ASA Journal on Uncertainty Quantification},
  volume        = {14},
  number        = {4},
  pages         = {1476--1503},
  year          = {2026},
  doi           = {10.1137/25M1735329}
}

@article{cui2022unified,
  title         = {A Unified Performance Analysis of Likelihood-Informed Subspace Methods},
  author        = {Cui, Tiangang and Tong, Xin T.},
  journal       = {Bernoulli},
  volume        = {28},
  number        = {4},
  pages         = {2788--2815},
  year          = {2022},
  doi           = {10.3150/21-BEJ1437}
}

@misc{baptista2026approximating,
  title         = {Approximating Measures on Function Spaces: Transport and Truncation},
  author        = {Baptista, Ricardo and Hosseini, Bamdad and Hsu, Alexander W.},
  howpublished  = {arXiv preprint},
  year          = {2026},
  eprint        = {2609.17802},
  archiveprefix = {arXiv}
}

@book{vanderVaartWellner2023,
  title         = {Weak Convergence and Empirical Processes: With Applications to Statistics},
  author        = {van der Vaart, Aad W. and Wellner, Jon A.},
  edition       = {Second},
  series        = {Springer Series in Statistics},
  publisher     = {Springer},
  address       = {Cham},
  year          = {2023},
  doi           = {10.1007/978-3-031-29040-4}
}

@article{bousquet2002bennett,
  title         = {A {B}ennett Concentration Inequality and Its Application to Suprema of Empirical Processes},
  author        = {Bousquet, Olivier},
  journal       = {Comptes Rendus Math{\'e}matique},
  volume        = {334},
  number        = {6},
  pages         = {495--500},
  year          = {2002},
  doi           = {10.1016/S1631-073X(02)02292-6}
}

@book{steinwart2008support,
  title         = {Support Vector Machines},
  author        = {Steinwart, Ingo and Christmann, Andreas},
  series        = {Information Science and Statistics},
  publisher     = {Springer},
  address       = {New York},
  year          = {2008},
  doi           = {10.1007/978-0-387-77242-4}
}

@article{sriperumbudur2011universality,
  title         = {Universality, Characteristic Kernels and {RKHS} Embedding of Measures},
  author        = {Sriperumbudur, Bharath K. and Fukumizu, Kenji and Lanckriet, Gert R. G.},
  journal       = {Journal of Machine Learning Research},
  volume        = {12},
  number        = {70},
  pages         = {2389--2410},
  year          = {2011},
  eprint        = {1003.0887},
  archiveprefix = {arXiv}
}

@article{shen2025engression,
  author  = {Shen, Xinwei and Meinshausen, Nicolai},
  title   = {Engression: extrapolation through the lens of distributional regression},
  journal = {Journal of the Royal Statistical Society Series B: Statistical Methodology},
  pages   = {653--677},
  year    = {2025},
  doi     = {10.1093/jrsssb/qkae108}
}

@article{sejdinovic2013equivalence,
  author  = {Sejdinovic, Dino and Sriperumbudur, Bharath and Gretton, Arthur and Fukumizu, Kenji},
  title   = {Equivalence of distance-based and {RKHS}-based statistics in hypothesis testing},
  journal = {The Annals of Statistics},
  volume  = {41},
  number  = {5},
  pages   = {2263--2291},
  year    = {2013},
  doi     = {10.1214/13-AOS1140}
}

@misc{chai2026neural,
  author = {Chai, Jinhang and Fan, Jianqing and Gu, Yihong},
  title  = {Neural generative distributional regression},
  year   = {2026},
  eprint = {2605.02062},
  archivePrefix = {arXiv},
  doi    = {10.48550/arXiv.2605.02062}
}

@misc{huang2026engression,
  author = {Huang, Jiaqi and Xu, Gongjun and Zhu, Ji},
  title  = {Theoretical analysis of engression and reverse {M}arkov engression},
  year   = {2026},
  eprint = {2606.01002},
  archivePrefix = {arXiv},
  doi    = {10.48550/arXiv.2606.01002}
}

@misc{chen2026engression,
  author = {Chen, Juntong and Guo, Zijian and Shen, Xinwei},
  title  = {Error analysis of neural-network-based engression},
  year   = {2026},
  eprint = {2607.27723},
  archivePrefix = {arXiv},
  doi    = {10.48550/arXiv.2607.27723}
}

@article{shen2025reverse,
  author  = {Shen, Xinwei and Meinshausen, Nicolai and Zhang, Tong},
  title   = {Reverse {M}arkov Learning: Multi-Step Generative Models for Complex Distributions},
  year    = {2025},
  eprint = {2502.13747},
  archivePrefix = {arXiv}
}

\appendix
\section{Explicit computation~\Cref{ex:counter-CMMD-counter-example}}
\label{app:CMMD-counter-example}
For a fixed realization $y$ of $Y$, we can explicitly compute
\begin{align*}
\begin{split}
D_{\text{Cond.}}\left(\nu(\cdot \mid y), \hat{\nu}(\cdot \mid y) \right)&=D_{\text{Cond.}}(\delta_{\alpha y}, \delta_{-\alpha y})=K(\alpha y , \alpha y)+K(-\alpha y , -\alpha y)-2K(\alpha y , -\alpha y)\\
& =1+1-2\exp\left(-\frac{1}{2}(2\alpha y)^2\right)= 2\left(1-\exp\left(-2\alpha^2 y^2\right)\right),
\end{split}
\end{align*}
yielding~\eqref{eq: first part example}.
Since $Y\sim N(0,1)$, we have that $\nu= N(0,\Sigma_{\alpha})$ and $\widehat \nu=N(0,\Sigma_{-\alpha})$ where
\begin{align*}
\Sigma_{\alpha} = \begin{bmatrix}
1 & \alpha \\
\alpha & \alpha^2
\end{bmatrix}\, , \qquad \Sigma_{-\alpha} = \begin{bmatrix}
1 & -\alpha \\
-\alpha & \alpha^2
\end{bmatrix}\, .
\end{align*}
Then, adopting the notation $N_{\Sigma}\coloneqq N(0,\Sigma)$ for a centered Gaussian measure, which may be degenerate as $\Sigma_{\pm\alpha}$ is singular, denoting $x:=(y,u)$, with $K(x,x') := K(y,y')K(u,u')$ and $z = x-x'$, we can explicitly compute also the joint divergence. Indeed, if $x\sim N_\Sigma$ and $x'\sim N_{\Sigma'}$ are independent, then $z\sim N_{\Sigma+\Sigma'}$, and $\int\exp(-\frac{\|z\|^2}{2})N_\Sigma(\dd z)=\left(\det\left(\textnormal{Id}+\Sigma\right)\right)^{-1/2}$ for any positive semidefinite $\Sigma$. Thus,
\begin{align*}
D_{\text{Joint}}(\nu,\widehat \nu) &= \int K(x,x')\,\nu(\dd x)\,\nu(\dd x')+\int K(x,x')\,\widehat\nu(\dd x)\,\widehat\nu(\dd x')\\
&\hspace{15em}-2\int K(x,x')\,\nu(\dd x)\,\widehat\nu(\dd x')\\
&= \int \exp\left(-\frac{\|z\|^2}{2}\right)N_{2\Sigma_\alpha}(\dd z)+\int \exp\left(-\frac{\|z\|^2}{2}\right)N_{2\Sigma_{-\alpha}}(\dd z)\\
&\hspace{15em}-2\int \exp\left(-\frac{\|z\|^2}{2}\right)N_{\Sigma_{\alpha}+\Sigma_{-\alpha}}(\dd z)\\
&=\left(\text{det}\left(\textnormal{Id}+2\Sigma_\alpha\right)\right)^{-1/2}+\left(\text{det}\left(\textnormal{Id}+2\Sigma_{-\alpha}\right)\right)^{-1/2}\\
&\hspace{15em}-2\left(\text{det}\left(\textnormal{Id}+\Sigma_\alpha+\Sigma_{-\alpha}\right)\right)^{-1/2}\\
&=\frac{2}{\sqrt{3 + 2\alpha^2}} -  \frac{2}{\sqrt{3(1 + 2\alpha^2)}},
\end{align*}
obtaining~\eqref{eq: second part example}.

\section{Proofs of preliminary results and problem setup of~\Cref{sec:preliminaries} and~\Cref{sec: setup}}\label{app: sec2}

We first present the proof of the master theorem of triangular transport in~\Cref{prop:triangular-transport-conditioning}, which is just a generalization of the original result in~\cite[Thm. 2.4]{baptista2024conditional}.
\begin{proof}[Proof of~\Cref{prop:triangular-transport-conditioning}] 
By the disintegration theorem~\cite[Sec.~10.4]{bogachev2007measure-volII}, there exists a family $\{\widetilde\eta(\cdot\mid\F(z)=y)\}_{y\in\Y}\subset\PP(\Z)$ such that
\begin{align}\label{eq: app disintegration along F}
\begin{split}
&\eta_\Z(\dd z)=\int_\Y\widetilde\eta(\dd z\mid\F(z)=y)\,(\F\#\eta_\Z)(\dd y),\\
&\widetilde\eta\left(\Z\setminus\F^{-1}(\{y\})\mid\F(z)=y\right)=0\ \text{ for }\F\#\eta_\Z\text{-a.e. }y.
\end{split}
\end{align}
For $y\in\Y$, define the probability measure $\mu_y\in\PP(\U)$ by
\begin{align*}
&\mu_y\coloneqq\int_{\F^{-1}(\{y\})}\T(y,\cdot)\#\eta(\cdot\mid z)\,\widetilde\eta(\dd z\mid\F(z)=y),\\
&\mu_y(A)=\int_\Z\eta\left(\T(y,\cdot)^{-1}(A)\mid z\right)\widetilde\eta(\dd z\mid\F(z)=y)
\end{align*}
for every $A\in\mathcal{B}(\U)$. By standard measurability properties of kernels, $y\mapsto\mu_y(A)$ is measurable for every $A\in\mathcal{B}(\U)$.
Let $B_\Y\in\mathcal{B}(\Y)$. Since the second component of $\T^\F_\t$ always lies in $\U$, we have $(\T^\F_\t)^{-1}(B_\Y\times\U)=\F^{-1}(B_\Y)\times\V$. Thus, by $\T^\F_\t\#\eta=\nu$,
\begin{align*}
\nu_\Y(B_\Y)=\nu(B_\Y\times\U)=\eta\left(\F^{-1}(B_\Y)\times\V\right)=\eta_\Z\left(\F^{-1}(B_\Y)\right)=(\F\#\eta_\Z)(B_\Y).
\end{align*}

Next, let $B\in\mathcal{B}(\Y\times\U)$, and denote by $B_y\coloneqq\{u\in\U\mid(y,u)\in B\}$ its slices, so that $\mathbf{1}_B(y,u)=\mathbf{1}_{B_y}(u)$. For $(y,z)\in\Y\times\Z$, let
\begin{align*}
\Phi(y,z)\coloneqq\int_\V\mathbf{1}_{B_y}\left(\T(y,v)\right)\eta(\dd v\mid z)=\left(\T(y,\cdot)\#\eta(\cdot\mid z)\right)(B_y).
\end{align*}
By $\T^\F_\t\#\eta=\nu$ and the disintegration of $\eta$ along $\Z$,
\begin{align*}
\nu(B)=\int_{\Z\times\V}\mathbf{1}_B\left(\F(z),\T(\F(z),v)\right)\eta(\dd z,\dd v)&=\int_\Z\int_\V\mathbf{1}_{B_{\F(z)}}\left(\T(\F(z),v)\right)\eta(\dd v\mid z)\,\eta_\Z(\dd z)\\
&=\int_\Z\Phi(\F(z),z)\,\eta_\Z(\dd z).
\end{align*}
By~\eqref{eq: app disintegration along F}, $\F(z)=y$ for $\widetilde\eta(\cdot\mid\F(z)=y)$-a.e.\ $z$, for $\F\#\eta_\Z$-a.e.\ $y$. Thus,
\begin{align*}
\nu(B)=\int_\Y\int_\Z\Phi(\F(z),z)\,\widetilde\eta(\dd z\mid\F(z)=y)\,(\F\#\eta_\Z)(\dd y)&=\int_\Y\int_\Z\Phi(y,z)\,\widetilde\eta(\dd z\mid\F(z)=y)\,\nu_\Y(\dd y)\\
&=\int_\Y\mu_y(B_y)\,\nu_\Y(\dd y).
\end{align*}
$\{\mu_y\}_{y\in\Y}$ is a system of conditional measures of $\nu$ in the sense of~\cite[Def.~2.1]{baptista2024conditional}. By the essential uniqueness of conditional measures~\cite[Prop.~2.2(b)]{baptista2024conditional}, it follows that \begin{align*}
\int_{\F^{-1}(\{y\})}\T(y,\cdot)\#\eta(\cdot\mid z)\,\widetilde\eta(\dd z\mid\F(z)=y):=\mu_y=\nu(\cdot\mid y),\qquad \textnormal{for }\nu_\Y\textnormal{-a.e.} y\in\Y.
\end{align*}
proving the main claim of~\Cref{prop:triangular-transport-conditioning}. 

Next, if $\F$ is invertible, then $\F^{-1}(\{y\})=\{\F^{-1}(y)\}$ is a singleton for every $y\in\F(\Z)$. Thus, $\widetilde\eta(\cdot\mid\F(z)=y)=\delta_{\F^{-1}(y)}$ for $\F\#\eta_\Z$-a.e.\ $y$, and therefore $\mu_y=\T(y,\cdot)\#\eta(\cdot\mid\F^{-1}(y))$.

If $\eta=\eta_\Z\otimes\eta_\V$, then $\eta(\cdot\mid z)=\eta_\V$ for every $z\in\Z$ is a system of conditional measures of $\eta$. Then, $\T(y,\cdot)\#\eta(\cdot\mid z)=\T(y,\cdot)\#\eta_\V$ does not depend on $z$, and, since $\widetilde\eta(\cdot\mid\F(z)=y)$ is a probability measure, $\mu_y = \T(y,\cdot)\#\eta_\V$.
{}
\end{proof}

Next, we present the proof of the result showing that controlling $\DD(\nu\pipe \widehat \nu)$ translates into a high-probability bound of the pointwise error $D_{\text{Cond.}}(\nu(\cdot\mid y),\widehat \nu(\cdot\mid y))$
\begin{proof}[Proof of~\Cref{prop: fixed sampled conditional control}]
Let $\SS$ denote the training samples $\{(y_i,u_i)\}_{i=1}^m\overset{\text{i.i.d.}}{\sim}\nu$, together with any additional randomness of the algorithm, so that $\Th$ is $\SS$-measurable and $y\sim\nu_\Y$ is independent of $\SS$. Define
\begin{align*}
\E_{\Th}^2(y)\coloneqq D_{\text{Cond.}}\left(\nu(\cdot\mid y),\Th(y,\cdot)\#\eta_\V\right),\qquad X(y)\coloneqq \E_{\Th}^2(y)-\E^2_{\text{Bias}}(y).
\end{align*}

By~\Cref{cor: conditional transport independent reference}, the conditionals of $\T_\t\#\eta$ are $\T(y,\cdot)\#\eta_\V$ for every $\T\in\TT$. Hence, by~\eqref{ACD-MMD-2},
\begin{align*}
\DD\left(\nu\pipe\T_\t\#\eta\right)=\int_\Y D_{\text{Cond.}}\left(\nu(\cdot\mid y'),\T(y',\cdot)\#\eta_\V\right)\nu_\Y(\dd y'),\qquad\forall\,\T\in\TT.
\end{align*}
By definition of $\E^2_{\text{Bias}}(y)$, the integrand on the right-hand side is at least $\E^2_{\text{Bias}}(y')$. Integrating and taking the infimum over $\T\in\TT$ gives $\int_\Y\E^2_{\text{Bias}}(y')\,\nu_\Y(\dd y')\leq\E^2_{\text{Bias}}$, so that $\Gamma_\TT\geq0$ is well defined. Moreover, since $\Th\in\TT$, we have $X(y)\geq0$. Finally, since $y$ is independent of $\SS$ and $\E^2_{\text{Bias}}(\cdot)$ is deterministic,
\begin{align*}
\mathbb{E}\left[X(y)\mid\SS\right]=\DD\left(\nu\pipe\Th_\t\#\eta\right)-\int_\Y\E^2_{\text{Bias}}(y')\,\nu_\Y(\dd y').
\end{align*}
Let $\Omega_1$ be the event, of probability at least $1-e^{-t}$, on which $\DD(\nu\pipe\Th_\t\#\eta)\leq\E^2_{\text{Bias}}+g(t)m^{-1/\mathfrak{r}}$. Then, on $\Omega_1$,
\begin{align*}
\mathbb{E}\left[X(y)\mid\SS\right]\leq\E^2_{\text{Bias}}-\int_\Y\E^2_{\text{Bias}}(y')\,\nu_\Y(\dd y')+g(t)m^{-1/\mathfrak{r}}=\Gamma_\TT+g(t)m^{-1/\mathfrak{r}}.
\end{align*}
Let $s_t\coloneqq e^t(\Gamma_\TT+g(t)m^{-1/\mathfrak{r}})$ and $\Omega_2\coloneqq\{X(y)>s_t\}$. Since $X(y)\geq0$, the conditional Markov inequality yields, on $\Omega_1$,
\begin{align*}
\mathbb{P}(\Omega_2\mid \SS)\leq\frac{\mathbb{E}[X(y)\mid\SS]}{s_t}\leq e^{-t}.
\end{align*}
If $\Gamma_\TT+g(t)m^{-1/\mathfrak{r}}=0$, we instead use that $\mathbb{E}[X(y)\mid\SS]=0$ and $X(y)\geq0$ force $X(y)=0$ almost surely, so that $\mathbb{P}(\Omega_2\mid\SS)=0$.
Since $\Omega_1$ is $\SS$-measurable, by the law of total probability,
\begin{align*}
\mathbb{P}(\Omega_2)=\mathbb{P}(\Omega_2\cap \Omega_1)+\mathbb{P}(\Omega_2\cap \Omega_1^c)\leq\mathbb{E}\left[\mathbf{1}_{\Omega_1}\,\mathbb{P}(\Omega_2\mid\SS)\right]+\mathbb{P}(\Omega_1^c)\leq e^{-t}+e^{-t}=2e^{-t}.
\end{align*}
On $\Omega_2^c$, which has probability at least $1-2e^{-t}$,
\begin{align*}
D_{\text{Cond.}}\left(\nu(\cdot\mid y),\Th(y,\cdot)\#\eta_\V\right)=\E^2_{\text{Bias}}(y)+X(y)\leq\E^2_{\text{Bias}}(y)+e^t\,\Gamma_\TT+e^tg(t)m^{-1/\mathfrak{r}}.
\end{align*}
If $m\geq(e^tg(t)/\delta)^\mathfrak{r}$, then $e^tg(t)m^{-1/\mathfrak{r}}\leq\delta$, which proves the claim.
\end{proof}

\section{Proofs of~\Cref{thm: main error bound 2}}\label{app: sec4}
In this section, we provide the average conditional estimation error bound on the estimators $\Th^m$ and $\Th^{m,n}$ of~\Cref{thm: main error bound 2}. Due to the length of the proof, we articulate it in several subsections. 

\subsection{Lifted regression problem over conditional mean embeddings}\label{subsec: lifted regression problem over CMEs}
We consider the space $\A$ of measurable maps $g:\Y\to\mathcal{K}$ with $\|g\|_\infty\coloneqq\operatorname*{ess\,sup}_{y\sim\nu_\Y}\|g(y)\|_\mathcal{K}<\infty$, endowed with the pseudo-norm $\|g\|_2\coloneqq\|g\|_{L^2(\Y,\mathcal{K};\nu_\Y)}\leq\|g\|_\infty$. Then, for $\xi=(y,u)\in\Y\times\U$ and $g\in\A$, we define the loss, population risk, and empirical risk:
\begin{align*}
\phi\big(g,(y,u)\big)\coloneqq\left\|K(u,\cdot)-g(y)\right\|^2_\mathcal{K},\qquad R(g)\coloneqq\int\phi(g,\xi)\,\nu(\dd\xi),\qquad  R^m(g)\coloneqq\frac1m\sum_{i=1}^m\phi(g,\xi_i),
\end{align*}
where $\xi_i\coloneqq(y_i,u_i)\overset{\textnormal{i.i.d.}}{\sim}\nu$ are the target training data. Moreover, we let $g^\dag(y) = I_K(\nu(\cdot\mid y))$, i.e. the target conditional mean embedding which $T^\dag$ recovers exactly for $\nu_\Y$-a.e. $y\in\Y$ via $\nu(\cdot\mid y) = \T^\dag(y,\cdot)\#\eta_\V$. We also introduce $g_\T$ and $\G_\TT$ as in~\eqref{eq: short embeddings} and let $\widehat{g}\coloneqq g_{\Th}\in\G_\TT$. With these definitions in hand, as anticipated in~\Cref{rem: cond. var. remark}, we can then lift~\Cref{prob: population level constrained minimization} and~\Cref{prob: empirical minimization} to a simple regression problem and its empirical counterpart over $\G_\TT\subseteq \A$.

\begin{lemma}[Lifted regression problem]\label{lem: app lifted problem}
The following hold.
\begin{enumerate}
\item $\|g^\dag(y)\|_\mathcal{K}\leq B_K$ and $\|g_\T(y)\|_\mathcal{K}\leq B_K$ for all $y\in\Y$ and $\T\in\TT$. In particular, $g^\dag\in\A$ and $\G_\TT\subseteq\A$.
\item For any $\xi\in\Y\times \U$, the function $g\to\phi(g,\xi)$ is convex on $\mathcal{A}$ and, for every $g\in\A$,
\begin{align}\label{eq: app excess risk identity}
R(g)-R(g^\dag)=\|g-g^\dag\|_2^2.
\end{align}
In particular, $g^\dag$ minimizes $R$ over $\A$.
\item For every $\T\in\TT$,
\begin{align*}
M(\nu,\T)&=R(g_\T),\qquad M(\nu^m,\T)=R^m(g_\T),\\
\DD(\nu,\T)&=\|g_\T-g^\dag\|_2^2=R(g_\T)-R(g^\dag).
\end{align*}
Consequently, $R(\widehat g)-R(g^\dag)=\E^2_{\text{Bias}}$, and, for every $\varrho\geq0$, a map $\T\in\TT$ satisfies $M(\nu^m,\T)\leq\inf_{\TT}M(\nu^m,\cdot)+\varrho$ if and only if $R^m(g_\T)\leq\inf_{\mathcal{G}_\TT}R^m+\varrho$.
\end{enumerate}
\end{lemma}

\begin{proof}
The boundedness statement follows by~\Cref{Boundedness kernel} in~\Cref{Assumptions} and Jensen's inequality yielding $\int \|K(u,\cdot)\|_\mathcal{K} \nu(\dd u\mid y)\leq B_K$ and $\|g_\T(y)\|_\mathcal{K}\leq\int\|K(\T(y,v), \cdot)\|_\mathcal{K}\,\eta_\V(\dd v)\leq B_K$. Next, for any fixed $\xi\in\Y\times\U$, $\phi(\cdot,\xi)$ is convex being the composition of an affine and a quadratic one. Next, fix $g\in\A$. Since $\int(K(u, \cdot)-g^\dag(y))\nu(\dd u\mid y)=0$, expanding the square gives
\begin{align*}
\int\left\|K(u,\cdot)-g(y)\right\|^2_\mathcal{K}\nu(\dd u\mid y)=\int\left\|K(u,\cdot)-g^\dag(y)\right\|_\mathcal{K}^2\nu(\dd u\mid y)+\left\|g^\dag(y)-g(y)\right\|_\mathcal{K}^2.
\end{align*}
Thus, integrating against $\nu_\Y$ yields~\eqref{eq: app excess risk identity}. Furthermore, the definition of $M$ in~\eqref{eq:M-derivation} gives $M(\nu,\T)=R(g_\T)$ and $M(\nu^m,\T)=R^m(g_\T)$, while~\eqref{ACD-MMD-2} and~\eqref{eq: app excess risk identity} gives $\DD(\nu,\T)=\|g_\T-g^\dag\|_2^2= R(g_\T)-R(g^\dag)$.  Since $\Th$ minimizes $\DD(\nu,\cdot)$ over $\TT$, we get $R(\widehat g)-R(g^\dag)=\DD(\nu,\Th)=\E^2_{\text{Bias}}$. The last claim follows from $M(\nu^m,\T)= R^m(g_\T)$ and $\mathcal{G}_\TT=\{g_\T\mid\T\in\TT\}$.
\end{proof}

We conclude the section with an upper bound on $\E_{\text{Bias}}$ for Lipschitz $K$ which essentially corresponds to the stability property of $\MMD_K$ defined through Lipschitz kernels (see~\cite[Thm. 3.2]{baptista2023approximation})

\begin{lemma}[Bias bound by $\MMD$ stability]\label{lem: app bias}
Suppose~\Cref{lipshitzness} and~\Cref{embedding} hold. Then, for all $\T,\S\in\TT$,
\begin{align*}
\|g_\T-g_\S\|_\infty\leq L_KC_{1,\eta_\V}\|\T-\S\|_\infty.
\end{align*}
Moreover, if $\T^\dagger$ is as in~\Cref{assumption:triangular-map-existence} with $\T^\dagger(y,\cdot)\in\HH$ for $\nu_\Y$-a.e.\ $y$, then
\begin{align*}
\E_{\text{Bias}}\leq L_KC_{1,\eta_\V}\,\textnormal{dist}_{L^\infty(\Y,\HH;\nu_\Y)}\left(\TT,\T^\dagger\right).
\end{align*}
\end{lemma}
\begin{proof}
For $\nu_\Y$-a.e.\ $y$, by Jensen's inequality,~\Cref{lipshitzness} and~\Cref{embedding} in~\Cref{Assumptions},
\begin{align*}
\left\|g_\T(y)-g_\S(y)\right\|_\mathcal{K}&\leq\int\left\|K(\T(y,v), \cdot)-K(\S(y,v),\cdot)\right\|_\mathcal{K}\eta_\V(\dd v)\\
&\leq L_K\left\|\T(y,\cdot)-\S(y,\cdot)\right\|_{L^1(\V,\U;\eta_\V)}\leq L_KC_{1,\eta_\V}\left\|\T(y,\cdot)-\S(y,\cdot)\right\|_{\HH}.
\end{align*}
Taking the essential supremum over $y$ yields the first claim. The same calculation applies to $\T^\dag$, for which $g_{\T^\dag}=g^\dag$. Hence, by~\Cref{lem: app lifted problem}, \begin{align*}\DD(\nu,\T)=\|g_\T-g^\dag\|_2^2\leq\|g_\T-g_{\T^\dagger}\|_\infty^2\leq(L_KC_{1,\eta_\V})^2\|\T-\T^\dagger\|_\infty^2,\end{align*}
for every $\T\in\TT$. Minimizing over $\T\in\TT$ gives the second claim.
\end{proof}

\subsection{Preliminary results from empirical process theory}\label{subsec: preliminary results from empirical process theory}
In this section we summarize the main tools from empirical process theory that we employ in the proof of~\Cref{thm: main error bound 2}. The first one that we employ is the \emph{one-shot localization technique} attributed to van de Geer \cite{van1987new,van2002m}. We present a small variation of the version of this method reported in~\cite{divol2025optimal-sm}. The variation is adapted to account for bounds on approximate minimizers $a^*\in\A$ such that
\begin{align}\label{eq: approximate minimizer defn}
R^m(a^*)\leq R^m(\overline{a})+ \varrho,
\end{align}
for some $\varrho\geq0$ and any fixed anchor element $\overline a$. As explained more in detail in the later sections, we let $\overline{a} = \widehat{g} = g_{\Th}$. While we will let $a^* = g_{\Th^m}$ and $\varrho = 0$ to derive the error bound of $\Th^m$, this adaptation will be useful for the proof of the fully empirical and practical minimizer $\Th^{m,n}$ where we instead set $a^* = g_{\Th^{m,n}}$ and control the fluctuations due to the sampling error in approximating the conditional mean embedding integral for $\varrho >0$.

\begin{theorem}[One-shot localization, adapted from~{\cite[Thm.~A.1]{divol2025optimal-sm}}]\label{thm: one-shot localization}
Let $\xi_1,\dots,\xi_m$ be i.i.d.\ with law $P$ on a measurable space $\Xi$, let $(\mathcal{A},\|\cdot\|)$ be a pseudo-normed vector space, and let $\phi:\mathcal{A}\times\Xi\to(-\infty,+\infty]$ be such that $\xi\mapsto\phi(a,\xi)$ is measurable for every $a\in\mathcal{A}$ and $a\mapsto\phi(a,\xi)$ is convex for $P$-a.e.\ $\xi$. Assume that $R(a)\coloneqq\int\phi(a,\xi)P(\dd\xi)$ admits a minimizer $a^\dag$ over $\mathcal{A}$, and set $\ell(a)\coloneqq R(a)-R(a^\dag)$ and $R^m(a)\coloneqq\frac1m\sum_{i=1}^m\phi(a,\xi_i)$. Assume that there is a nondecreasing $w:[0,\infty)\to[0,\infty)$ such that $\|a-a^\dag\|\leq w(\ell(a))$ for all $a\in\mathcal{A}$. Let $A\subseteq\mathcal{A}$ be a model, let $|||\cdot|||$ be another pseudo-norm on $\mathcal{A}$ with $\sup_{a\in A}|||a|||\leq B$ for some $B>0$, let $0<r\leq B$ and $\bar a\in A$, and let
\begin{align}\label{eq: contrained cond with auxiliary norm}
\bar A\coloneqq\left\{\bar a+s(a-\bar a)\mid a\in A,\ s\in[0,1],\ s|||a-\bar a|||\leq r\right\}.
\end{align}
Let $F_{\text{env.}}$ be a measurable envelope with $|\phi(a,\xi)|\leq F_{\text{env.}}(\xi)$ for all $a\in\bar A$ and $\xi\in\Xi$, such that $F_{\text{env.}}(\xi)$, $\xi\sim P$, belongs to the Orlicz space of index $p\in(0,1]$ with Orlicz norm at most $L_{\rm env}$. Let $\varrho\geq0$ and $t>0$, and let $\tau>0$ and $\Psi(\tau)\geq0$ be such that
\begin{align}\label{eq: Psi bound}
\sqrt m\,\mathbb{E}\left[\sup_{a\in\bar A,\ \|a-\bar a\|\leq\tau}\left|(R^m-R)(a)-(R^m-R)(\bar a)\right|\right]\leq\Psi(\tau)
\end{align}
and
\begin{align}\label{eq: app localization condition rho}
\tau\geq 4\,w\left(\ell(\bar a)+\varrho+\kappa_p\,t^{1/p}\left(\frac{\Psi(\tau)}{\sqrt m}+\frac{L_{\rm env}(\log m)^p}{m}\right)\right),
\end{align}
where $\kappa_p>0$ is a constant depending only on $p$. Then there is an event $\Omega$, depending only on $\xi_1,\dots,\xi_m$ and of probability at least $1-e^{-t}$, on which every $a^*\in A$ with $R^m(a^*)\leq R^m(\bar a)+\varrho$ (i.e., as in~\eqref{eq: approximate minimizer defn}) satisfies
\begin{align*}
\|a^*-a^\dag\|\leq w(\ell(\bar a))+\left(\frac{2B}{r}-1\right)\tau.
\end{align*}
In particular, for $\varrho=0$, this holds for every $a^*\in\argmin_A R^m$.
\end{theorem}

\begin{proof}[Proof of~\Cref{thm: one-shot localization}]
We follow almost identically the proof of~\cite[Thm.~A.1]{divol2025optimal-sm}, which corresponds to the case $\varrho=0$ and $a^*\in\argmin_AR^m$. Let $t>0$ and let $\tau$ be as in the statement of the theorem. Fix $a^*\in A$ with $R^m(a^*)\leq R^m(\bar a)+\varrho$, and let $a^*_s=sa^*+(1-s)\bar a$, where $s=\frac{\tau}{\tau+\|a^*-\bar a\|}$. Note that
\begin{align*}
\|a^*_s-\bar a\|=s\|a^*-\bar a\|=\frac{\tau\|a^*-\bar a\|}{\tau+\|a^*-\bar a\|}\leq\tau.
\end{align*}
If $s\geq r/(2B)$, then $\|a^*-\bar a\|\leq\left(\frac{2B}{r}-1\right)\tau$, and the conclusion holds. Otherwise, we have $s<r/(2B)$, which implies that
\begin{align}\label{eq: app hull membership}
|||a^*_s-\bar a|||=s|||a^*-\bar a|||\leq r,
\end{align}
since $|||a^*-\bar a|||\leq 2B$. Hence $a^*_s\in\bar A$ and $\|a^*_s-\bar a\|\leq\tau$. On the almost sure event on which $a\mapsto\phi(a,\xi_i)$ is convex for every $i\leq m$, and since $R^m(a^*)\leq R^m(\bar a)+\varrho$, it holds that
\begin{align*}
R^m(a^*_s)\leq sR^m(a^*)+(1-s)R^m(\bar a)\leq R^m(\bar a)+s\varrho\leq R^m(\bar a)+\varrho.
\end{align*}
Hence,
\begin{align*}
\ell(a^*_s)&=\ell(\bar a)+R^m(a^*_s)-R^m(\bar a)+(R^m-R)(\bar a)-(R^m-R)(a^*_s)\\
&\leq\ell(\bar a)+\varrho+(R^m-R)(\bar a)-(R^m-R)(a^*_s).
\end{align*}
Let
\begin{align*}
Z_m=\sup_{a\in\bar A,\ \|a-\bar a\|\leq\tau}\left|(R^m-R)(a)-(R^m-R)(\bar a)\right|,
\end{align*}
so that $(R^m-R)(\bar a)-(R^m-R)(a^*_s)\leq Z_m$. As $w$ is nondecreasing, we have
\begin{align*}
\frac{\tau\|a^*-\bar a\|}{\tau+\|a^*-\bar a\|}=\|a^*_s-\bar a\|&\leq\|a^*_s-a^\dag\|+\|\bar a-a^\dag\|\\
&\leq w(\ell(a^*_s))+w(\ell(\bar a))\\
&\leq w\left(\ell(\bar a)+\varrho+(R^m-R)(\bar a)-(R^m-R)(a^*_s)\right)+w(\ell(\bar a))\\
&\leq 2\,w\left(\ell(\bar a)+\varrho+Z_m\right).
\end{align*}
Thus, if $w(\ell(\bar a)+\varrho+Z_m)\leq\tau/4$, then $\|a^*-\bar a\|\leq\tau$.

The quantity $Z_m$ does not depend on $a^*$. Accordingly, let $\Omega$ be the intersection of the almost sure event above with the event $\{w(\ell(\bar a)+\varrho+Z_m)\leq\tau/4\}$. Then $\Omega$ depends only on $\xi_1,\dots,\xi_m$, and the above argument applies simultaneously, on $\Omega$, to every $a^*\in A$ with $R^m(a^*)\leq R^m(\bar a)+\varrho$. Since $\mathbb{E}[Z_m]\leq\Psi(\tau)/\sqrt m$, according to~\cite[Thm.~2.14.23]{vanderVaartWellner2023} and~\eqref{eq: app localization condition rho},
\begin{align*}
\mathbb{P}(\Omega^c)&\leq\mathbb{P}\left(w\left(\ell(\bar a)+\varrho+Z_m\right)>\tau/4\right)\\
&\leq\mathbb{P}\left(Z_m\geq\kappa_p\,t^{1/p}\left(\frac{\Psi(\tau)}{\sqrt m}+\frac{(\log m)^pL_{\rm env}}{m}\right)\right)\\
&\leq\mathbb{P}\left(Z_m\geq\kappa_p\,t^{1/p}\left(\mathbb{E}[Z_m]+\frac{(\log m)^pL_{\rm env}}{m}\right)\right)\\
&\leq\exp(-t)
\end{align*}
for a certain constant $\kappa_p$. Hence, on $\Omega$, which has probability at least $1-e^{-t}$, for every $a^*\in A$ with $R^m(a^*)\leq R^m(\bar a)+\varrho$ the following holds. If $s<r/(2B)$, then
\begin{align*}
\|a^*-\bar a\|\leq\tau\leq\left(\frac{2B}{r}-1\right)\tau,
\end{align*}
and otherwise, if $s\geq r/(2B)$, we have already proved that $\|a^*-\bar a\|\leq\left(\frac{2B}{r}-1\right)\tau$. Hence, by the triangle inequality, on $\Omega$, for every such $a^*$,
\begin{align*}
\|a^*-a^\dag\|&\leq\|\bar a-a^\dag\|+\|a^*-\bar a\|\\
&\leq w(\ell(\bar a))+\left(\frac{2B}{r}-1\right)\tau.
\end{align*}
\end{proof}

\begin{remark}\label{rem: on the localization statement}
In the proof of~\Cref{thm: main error bound 2}, we apply~\Cref{thm: one-shot localization} to a loss that is uniformly bounded under~\Cref{Assumptions}, so that the envelope $F_{\text{env.}}$ can be taken constant. Since constant random variables belong to every Orlicz space, the Orlicz norm condition on $F_{\text{env.}}$ is then trivially satisfied, e.g., with $p=1$. We nevertheless state~\Cref{thm: one-shot localization} with a general envelope, since this is the version that allows the uniform boundedness and Lipschitz conditions of~\Cref{Assumptions} and~\Cref{assum: strengthened assum} to be relaxed.
\end{remark}

The next important tool from empirical process theory that we employ concerns the estimation of Dudley integrals crucial in identifying the function $\Psi$ reported in~\eqref{eq: Psi bound} in~\Cref{thm: one-shot localization}. We state it without proof from~\cite[Prop. A.2.]{divol2025optimal-sm}. The result is an extension to uniformly weighted bounded hypothesis classes of the original result by~\cite[Thm. 2.14.21]{vanderVaartWellner2023} which instead concerns uniformly unweighted bounded classes.

\begin{proposition}[Dudley-type bound {\cite[Prop.~A.2]{divol2025optimal-sm}}]\label{prop: dudley general}
Let $P$ be a probability distribution on $\RR^D$ such that $\xi\sim P$ belongs to the Orlicz space of index $p\in(0,1]$, and let $P_m$ be the empirical measure of $m$ i.i.d.\ samples from $P$. For $x\in\RR^D$, let $\langle x\rangle\coloneqq1+\|x\|$, and, for $q\geq0$, define the weighted uniform norm of a function $g:\textnormal{supp}(P)\to\RR$ by
\begin{align*}
\|g\|_{L^\infty(\langle\cdot\rangle^{-q})}\coloneqq\sup_{x\in\textnormal{supp}(P)}\langle x\rangle^{-q}|g(x)|.
\end{align*}
Let $\mathcal{G}$ be a class of functions defined on $\textnormal{supp}(P)$ with
\begin{align*}
\sup_{g\in\mathcal{G}}\|g\|_{L^2(P)}\leq\varsigma\qquad\text{and}\qquad\sup_{g\in\mathcal{G}}\|g\|_{L^\infty(\langle\cdot\rangle^{-q})}\leq \Pi,
\end{align*}
for some $q\geq0$. Let $H_2$ (resp.\ $H_\infty$) be a decreasing function that diverges at $\delta=0$ and such that, for all $\delta>0$,
\begin{align*}
\Hf\left(\delta,\mathcal{G},L^2(P)\right)\leq H_2(\delta)\qquad\left(\text{resp. }\Hf\left(\delta,\mathcal{G},L^\infty(\langle\cdot\rangle^{-q})\right)\leq H_\infty(\delta)\right),
\end{align*}
with $\Hf$ the metric entropy as in~\eqref{eq: eps-entropy}, for a general class and pseudo-norm. There exists $C>0$ depending on $q$, $p$, and the $p$-Orlicz norm of $\xi\sim P$ such that the following holds. For $\varepsilon\in(0,\Pi)$, let $L\coloneqq(C\log_+(1/\varepsilon))^{1/p}$, where $\log_+(x)\coloneqq\max\{\log x,0\}$, let $\kappa_\infty>0$ be such that $H_\infty(\kappa_\infty\varepsilon)\leq H_\infty(\varepsilon)/2$, and let $\tilde\varepsilon<\varsigma$ be such that $H_2(\tilde\varepsilon)\geq H_\infty(\varepsilon)$ (with possibly $\tilde\varepsilon=0$). Then
\begin{align}\label{eq: dudley general}
\sqrt m\,\mathbb{E}\left[\sup_{g\in\mathcal{G}}\left|(P_m-P)(g)\right|\right]\lesssim\sqrt m\,(\kappa_\infty+\Pi)\,\varepsilon\,\langle L\rangle^q+\int_{\tilde\varepsilon/2}^\varsigma\sqrt{H_2(\delta)}\,\dd \delta+\frac{\langle L\rangle^q}{\sqrt m}\int_{\varepsilon/2}^\Pi H_\infty(\delta)\,\dd \delta,
\end{align}
up to a numerical constant. The same inequality holds for $\varepsilon=0$ with $L=0$.
\end{proposition}

\begin{remark}\label{rem: on the dudley statement}
Under~\Cref{Assumptions} and~\Cref{assum: strengthened assum}, all function classes to which we apply~\Cref{prop: dudley general} are uniformly bounded. We therefore only use it with $\varepsilon=\tilde\varepsilon=0$ (so that $L=0$) and $q=0$, in which case~\eqref{eq: dudley general} reduces to
\begin{align}\label{eq: actual Dudley bound used}
\sqrt m\,\mathbb{E}\left[\sup_{g\in\mathcal{G}}\left|(P_m-P)(g)\right|\right]\lesssim\int_0^\varsigma\sqrt{H_2(\delta)}\,\dd \delta+\frac{1}{\sqrt m}\int_0^\Pi H_\infty(\delta)\,\dd \delta.
\end{align}
This is the classical bound of~\cite[Thm.~2.14.21]{vanderVaartWellner2023}. In this case the Orlicz norm condition on $\xi\sim P$ is not needed. Moreover, the weight $\langle\cdot\rangle$ plays no role, and the proof of~\eqref{eq: actual Dudley bound used} (see~\cite[Thm.~2.14.21]{vanderVaartWellner2023} or the proof of~\cite[Prop.~A.2]{divol2025optimal-sm}) relies only on Bernstein's inequality and chaining. Hence,~\eqref{eq: actual Dudley bound used} holds for $P$ on an arbitrary measurable space, in particular on $\Y\times\U$ in~\Cref{prop: app local fluctuation} and on $\Y\times\U\times\V^n$ in~\Cref{prop: app reference error}. As for~\Cref{thm: one-shot localization}, we nevertheless state~\Cref{prop: dudley general} in its general form, since this is the version that allows the boundedness conditions of~\Cref{Assumptions} and~\Cref{assum: strengthened assum} to be relaxed, at least for finite-dimensional $\Y$, $\U$ and $\V$, where~\Cref{prop: dudley general} applies as stated.
\end{remark}

The third tool we use is Bousquet version~\cite[Thm. 2.3]{bousquet2002bennett} of Talagrand's inequality, which we state in the following rescaled form.

\begin{lemma}[Concentration of suprema, rescaled form of {\cite[Thm. 2.3]{bousquet2002bennett}}]\label{lem: bousquet}
Let $P$ be a probability measure on a measurable space, let $P_m$ be the empirical measure of $m$ i.i.d.\ samples from $P$, and let $\mathcal{G}$ be a class of functions with $\sup_{f\in\mathcal{G}}\|f\|_{L^2(P)}\leq\varsigma$, $\sup_{f\in\mathcal{G}}\|f\|_{\infty}\leq\Pi$ and $Pf=0$ for all $f\in\mathcal{G}$. Let $Z\coloneqq\sup_{f\in\mathcal{G}}(P_m-P)f$. Then, for every $t>0$, with probability at least $1-e^{-t}$,
\begin{align*}
Z\leq 2\,\mathbb{E}Z+\varsigma\sqrt{\frac{2t}{m}}+\frac{4\Pi t}{3m}.
\end{align*}
\end{lemma}

\begin{proof}[Proof of~\Cref{lem: bousquet}]
For $f\in\mathcal{G}$, let $\tilde f\coloneqq f/\Pi$. Then $\mathbb{E}[\tilde f(Z_1)]=0$, $\operatorname*{ess\,sup}\tilde f\leq1$, and
\begin{align*}
\textnormal{Var}[\tilde f(Z_1)]=P\tilde f^2=\frac{\|f\|^2_{L^2(P)}}{\Pi^2}\leq\frac{\varsigma^2}{\Pi^2}.
\end{align*}
Let
\begin{align*}
\widetilde Z\coloneqq\sup_{f\in\mathcal{G}}\sum_{i=1}^m\tilde f(Z_i)=\frac m\Pi\sup_{f\in\mathcal{G}}P_mf=\frac m\Pi\,Z,
\end{align*}
where we used $Pf=0$, and note that $\mathbb{E}[\widetilde Z]\geq\sup_{f\in\mathcal{G}}\mathbb{E}\big[\sum_{i=1}^m\tilde f(Z_i)\big]=0$. By~\cite[Thm.~2.3]{bousquet2002bennett}, applied with $n=m$, $\sigma=\varsigma/\Pi$ and $x=t$, with probability at least $1-e^{-t}$,
\begin{align*}
\widetilde Z\leq\mathbb{E}[\widetilde Z]+\sqrt{2t\tilde v}+\frac t3,\qquad\text{where}\qquad\tilde v\coloneqq m\frac{\varsigma^2}{\Pi^2}+2\frac m\Pi\,\mathbb{E}[Z].
\end{align*}
Multiplying by $\Pi/m$ and using $\frac \Pi m\sqrt{2t\tilde v}=\sqrt{2t(\varsigma^2+2\Pi\mathbb{E}[Z])/m}$, we obtain, with probability at least $1-e^{-t}$,
\begin{align*}
Z\leq\mathbb{E}[Z]+\sqrt{\frac{2t\left(\varsigma^2+2\Pi\,\mathbb{E}[Z]\right)}{m}}+\frac{\Pi t}{3m}.
\end{align*}
Finally, by $\sqrt{a+b}\leq\sqrt a+\sqrt b$ and $2\sqrt{cd}\leq c+d$ with $c=\mathbb{E}[Z]$ and $d=t\Pi/m$,
\begin{align*}
\sqrt{\frac{2t\left(\varsigma^2+2\Pi\,\mathbb{E}[Z]\right)}{m}}\leq\varsigma\sqrt{\frac{2t}{m}}+2\sqrt{\mathbb{E}[Z]\frac{t\Pi}{m}}\leq\varsigma\sqrt{\frac{2t}{m}}+\mathbb{E}[Z]+\frac{\Pi t}{m},
\end{align*}
which concludes the proof since $\frac{\Pi t}{m}+\frac{\Pi t}{3m}=\frac{4\Pi t}{3m}$.
\end{proof}

We later adopt this result to control the fluctuations induced by the sampling error in approximating the integral of the conditional mean embedding to obtain the estimator $\Th^{m,n}$.

Lastly, we summarize explicit integral estimation calculations that will be used after the application of~\Cref{prop: dudley general} in order to explicitate the function $\Psi$ appearing in~\eqref{eq: Psi bound} of~\Cref{thm: one-shot localization} in the presented error bounds for $\Th^m$ and $\Th^{m,n}$.

\begin{lemma}[Entropy integral bounds]\label{lem: app entropy integrals}
Let $C>0$, $\gamma\in[0,1)$ and $A,D,\lambda>0$. Then
\begin{enumerate}
\item\label{calc0}$\log(e+\lambda \delta^{-1})\leq 2\log(e+\lambda)\log(e+\delta^{-1}).$
\item\label{calc1}$\int_0^A\sqrt{C\delta^{-\gamma}\log(e+\delta^{-1})}\,\dd \delta\leq 6\sqrt{C}A^{1-\gamma/2}\sqrt{\log(e+A^{-1})}.$
\item\label{calc2}$\int_0^A C\delta^{-\gamma}\log(e+\delta^{-1})\,\dd \delta\leq\frac{4CA^{1-\gamma}\log(e+A^{-1})}{(1-\gamma)^2}.$
\item\label{calc3}$\int_0^A\sqrt{\log(e+D\delta^{-1})}\,\dd \delta\leq 2A\sqrt{\log(e+DA^{-1})}.$
\item\label{calc4}$\int_0^A\log(e+D\delta^{-1})\,\dd \delta\leq 2A\log(e+DA^{-1}).$
\end{enumerate}
\end{lemma}
\begin{proof}
\Cref{calc0} follows since $e+\lambda/\delta\leq(e+\lambda)(e+1/\delta)$, and $x+y\leq 2xy$ for $x,y\geq1$.

For the remaining items, we substitute $\delta=Aq$ with $q\in(0,1]$. For~\Cref{calc1,calc2}, we follow this substitution with the bound $\log(e+(Aq)^{-1})\leq\log(e+A^{-1})+\log(e+q^{-1})$. This simply follows from $e+(Aq)^{-1}\leq(e+A^{-1})(e+q^{-1})$, together with $\sqrt{x+y}\leq\sqrt x+\sqrt y$ and $\log(e+A^{-1})\geq1$.

Specifically for~\Cref{calc1}, we further employ $\int_0^1q^{-\gamma/2}\dd q\leq2$, $q^{-\gamma/2}\leq q^{-1/2}$ and $e+q^{-1}\leq(e+1)/q$, the substitution $q=e^{-s}$, and Jensen's inequality to give
\begin{align*}
\int_0^1q^{-\gamma/2}\sqrt{\log(e+q^{-1})}\,\dd q\leq\int_0^\infty e^{-s/2}\sqrt{\log(e+1)+s}\,\dd s\leq 2\sqrt{\log(e+1)+2}\leq4.
\end{align*}
Combining everything yields
\begin{align*}
\int_0^A\sqrt{\delta^{-\gamma}\log(e+\delta^{-1})}\,\dd \delta&\leq A^{1-\frac{\gamma}{2}}\left(\sqrt{\log(e+A^{-1})}\int_0^1 q^{-\frac\gamma2} \dd q+\int_0^1q^{-\frac\gamma2}\sqrt{\log(e+q^{-1})}\,\dd q\right)\\
&\leq A^{1-\frac\gamma2}\left(2\sqrt{\log(e+A^{-1})} + 4\right)\leq 6A^{1-\frac\gamma2}\sqrt{\log(e+A^{-1})},
\end{align*}
which is~\Cref{calc1} multiplying by $\sqrt{C}$.

Specifically for~\Cref{calc2}, we further use $\int_0^1q^{-\gamma}\dd q=(1-\gamma)^{-1}$ and $\int_{0}^{1}q^{-\gamma}\log(q^{-1})\dd q
      =\frac{1}{(1-\gamma)^{2}}$ to give
\begin{align*}
\int_0^1 q^{-\gamma}\log(e+q^{-1})\,\dd q&\leq \log(e+1)\int_0^1q^{-\gamma}\dd q+ \int_0^1 q^{-\gamma}\log(q^{-1})\dd q\\
& = \frac{\log(e+1)}{1-\gamma} + (1-\gamma)^{-2}.
\end{align*}
Combining everything yields
\begin{align*}
\int_0^A \delta^{-\gamma}\log(e+\delta^{-1})\,\dd \delta&\leq A^{1-\gamma}\left(\log(e+A^{-1})\int_0^1 q^{-\gamma} \dd q+\int_0^1 q^{-\gamma}\log(e+q^{-1})\,\dd q\right)\\
&\leq A^{1-\gamma}\left(\frac{\log(e+A^{-1})+\log(e+1)}{1-\gamma}+(1-\gamma)^{-2}\right)\\
&\leq 4A^{1-\gamma}\frac{\log(e+A^{-1})}{(1-\gamma)^2},
\end{align*}
which is~\Cref{calc2} multiplying by $C$.

Specifically for~\Cref{calc3}, we employ $e+D(Aq)^{-1}\leq(e+DA^{-1})/q$ together with the identity $\int_0^1\sqrt{\log(q^{-1})}\,\dd q=\sqrt\pi/2\leq1$ to give
\begin{align*}
\int_0^A\sqrt{\log(e+D\delta^{-1})}\,\dd \delta&\leq A\left(\sqrt{\log(e+DA^{-1})} + \int_0^1\sqrt{\log(q^{-1})}\dd q\right)\\
&\leq A\left(\sqrt{\log(e+DA^{-1})}+1\right)\leq 2A\sqrt{\log(e+DA^{-1})}.
\end{align*}
Lastly, \Cref{calc4} is analogous to~\Cref{calc3} but without square root and using $\int_0^1\log(q^{-1})\,\dd q=1$.
\end{proof}

\subsection{Local complexity control of the lifted class}\label{subsec: local complexity control of the lifted class}
After introducing the regression setup of~\Cref{subsec: lifted regression problem over CMEs} and the empirical process theory background of~\Cref{subsec: preliminary results from empirical process theory}, we can finally directly tackle the proof of~\Cref{thm: main error bound 2}. We begin doing so by controlling the complexity of a class which is localized around the conditional mean embedding corresponding to $\Th$, i.e., $\widehat{g}$. To this end, we introduce the segment hull of $\G_\TT$ around $\widehat{g}$, namely,
\begin{align}\label{eq: segment hull}
\widehat {\G}_\TT\coloneqq\{\widehat  g+s(g-\widehat{ g})\mid g\in\G_\TT,\ s\in[0,1]\},
\end{align}
and the localized loss class
\begin{align}\label{eq: localized loss class}
\mathcal{L}(\tau)\coloneqq\left\{\phi(h,\cdot)-\phi(\widehat g,\cdot)\mid h\in\widehat \G_\TT,\ \|h-\widehat g\|_2\leq\tau\right\}.
\end{align}

The goal of this section is then the one to control the complexity of $\mathcal{L}(\tau)$. For doing so, we first recap some preliminary bounds yielded by the boundedness of $K$ that we use throughout the rest of the section.

\begin{lemma}[Segment hull, envelope and Lipschitz loss]\label{lem: app hull}
Suppose that~\Cref{Boundedness kernel} in~\Cref{Assumptions} holds. Then the following hold.
\begin{enumerate}
\item $\|g-\widehat g\|_\infty\leq 2B_K$ for every $g\in\G_\TT$, and $\|h\|_\infty\leq B_K$ for every $h\in\widehat\G_\TT$.
\item $0\leq\phi(h,\xi)\leq 4B_K^2$ for every $h\in\widehat\G_\TT$ and every $\xi\in\Y\times\U$.
\item For all $h,h'\in\widehat\G_\TT$ and $\xi=(y,u)\in\Y\times\U$,
\begin{align*}
\left|\phi(h,\xi)-\phi(h',\xi)\right|\leq 4B_K\left\|h(y)-h'(y)\right\|_\mathcal{K}.
\end{align*}
In particular,
\begin{align*}
\left\|\phi(h,\cdot)-\phi(h',\cdot)\right\|_{L^2(\nu)}&\leq 4B_K\|h-h'\|_2,\\
\left\|\phi(h,\cdot)-\phi(h',\cdot)\right\|_{L^\infty(\nu)}&\leq 4B_K\|h-h'\|_\infty.
\end{align*}
\end{enumerate}
\end{lemma}

\begin{proof}
By~\Cref{lem: app lifted problem}, $\|g(y)\|_\mathcal{K}\leq B_K$ for every $g\in\G_\TT$ and $y\in\Y$. Since $\widehat g\in\G_\TT$, this gives $\|g-\widehat g\|_\infty\leq 2B_K$ for every $g\in\G_\TT$. Moreover, every $h\in\widehat\G_\TT$ is of the form $h=(1-s)\widehat g+sg$ with $g\in\G_\TT$ and $s\in[0,1]$, so that $\|h(y)\|_\mathcal{K}\leq(1-s)\|\widehat g(y)\|_\mathcal{K}+s\|g(y)\|_\mathcal{K}\leq B_K$ for every $y\in\Y$. 

By~\Cref{Boundedness kernel} in~\Cref{Assumptions}, $\|K(u,\cdot)\|_\mathcal{K}=\sqrt{K(u,u)}\leq B_K$. Hence, $\|K(u,\cdot)-h(y)\|_\mathcal{K}\leq 2B_K$ for every $h\in\widehat\G_\TT$, and $0\leq\phi(h,\xi)\leq 4B_K^2$ follows from the definition of $\phi$.

For $a,b\in\mathcal{K}$ and $u\in\U$, expanding the squares gives
\begin{align*}
\left\|K(u,\cdot)-a\right\|^2_\mathcal{K}-\left\|K(u,\cdot)-b\right\|^2_\mathcal{K}=\left\langle a-b,\,a+b-2K(u,\cdot)\right\rangle_\mathcal{K}.
\end{align*}
Take $a=h(y)$ and $b=h'(y)$. By~\Cref{Boundedness kernel} in~\Cref{Assumptions}, $\|a+b-2K(u,\cdot)\|_\mathcal{K}\leq 4B_K$, so the Cauchy-Schwarz inequality yields the pointwise bound $
\left|\phi(h,\xi)-\phi(h',\xi)\right|\leq 4B_K\left\|h(y)-h'(y)\right\|_\mathcal{K}$. Squaring and integrating against $\nu$, and noting that the right-hand side depends on $\xi=(y,u)$ only through $y$, gives
\begin{align*}
\left\|\phi(h,\cdot)-\phi(h',\cdot)\right\|^2_{L^2(\nu)}\leq 16B_K^2\int_\Y\left\|h(y)-h'(y)\right\|^2_\mathcal{K}\nu_\Y(\dd y)=16B_K^2\|h-h'\|_2^2.
\end{align*}
Similarly, the pointwise bound gives $|\phi(h,\xi)-\phi(h',\xi)|\leq 4B_K\|h-h'\|_\infty$ for $\nu$-a.e.\ $\xi$, which is the $L^\infty(\nu)$ bound.
\end{proof}

Now, we bound the entropy of the localized loss class.

\begin{lemma}[Entropy of the localized loss class]\label{lem: app loss entropy}
Suppose that~\Cref{Boundedness kernel} and~\Cref{Entropy lifted} in~\Cref{Assumptions} hold, and let $C_\mathcal{L}$ be as in~\Cref{thm: main error bound 2}. Then, for every $\tau>0$,
\begin{align*}
\sup_{f\in\mathcal{L}(\tau)}\|f\|_{L^2(\nu)}\leq 4B_K\tau,\qquad\sup_{f\in\mathcal{L}(\tau)}\|f\|_{L^\infty(\nu)}\leq 8B_K^2,
\end{align*}
and, for every $\delta>0$,
\begin{align*}
\Hf\left(\delta,\mathcal{L}(\tau),L^2(\nu)\right)\leq H(\delta)\qquad\text{and}\qquad\Hf\left(\delta,\mathcal{L}(\tau),L^\infty(\nu)\right)\leq H(\delta),
\end{align*}
where
\begin{align}\label{eq: entropy envolpe H}
H(\delta)\coloneqq C_\mathcal{L}\delta^{-\gamma}\log(e+\delta^{-1})+2\log\left(e+4B_K^2\delta^{-1}\right).
\end{align}
\end{lemma}

\begin{proof}
Let $h\in\widehat\G_\TT$, so that $h=\widehat g+s(g-\widehat g)$ with $g\in\G_\TT$ and $s\in[0,1]$. By, Item 1 of~\Cref{lem: app hull}, $\|h-\widehat g\|_\infty=s\|g-\widehat g\|_\infty\leq 2B_K$. Then, using Item 3 of~\Cref{lem: app hull} with $h'=\widehat g$, we have that, whenever $\|h-\widehat g\|_2\leq\tau$,
\begin{align*}
\left\|\phi(h,\cdot)-\phi(\widehat g,\cdot)\right\|_{L^2(\nu)}&\leq 4B_K\|h-\widehat g\|_2\leq 4B_K\tau,\\
\left\|\phi(h,\cdot)-\phi(\widehat g,\cdot)\right\|_{L^\infty(\nu)}&\leq 4B_K\|h-\widehat g\|_\infty\leq 8B_K^2.
\end{align*}

Next, fix $\varepsilon>0$. Let $\{g_k\}\subseteq\G_\TT$ be an $\varepsilon/2$-net of $\G_\TT$ in $\|\cdot\|_\infty$ with at most $\N(\varepsilon/2,\G_\TT,\|\cdot\|_\infty)$ elements, and let $\{s_j\}\subseteq[0,1]$ be a grid with spacing $\varepsilon/(2B_K)$ and at most $2+2B_K/\varepsilon$ points. Then, given $h=\widehat g+s(g-\widehat g)\in\widehat\G_\TT$, we have that $\|g-g_k\|_\infty\leq\varepsilon/2$ and $s_j$ with $|s-s_j|\leq\varepsilon/(4B_K)$. Consequently, using Item 1 of~\Cref{lem: app hull},
\begin{align*}
\left\|h-\left(\widehat g+s_j(g_k-\widehat g)\right)\right\|_\infty\leq s\|g-g_k\|_\infty+|s-s_j|\,\|g_k-\widehat g\|_\infty\leq\frac\varepsilon2+\frac{\varepsilon}{4B_K}2B_K=\varepsilon.
\end{align*}
Hence $\{\widehat g+s_j(g_k-\widehat g)\}$ (which is contained in $\widehat\G_\TT$ by definition) consists of an $\varepsilon$-net of $\widehat\G_\TT$ in $\|\cdot\|_\infty$ and has cardinality at most $\N(\varepsilon/2,\G_\TT,\|\cdot\|_\infty)\,(2+2B_K/\varepsilon)$.

The construction we just outlined tells us that, for a fixed $\delta>0$, we can find $\{h_l\}\subseteq\widehat\G_\TT$ constituting an $\varepsilon = \delta/(4B_K)$-net of $\widehat\G_\TT$ in $\|\cdot\|_\infty$ as above. By Item 3 of~\Cref{lem: app hull} and $\|\cdot\|_2\leq\|\cdot\|_\infty$, the corresponding functions $\{\phi(h_l,\cdot)-\phi(\widehat g,\cdot)\}$ then form a $\delta$-net of $\mathcal{L}(\tau)$, both in $L^2(\nu)$ and in $L^\infty(\nu)$ of cardinality at most $\N(\delta/(8B_K),\G_\TT,\|\cdot\|_\infty)\,(2+8B_K^2/\delta)$. Since $\Hf=\log\N$, we then have
\begin{align*}
\max\left\{\Hf\left(\delta,\mathcal{L}(\tau),L^2(\nu)\right),\Hf\left(\delta,\mathcal{L}(\tau),L^\infty(\nu)\right)\right\}\leq\Hf\left(\frac{\delta}{8B_K},\G_\TT,\|\cdot\|_\infty\right)+\log\left(2+\frac{8B_K^2}{\delta}\right).
\end{align*}
By~\Cref{Entropy lifted} in~\Cref{Assumptions}, $\log(1+x)\leq\log(e+x)$ and~\Cref{calc0} of~\Cref{lem: app entropy integrals},
\begin{align*}
\Hf\left(\frac{\delta}{8B_K},\G_\TT,\|\cdot\|_\infty\right)&\leq C_{\G_\TT}(8B_K)^\gamma \delta^{-\gamma}\log\left(e+8B_K\delta^{-1}\right)\\
&\leq 2C_{\G_\TT}(8B_K)^\gamma\log(e+8B_K)\,\delta^{-\gamma}\log(e+\delta^{-1})=C_\mathcal{L}\delta^{-\gamma}\log(e+\delta^{-1}).
\end{align*}
Finally, $\log(2+8B_K^2/\delta)\leq 2\log(e+4B_K^2/\delta)$, since $2+2x\leq(e+x)^2$ for every $x\geq0$.
\end{proof}
 Next, we control the fluctuations between the empirical and population loss differences for the elements of $\widehat \G_\TT$ around $\widehat{g}$. This result essentially makes explicit the function which we substitute as $\Psi$ in~\eqref{eq: Psi bound} of~\Cref{thm: one-shot localization} for deriving the error bounds of the estimators $\Th^m$ and $\Th^{m,n}$.

 \begin{proposition}[Local fluctuation bound]\label{prop: app local fluctuation}
Suppose that~\Cref{Boundedness kernel} and~\Cref{Entropy lifted} of~\Cref{Assumptions} hold. Then, for every $\tau>0$,
\begin{align*}
\sqrt m\,\mathbb{E}\left[\sup_{h\in\widehat\G_\TT,\ \|h-\widehat g\|_2\leq\tau}\left|(R^m-R)(h)-(R^m-R)(\widehat g)\right|\right]\leq\Psi(\tau),
\end{align*}
where
\begin{align}\label{eq: explicit psi}
\Psi(\tau)\coloneqq c_\Psi\left(\sqrt{\C_1}\tau^{1-\frac\gamma2}\sqrt{\log(e+\tau^{-1})}+\mathsf{a}\tau\sqrt{\log(e+\tau^{-1})}+\frac{\C_2}{\sqrt m}\right),
\end{align}
$c_\Psi\geq1$ is a numerical constant, we shorten $\mathsf{a}\coloneqq B_K\sqrt{\log(e+B_K)}$, and where we recall
\begin{align*}
\C_1\coloneqq C_\mathcal{L}\,B_K^{2-\gamma},\qquad \C_2\coloneqq\frac{C_\mathcal{L}B_K^2}{(1-\gamma)^2},
\end{align*}
as in the statement of~\Cref{thm: main error bound 2}.
\end{proposition}

\begin{proof}
For every $h\in\widehat\G_\TT$, we have $(R^m-R)(h)-(R^m-R)(\widehat g)=(\nu^m-\nu)\big(\phi(h,\cdot)-\phi(\widehat g,\cdot)\big)$. Hence the supremum in the statement equals $\sup_{f\in\mathcal{L}(\tau)}|(\nu^m-\nu)f|$. By~\Cref{lem: app loss entropy}, the class $\mathcal{L}(\tau)$ satisfies the assumptions of~\Cref{prop: dudley general} with $P=\nu$, $q=0$, $\varsigma=4B_K\tau$, $\Pi=8B_K^2$ and $H_2=H_\infty=H$ in~\eqref{eq: entropy envolpe H}. Applying~\Cref{prop: dudley general} with $\varepsilon=\tilde\varepsilon=0$ and $L=0$, as discussed in~\Cref{rem: on the dudley statement} (i.e., using~\eqref{eq: actual Dudley bound used}), yields
\begin{align}\label{eq: integral bound using H}
\sqrt m\,\mathbb{E}\left[\sup_{f\in\mathcal{L}(\tau)}\left|(\nu^m-\nu)f\right|\right]\lesssim\int_0^{4B_K\tau}\sqrt{H(\delta)}\,\dd \delta+\frac{1}{\sqrt m}\int_0^{8B_K^2}H(\delta)\,\dd \delta.
\end{align}
To conclude the proof, we then just need to bound the two integrals of~\eqref{eq: integral bound using H} using the explicit computations of~\Cref{lem: app entropy integrals}.

\emph{First integral.} By~\Cref{calc1} in~\Cref{lem: app entropy integrals} with $A=4B_K\tau$, together with $(4B_K)^{1-\gamma/2}\leq4B_K^{1-\gamma/2}$ and $\log(e+(4B_K\tau)^{-1})\leq\log(e+\tau^{-1})$ since $B_K\geq1$,
\begin{align*}
\int_0^{4B_K\tau}\sqrt{C_\mathcal{L}\delta^{-\gamma}\log(e+\delta^{-1})}\,\dd \delta\leq 24\,\sqrt{\C_1}\tau^{1-\frac\gamma2}\sqrt{\log(e+\tau^{-1})}.
\end{align*}
By~\Cref{calc3} in~\Cref{lem: app entropy integrals} with $A=4B_K\tau$ and $D=4B_K^2$, followed by~\Cref{calc0} in~\Cref{lem: app entropy integrals},
\begin{align*}
\int_0^{4B_K\tau}\sqrt{2\log\left(e+4B_K^2\delta^{-1}\right)}\,\dd \delta\leq 8\sqrt2B_K\tau\sqrt{\log(e+B_K\tau^{-1})}\leq 16\,\mathsf{a}\tau\sqrt{\log(e+\tau^{-1})}.
\end{align*}
Thus, we can bound the first integral by
\begin{align}\label{eq: first integral bound of H}
\int_0^{4B_K\tau}\sqrt{H(\delta)}\,\dd \delta\leq 24\,\sqrt{\C_1}\tau^{1-\frac\gamma2}\sqrt{\log(e+\tau^{-1})} + 16\,\mathsf{a}\tau\sqrt{\log(e+\tau^{-1})}.
\end{align}

\emph{Second integral.} By~\Cref{calc2} in~\Cref{lem: app entropy integrals} with $A=8B_K^2$, using $(8B_K^2)^{1-\gamma}\leq 8B_K^2$ and $\log(e+(8B_K^2)^{-1})\leq\log(e+1/8)$,
\begin{align*}
\int_0^{8B_K^2}C_\mathcal{L}\delta^{-\gamma}\log(e+\delta^{-1})\,\dd \delta\leq\frac{34\,C_LB_K^2}{(1-\gamma)^2}=34\,\C_2.
\end{align*}
By~\Cref{calc4} in~\Cref{lem: app entropy integrals} with $A=8B_K^2$ and $D=4B_K^2$,
\begin{align*}
\int_0^{8B_K^2}2\log\left(e+4B_K^2\delta^{-1}\right)\dd \delta\leq 32B_K^2\log(e+1/2)\leq 38B_K^2\leq 19\,\C_2,
\end{align*}
where the last inequality uses $C_\mathcal{L}\geq2$. Thus, we can bound the second integral by
\begin{align}\label{eq: second integral bound of H}
\int_0^{8B_K^2}H(\delta)\,\dd \delta\leq (34+19)\C_2 =53\C_2.
\end{align}

Combining~\eqref{eq: first integral bound of H} and~\eqref{eq: second integral bound of H}, absorbing all numerical factors into $c_\Psi$, concludes the proof.
\end{proof}

So far, we have not selected the localization radius $\tau$ defining the localized loss class of~\eqref{eq: localized loss class}. The last step towards applying the \emph{one-shot localization} of~\Cref{thm: one-shot localization} is then to select a suitable $\tau$ for which the necessary condition~\eqref{eq: app localization condition rho} hold. The next result achieves so.

\begin{lemma}[$\tau$ selection]\label{lem: app fixed point}
Let $\kappa_1$ be the constant of~\Cref{thm: one-shot localization} for $p=1$, let $c_\Psi$ and $\Psi$ be as in~\Cref{prop: app local fluctuation}, and let $L_{\rm env}\coloneqq 6B_K^2$. Moreover, shorten $\mathsf{a}\coloneqq B_K\sqrt{\log(e+B_K)}$ and let $\C_1$ and $\C_2$ as in~\Cref{thm: main error bound 2,prop: app local fluctuation}. For $t>0$, $\varrho\geq0$ and $m\geq3$, set $\kappa_t\coloneqq\max\{16\kappa_1c_\Psi(1+t),1\}$ and
\begin{align}\label{eq: tau expression}
\begin{split}
\tau^\star\coloneqq \underbrace{8\sqrt{\E^2_{\text{Bias}}+\varrho}}_{\tau^*_1}+\underbrace{\left(\frac{4\kappa_t\sqrt{\C_1\log(e+m)}}{\sqrt m}\right)^{\frac{2}{2+\gamma}}}_{\tau_2^*}&+\underbrace{4\kappa_t\mathsf{a}\sqrt{\frac{\log(e+m)}{m}}}_{\tau_3^*}\\
&+\underbrace{\sqrt{\frac{4\kappa_t(\C_2+L_{\rm env}\log m)}{m}}}_{\tau_4^*}.
\end{split}
\end{align}
Then $\tau=\tau^\star$ satisfies the inequality condition
\begin{align}\label{eq: app fixed point condition}
\tau^2\geq 16\left(\E^2_{\text{Bias}}+\varrho+\kappa_1t\left(\frac{\Psi(\tau)}{\sqrt m}+\frac{L_{\rm env}\log m}{m}\right)\right),
\end{align}
and
\begin{align}\label{eq: app fixed point bound}
\begin{split}
(\tau^\star)^2\leq 256\left(\E^2_{\text{Bias}}+\varrho\right)&+64\kappa_t^2\left(\left(\frac{\C_1\log(e+m)}{m}\right)^{\frac{2}{2+\gamma}}+\frac{\mathsf{a}^2\log(e+m)}{m}\right)\\
&+\frac{16\kappa_t(\C_2+L_{\rm env}\log m)}{m}.
\end{split}
\end{align}
\end{lemma}

\begin{proof}
We write $\tau^\star=\tau_1^*+\tau_2^*+\tau_3^*+\tau_4^*$ as in~\eqref{eq: tau expression}. Using the definition of $\Psi$ in~\eqref{eq: explicit psi}, together with $t\leq1+t$ and $c_\Psi\geq1$, the right-hand side of~\eqref{eq: app fixed point condition} evaluated at $\tau = \tau^*$ is at most
\begin{align*}
\underbrace{16\left(\E^2_{\text{Bias}}+\varrho\right)}_{r_1}+\kappa_t\Bigg(\underbrace{\frac{\sqrt{\C_1}(\tau^*)^{1-\frac\gamma2}\sqrt{\log(e+(\tau^*)^{-1})}}{\sqrt m}}_{r_2} &+\underbrace{\frac{\mathsf{a}\tau^*\sqrt{\log(e+(\tau^*)^{-1})}}{\sqrt m}}_{r_3}\\
&+\underbrace{\frac{\C_2+L_{\rm env}\log m}{m}}_{r_4}\Bigg).
\end{align*}
It therefore suffices to show that each of these four terms is at most $(\tau^*)^2/4$.

\emph{$r_1$ term.} Since $\tau^\star\geq\tau_1^*=8\sqrt{\E^2_{\text{Bias}}+\varrho}$, we have $16(\E^2_{\text{Bias}}+\varrho)\leq(\tau^\star)^2/4$.

\emph{$\kappa_t r_2$ term.} Since $\kappa_t\geq1$, $\C_1\geq1$ and $\log(e+m)\geq1$, we have $\tau^*\geq \tau_2^*\geq m^{-1/(2+\gamma)}\geq m^{-1}$. Thus, we have $\log(e+(\tau^\star)^{-1})\leq\log(e+m)$, and
\begin{align*}
(\tau^\star)^{1+\frac\gamma2}\geq(\tau_2^*)^{1+\frac\gamma2}=\frac{4\kappa_t\sqrt{\C_1}\sqrt{\log(e+m)}}{\sqrt m}\geq\frac{4\kappa_t\sqrt{\C_1}\sqrt{\log(e+(\tau^\star)^{-1})}}{\sqrt m}.
\end{align*}
Multiplying both sides by $(\tau^\star)^{1-\gamma/2}/4$ shows that $\kappa_t r_2\leq(\tau^\star)^2/4$.

\emph{$\kappa_t r_3$ term.} Similarly, since $\kappa_t,\mathsf{a}\geq1$, we have $\tau_3^*\geq m^{-1/2}\geq m^{-1}$. Hence, $\log(e+(\tau^\star)^{-1})\leq\log(e+m)$, and
\begin{align*}
\tau^\star\geq\tau^*_3=4\kappa_t\mathsf{a}\sqrt{\frac{\log(e+m)}{m}}\geq\frac{4\kappa_t\mathsf{a}\sqrt{\log(e+(\tau^\star)^{-1})}}{\sqrt m}.
\end{align*}
Multiplying both sides by $\tau^\star/4$ shows that the $\kappa_t r_3\leq(\tau^\star)^2/4$.

\emph{$\kappa_t r_4$ term.} Since $\tau^\star\geq\tau^*_4$, we have $\kappa_t(\C_2+L_{\rm env}\log m)/m=(\tau_4^*)^2/4\leq(\tau^\star)^2/4$.

Combining the four bounds proves~\eqref{eq: app fixed point condition}. Finally, note that $(\tau^\star)^2\leq 4((\tau_1^*)^2+(\tau_2^*)^2+(\tau_3^*)^2+(\tau_4^*)^2)$ which then yields~\eqref{eq: app fixed point bound}.
\end{proof}

\subsection{Error bound for approximate $\DD$ minimizers and proof for $\Th^m$ }\label{subsec: error bound for approximate minimizers}
We now combine~\Cref{thm: one-shot localization} with~\Cref{prop: app local fluctuation} and~\Cref{lem: app fixed point}. This yields an error bound on $\DD(\nu, \T)$ that holds simultaneously for all approximate minimizers $\T^*$ of $M(\nu^m,\cdot)$ over $\TT$. With $\varrho=0$, it gives the first claim of~\Cref{thm: main error bound 2} for the idealized estimator $\Th^m$. Later, we will use this result with $\varrho>0$ for the bound on $\Th^{m,n}$.

\begin{proposition}[$\DD$ error bound approximate minimizers of $M(\nu^m,\cdot)$]\label{prop: app oracle inequality}
Suppose~\Cref{Boundedness kernel} and~\Cref{Entropy lifted} hold, and let $m\geq3$. Then, for every $\varrho\geq0$ and $t>0$, there is an event of probability at least $1-e^{-t}$, depending only on $\{(y_i,u_i)\}_{i=1}^m$, on which every $\T^*\in\TT$ with
\begin{align*}
M(\nu^m,\T^*)\leq\inf_{\T\in\TT}M(\nu^m,\T)+\varrho
\end{align*}
satisfies
\begin{align*}
\DD(\nu,\T^*)\leq c\left(\E^2_{\text{Bias}}+\varrho+(1+t)^2\left(\left(\frac{\C_1\log(e+m)}{m}\right)^{\frac{2}{2+\gamma}}+\frac{\C_2\log(e+m)}{m}\right)\right),
\end{align*}
where $c$ is a numerical constant and $\C_1,\C_2$ are as in~\Cref{thm: main error bound 2}.
\end{proposition}

\begin{proof}
We apply~\Cref{thm: one-shot localization} with the following choices.
\begin{itemize}
\item \emph{Sample.} $\xi_i=(y_i,u_i)$, $i=1,\dots,m$, which are i.i.d.\ with law $P=\nu$, so that $R^m$ is the empirical risk.
\item \emph{Loss and minimizer.} The pseudo-normed vector space $(\mathcal{A},\|\cdot\|_2)$ with the loss $\phi$, which is convex in its first argument, and $a^\dag=g^\dag$, which minimizes $R$ over $\mathcal{A}$ by~\Cref{lem: app lifted problem}. By the excess risk identity~\eqref{eq: app excess risk identity}, $\|g-g^\dag\|_2=\sqrt{\ell(g)}$ for all $g\in\mathcal{A}$, so the growth condition holds with $w(x)=\sqrt x$.
\item \emph{Model and anchor.} $A=\G_\TT$ and $\bar a=\widehat g=g_{\Th}$, with $\ell(\widehat g)=\E^2_{\text{Bias}}$ by~\Cref{lem: app lifted problem}.
\item \emph{Auxiliary norm.} $|||\cdot|||=\|\cdot\|_\infty$ and $B=r=2B_K$. By Item 1 of~\Cref{lem: app hull}, $\sup_{g\in\G_\TT}\|g\|_\infty\leq B_K\leq B$, and $s\|g-\widehat g\|_\infty\leq 2B_K=r$ for all $g\in\G_\TT$ and $s\in[0,1]$. Thus, the constraint~\eqref{eq: contrained cond with auxiliary norm} in the definition of $\bar A$ is void, $\bar A=\widehat\G_\TT$, and $2B/r-1=1$.
\item \emph{Envelope.} By Item 2 of~\Cref{lem: app hull}, the constant function $F_{\text{env.}}\equiv4B_K^2$ is an envelope of $\{\phi(h,\cdot)\mid h\in\widehat\G_\TT\}$. It belongs to the Orlicz space of index $p=1$, with Orlicz norm $4B_K^2/\log2\leq L_{\rm env}=6B_K^2$; see~\Cref{rem: on the localization statement}.
\item \emph{Local fluctuations.} $\Psi$ as in~\Cref{prop: app local fluctuation}, which is valid for every $\tau>0$.
\end{itemize}
Let $\tau=\tau^\star$, with $\tau^\star$ as in~\Cref{lem: app fixed point}. For $w(x)=\sqrt x$ and $p=1$, condition~\eqref{eq: app localization condition rho} is equivalent to~\eqref{eq: app fixed point condition}, which holds by~\Cref{lem: app fixed point}. Let $\Omega$ be the event of~\Cref{thm: one-shot localization}. It depends only on $\{(y_i,u_i)\}_{i=1}^m$ and has probability at least $1-e^{-t}$.

Now let $\T^*\in\TT$ satisfy $M(\nu^m,\T^*)\leq\inf_{\T\in\TT}M(\nu^m,\T)+\varrho$. By~\Cref{lem: app lifted problem}, $M(\nu^m,\T)=R^m(g_\T)$ for every $\T\in\TT$. Thus,
\begin{align*}
R^m(g_{\T^*})=M(\nu^m,\T^*)\leq M(\nu^m,\Th)+\varrho=R^m(\widehat g)+\varrho,
\end{align*}
that is, $a^*=g_{\T^*}$ is an approximate minimizer in the sense of~\eqref{eq: approximate minimizer defn}. Therefore, on $\Omega$, \Cref{thm: one-shot localization} gives
\begin{align*}
\|g_{\T^*}-g^\dag\|_2\leq\sqrt{\ell(\widehat g)}+\tau^\star=\E_{\text{Bias}}+\tau^\star.
\end{align*}
By~\Cref{lem: app lifted problem}, it follows that
\begin{align*}
\DD(\nu,\T^*)=\|g_{\T^*}-g^\dag\|_2^2\leq 2\E^2_{\text{Bias}}+2(\tau^\star)^2.
\end{align*}

It remains to bound $(\tau^\star)^2$ using~\eqref{eq: app fixed point bound}. First, $\kappa_t\leq\max\{16\kappa_1c_\Psi,1\}(1+t)$. So, $\kappa_t,\kappa_t^2\leq c(1+t)^2$ for a numerical constant $c>0$. Second, since $C_{\G_\TT}\geq1$ and $B_K\geq1$, we have $C_\mathcal{L}\geq2\log(e+8B_K)\geq\max\{2,\log(e+B_K)\}$. This gives $\mathsf{a}^2=B_K^2\log(e+B_K)\leq C_LB_K^2\leq\C_2$ and $L_{\rm env}=6B_K^2\leq3C_LB_K^2\leq3\C_2$. Finally, $\log m\leq\log(e+m)$. Inserting these bounds into~\eqref{eq: app fixed point bound} yields
\begin{align*}
(\tau^\star)^2\leq c\left(\E^2_{\text{Bias}}+\varrho+(1+t)^2\left(\left(\frac{\C_1\log(e+m)}{m}\right)^{\frac{2}{2+\gamma}}+\frac{\C_2\log(e+m)}{m}\right)\right),
\end{align*}
which concludes the proof.
\end{proof}

\begin{proof}[Proof of the first claim of~\Cref{thm: main error bound 2} for $\Th^m$]
Since $\Th^m$ minimizes $M(\nu^m,\cdot)$ over $\TT$, it follows that 
~\Cref{prop: app oracle inequality} applies to $\T^*=\Th^m$ with $\varrho=0$. For $m\geq3$, we have $\log(e+m)\leq\log(2m)\leq2\log m$, which yields the claimed bound with probability at least $1-e^{-t}$. The bound on $\E_{\text{Bias}}$ is~\Cref{lem: app bias}.
\end{proof}

\subsection{Reference sampling error control}\label{subsec: reference sampling error control}
In order to exploit the result of~\Cref{prop: app oracle inequality} to extend an error bound on the practical and fully empirical estimator $\Th^{m,n}$, it is necessary to control uniformly over $\TT$ the centered sampling error arising in approximating the integrals of the transport map-parametrized conditional mean embedding using the samples from $\eta_\V$, i.e,
\begin{align}\label{eq: reference sampling error defn}
\E_n \coloneqq \sup_{\T\in\TT}\left|\delta_n(\T)-\delta_n(\Th)\right|,\qquad \delta_n(\T)\coloneqq \M^n(\nu^m,\T)-\M(\nu^m, \T).
\end{align}
To this end, we collect the samples approximating the integrals in vector form by $\mathbf{v}_i\coloneqq [v_{i,1}, ..., v_{i,n}]\overset{\textnormal{i.i.d}}{\sim}\eta_\V^{\otimes n}$ for any $i=1,...,m$ and we denote the extended measure $\widetilde{P}\coloneqq \nu\otimes \eta_\V^{\otimes n}\in\mathcal{P}(\Y\times\U\times \V^n)$, and let $\mathbf{z}_i\coloneqq (y_i, u_i, \mathbf{v}_i)\overset{\textnormal{i.i.d.}}{\sim}\widetilde P$. Furthermore, we introduce in compact notation, for any $\mathbf{z}= (y,u,(v_j)_{j\leq n})$,
\begin{align}\label{eq: compact notation for reference sampling error}
\begin{split}
c_j(\T, \mathbf{z})&\coloneqq K(\T(y,v_j), \cdot)-g_\T(y),\quad W_1(\T,\mathbf{z})\coloneqq\frac{1}{n}\sum_{j=1}^n c_j(\T, \mathbf{z}), \\
W_2(\T,\mathbf{z})&\coloneqq\frac{1}{n(n-1)}\sum_{j=1}^n\sum_{\substack{k=1\\
    k\neq j}}^n\langle c_j(\T, \mathbf{z}), c_k(\T, \mathbf{z})\rangle_{\mathcal{K}}, \\
\widetilde{W}(\T,\mathbf{z})&\coloneqq 2\langle g_\T(y)-K(u,\cdot), W_1(\T,\mathbf{z})\rangle_{\mathcal{K}}+W_2(\T,\mathbf{z}).
\end{split}
\end{align}
Before delving into the proof for the uniform sampling error bound of $\E_n$, we first present the following estimates which will be then employed in the proof.

\begin{lemma}[Decomposition of the reference-sampling error]\label{lem: app reference decomposition}
Suppose~\Cref{Boundedness kernel} and~\Cref{lipshitzness} in~\Cref{Assumptions}. Moreover, assume~\Cref{inf embedding} in~\Cref{assum: strengthened assum}, let $n\geq2$, and set $\Lambda_\infty\coloneqq L_KC_{\infty,\eta_\V}$. Let $\widetilde P^m\coloneqq\frac1m\sum_{i=1}^m\delta_{\mathbf{z}_i}$ be the empirical measure of $\mathbf{z}_1,\dots,\mathbf{z}_m$. Then, for all $\T,\S\in\TT$, the following hold.
\begin{enumerate}
\item \label{item 1 C14}$\delta_n(\T)=\frac1m\sum_{i=1}^m\widetilde W(\T,\mathbf{z}_i)$, and $\int\widetilde W(\T,(y,u,\mathbf{v}))\,\eta_\V^{\otimes n}(\dd\mathbf{v})=0$ for every $(y,u)\in\Y\times\U$. In particular, $\widetilde P\,\widetilde W(\T,\cdot)=0$ and $\delta_n(\T)=(\widetilde P^m-\widetilde P)\,\widetilde W(\T,\cdot)$.
\item\label{item 2 C14} $|\widetilde W(\T,\mathbf{z})|\leq 12B_K^2$ for every $\mathbf{z}$.
\item\label{item 3 C14} The increments satisfy
\begin{align}
\left\|\widetilde W(\T,\cdot)-\widetilde W(\S,\cdot)\right\|_{L^\infty(\widetilde P)}&\leq 20B_K\Lambda_\infty\|\T-\S\|_\infty,\label{eq: Linfty W}\\
\left\|\widetilde W(\T,\cdot)-\widetilde W(\S,\cdot)\right\|_{L^2(\widetilde P)}&\leq\frac{10B_K\min\left\{2B_K,\Lambda_\infty\|\T-\S\|_\infty\right\}}{\sqrt n}.\label{eq: L2 W}
\end{align}
\end{enumerate}
\end{lemma}

\begin{proof}
We prove each item in order.

\emph{Proof of~\Cref{item 1 C14}}. Let $\left(M^n(\nu^m, \T)\right)_i$ and $\left(M(\nu^m, \T)\right)_i$ be the $i^{\textnormal{th}}$ summands of respectively $M^n(\nu^m, \T)$ and $M(\nu^m, \T)$. 
Note that,
\begin{align*}
(M(\nu^m, \T))_i &= K(u_i,u_i)-2\langle K(u_i, \cdot), g_\T(y_i)\rangle_\mathcal{K}+\|g_\T(y_i)\|_\mathcal{K}^2
\end{align*}
On the other hand,
\begin{align*}
&\left(M^n(\nu^m, \T)\right)_i\\ &=K(u_i,u_i)-\frac{2}{n}\sum_{j=1}^n K(u_i,\T(y_i, v_{i,j})) + \frac{1}{n(n-1)}\sum_{j=1}^n\sum_{\substack{k=1\\
    k\neq j}}^n K(\T(y_i, v_{i,j}),\T(y_i, v_{i,k})),
\end{align*}
Using the reproducing property,
\begin{align*}
\frac{2}{n}\sum_{j=1}^n K(u_i,\T(y_i, v_{i,j})) &= \frac{2}{n}\sum_{j=1}^n \langle K(u_i,\cdot),K(\T(y_i, v_{i,j}), \cdot)\rangle_\mathcal{K}\\
&=\frac{2}{n}\sum_{j=1}^n \langle K(u_i,\cdot),g_\T(y_i) +c_j(\T,\mathbf{z}_i)\rangle_\mathcal{K} \\
&= 2\langle K(u_i,\cdot), g_\T(y_i)+ W_1(\T,\mathbf{z}_i)\rangle_\mathcal{K}.
\end{align*}
Furthermore, noting that $K(\T(y_i, v_{i,j}),\T(y_i, v_{i,k}))=\langle g_{\T}(y_i)+c_j(\T,\mathbf{z}_i), g_{\T}(y_i)+c_k(\T,\mathbf{z}_i)\rangle_{\mathcal{K}}$ and $\sum_{j=1}^n\sum_{\substack{k=1\\
    k\neq j}}^n (c_j(\T,\mathbf{z}_i) + c_k(\T,\mathbf{z}_i)) = 2(n-1)\sum_{j=1}^n c_j(\T,\mathbf{z}_i) $,
\begin{align*}
\frac{1}{n(n-1)}\sum_{j=1}^n\sum_{\substack{k=1\\
    k\neq j}}^n K(\T(y_i, v_{i,j}),\T(y_i, v_{i,k})) = \|g_\T(y_i)\|_\mathcal{K}^2+ 2\langle g_\T(y_i) ,W_1(\T,\mathbf{z}_i)\rangle_\mathcal{K} + W_2(\T,\mathbf{z}_i).
\end{align*}
Combining all these calculations, we then have that
\begin{align*}
\left(M^n(\nu^m,\T)-M(\nu^m,\T)\right)_i = 2\langle g_\T(y_i)-K(u_i,\cdot), W_1(\T,\mathbf{z}_i)\rangle_{\mathcal{K}}+W_2(\T,\mathbf{z}_i) = \widetilde{W}(\T,\mathbf{z}_i), 
\end{align*}
and, summing over the $i$ summands, $\delta_n(\T) = \frac1m\sum_{i=1}^m\widetilde W(\T,\mathbf{z}_i)$. Averaging over $i$ gives the first claim.

For the remainder of the proof, we use the notation $\mathbb{E}$ to denote the conditional expectation with respect to $\mathbf{v}\sim\eta^{\otimes n}$ for fixed pair $(y,u)$.

Since the $c_j(\T,\mathbf{z})$ are i.i.d. and centered under $\mathbb{E}$, we have $\mathbb{E}[W_1(\T,\mathbf{z})]=0$ and also $\mathbb{E}[W_2(\T,\mathbf{z})]=0$. Thus, $\mathbb{E}[\widetilde W(\T,\mathbf{z})]=0$ for every $(y,u)$. Integrating over $(y,u)\sim\nu$ gives $\widetilde P\,\widetilde W(\T,\cdot)=0$, and therefore $\delta_n(\T)=(\widetilde P^m-\widetilde P)\widetilde W(\T,\cdot)$. 

\emph{Proof of~\Cref{item 2 C14}}. By~\Cref{Boundedness kernel} in~\Cref{Assumptions}, $\|W_1(\T,\mathbf{z})\|_\mathcal{K}\leq 2B_K$ and $|W_2(\T,\mathbf{z})|\leq 4B_K^2$. Moreover, by~\Cref{lem: app lifted problem}, $\|g_\T(y)-K(u,\cdot)\|_\mathcal{K}\leq 2B_K$. Then, $|\widetilde W(\T,\mathbf{z})|\leq 2\cdot 2B_K\cdot 2B_K+4B_K^2=12B_K^2$.

\emph{Proof of~\Cref{item 3 C14}}. Using the same convention for $\mathbb{E}$, we further introduce
\begin{align*}
d_j&\coloneqq K(\T(y,v_j),\cdot)-K(\S(y,v_j),\cdot),\qquad d_g\coloneqq g_\T(y)-g_\S(y)=\mathbb{E}[d_j],\\
\widetilde d_j&\coloneqq d_j-d_g=c_j(\T,\mathbf{z})-c_j(\S,\mathbf{z}),\qquad \quad \ \widetilde{d} \coloneqq \frac{1}{n}\sum_{j=1}^n \widetilde{d}_j=W_1(\T,\mathbf{z})-W_1(\S,\mathbf{z}).
\end{align*}
Then, we decompose
\begin{align*}
\widetilde{W}(\T, \mathbf{z})-\widetilde{W}(\S, \mathbf{z}) &= 2\langle g_\T(y)-g_\S(y), W_1(\T,\mathbf{z})\rangle_\mathcal{K}+ 2\langle g_\S(y)-K(u,\cdot), W_1(\T,\mathbf{z})-W_1(\S,\mathbf{z})\rangle_\mathcal{K}\\
&\hspace{13.3em}+ W_2(\T,\mathbf{z})-W_2(\S,\mathbf{z})\\
&=2\langle d_g,W_1(\T,\mathbf{z})\rangle_\mathcal{K}+2\langle g_\S(y)-K(u,\cdot),\widetilde d\rangle_\mathcal{K} \\
&\hspace{8.5em}+ \frac{1}{n(n-1)}\sum_{j=1}^n\sum_{\substack{k=1\\
    k\neq j}}^n \langle\widetilde d_j,c_k(\T,\mathbf{z})\rangle_\mathcal{K} + \langle c_j(\S,\mathbf{z}),\widetilde d_k\rangle_\mathcal{K}.
\end{align*}
By~\Cref{lipshitzness} in~\Cref{Assumptions} and~\Cref{inf embedding} in~\Cref{assum: strengthened assum}, for $\nu_\Y$-a.e.\ $y$ and $\eta_\V$-a.e.\ $v$,
\begin{align}\label{eq: app lipschitz feature map}
\left\|d_j\right\|_\mathcal{K}\leq L_K\|\T(y,v_j)-\S(y,v_j)\|_\U\leq L_KC_{\infty,\eta_\V}\|\T(y,\cdot)-\S(y,\cdot)\|_{\HH}\leq\Lambda_\infty\|\T-\S\|_\infty.
\end{align}
By Jensen's inequality,~\eqref{eq: app lipschitz feature map} also yields $\|d_g\|_\mathcal{K}\leq \Lambda_\infty\|\T-\S\|_\infty$ and $\|\widetilde d_j\|_\mathcal{K}\leq 2\Lambda_\infty\|\T-\S\|_\infty$ as well. Moreover, $\|c_j(\T, \mathbf{z})\|_\mathcal{K},\|W_1\|_\mathcal{K},\|g_\T(y)-K(u,\cdot)\|_\mathcal{K}\leq 2B_K$ for any $\T\in\TT$. Plugging these bounds in the decomposition then yields~\eqref{eq: Linfty W}.

Next, setting $\alpha^2 \coloneqq\mathbb{E}\|d_1\|_\mathcal{K}^2$, we have that $\|d_g\|_\mathcal{K}\leq \alpha$ and $\mathbb{E}\|\widetilde d\|_\mathcal{K}^2 = \frac1n \mathbb{E}\|\widetilde d_1\|_\mathcal{K}^2\leq \frac{\alpha^2}{n}$. Moreover, $\mathbb{E}\|W_1(\T, \mathbf{z})\|_\mathcal{K}^2=\frac1n \mathbb{E}\|c_1(\T,\mathbf{z})\|_\mathcal{K}^2\leq \frac{B_K^2}{n}$ for any $\T\in\TT$. Using these bounds, we then have a bound on the second moments of the first two terms of the decomposition, i.e., 
\begin{align*}
\mathbb{E}\left[\left(2\langle d_g,W_1(\T,\mathbf{z})\rangle_\mathcal{K}\right)^2\right]\leq 4\alpha^2\frac{B_K^2}{n},\qquad \mathbb{E}\left[\left(2\langle g_\S(y)-K(u,\cdot),\widetilde d\rangle_\mathcal{K}\right)^2\right]\leq 16\alpha^2\frac{B_K^2}{n}.
\end{align*}
To control the remaining term of the decomposition, we first note that, by the independence of the $c_j(\T, \mathbf{z})$ terms which are centered, with $j\neq k$ and $j'\neq k'$,
\begin{align*}
\mathbb{E}[\langle\widetilde d_j,c_k(\T,\mathbf{z})\rangle_\mathcal{K}\langle\widetilde d_{j'},c_{k'}(\T,\mathbf{z})\rangle_\mathcal{K}] = \begin{cases}
0,\qquad \{j,k\}\neq\{j',k'\},\\
\mathbb{E}[\langle\widetilde d_j,c_k(\T,\mathbf{z})\rangle_\mathcal{K}^2]\leq B_K^2\alpha^2\qquad \{j,k\}=\{j',k'\}.
\end{cases}
\end{align*}
for any $\T\in\TT$, where the inequality follows by the Cauchy-Schwarz inequality. Thus, we can bound
\begin{align*}
\mathbb{E}\left[\left(\frac{1}{n(n-1)}\sum_{j=1}^n\sum_{\substack{k=1\\
    k\neq j}}^n \langle\widetilde d_j,c_k(\T,\mathbf{z})\rangle_\mathcal{K} + \langle c_j(\S,\mathbf{z}),\widetilde d_k\rangle_\mathcal{K}\right)^2\right]\leq 8 B_K^2\frac{\alpha^2}{n(n-1)}\leq 8 B_K^2\frac{\alpha^2}{n}.
\end{align*}
Consequently, we can combine all the three bounds on the decomposition terms to get
\begin{align*}
\mathbb{E}\left[\left(\widetilde W(\T,\mathbf{z})-\widetilde W(\S,\mathbf{z})\right)^2\right]\leq 3\left(4+16+8\right)\frac{B_K^2\alpha^2}{n}=\frac{84B_K^2\alpha^2}{n}.
\end{align*}
Finally, $\alpha\leq \min\{2B_K, \Lambda_\infty \|\T-\S\|_\infty\}$ by~\eqref{eq: app lipschitz feature map}. So, using $\sqrt{84}\leq 10$, yields~\eqref{eq: L2 W}.
\end{proof}

Now we can bound the uniform sampling error centered at $\Th$ defined in~\eqref{eq: reference sampling error defn}.

\begin{proposition}[Uniform control of the reference-sampling error]\label{prop: app reference error}
Suppose that~\Cref{Boundedness kernel} and~\Cref{lipshitzness} in~\Cref{Assumptions} and~\Cref{assum: strengthened assum} hold, let $n\geq2$, and let $\Lambda_\infty$, $R_\TT$, $\cc_1$, $\cc_2$, $\C_3$ and $\C_4$ be as in~\Cref{thm: main error bound 2}. Then, for every $t>0$, with probability at least $1-e^{-t}$,
\begin{align*}
\E_n\leq c\left(\frac{\C_4+B_K^2\sqrt t}{\sqrt{mn}}+\frac{\C_3+B_K^2t}{m}\right),
\end{align*}
where $c$ is a numerical constant.
\end{proposition}

\begin{proof}
If $R_\TT\coloneqq\textnormal{diam}_{\|\cdot\|_\infty}(\TT)=0$, then $\TT=\{\Th\}$ and $\E_n=0$. So we assume $R_\TT>0$. Consider the class
\begin{align*}
\mathcal{W}\coloneqq\left\{\widetilde W(\T,\cdot)-\widetilde W(\Th,\cdot)\mid\T\in\TT\right\}.
\end{align*}
By~\Cref{item 1 C14} of~\Cref{lem: app reference decomposition}, every $f\in\mathcal{W}$ satisfies $\widetilde Pf=0$, and
\begin{align*}
\E_n=\sup_{f\in\mathcal{W}}\left|(\widetilde P^m-\widetilde P)f\right|=\sup_{f\in\mathcal{W}\cup(-\mathcal{W})}(\widetilde P^m-\widetilde P)f.
\end{align*}
By~\Cref{item 3 C14} of~\Cref{lem: app reference decomposition} with $\S=\Th$, and by~\Cref{item 2 C14} of the same lemma, every $f\in\mathcal{W}\cup(-\mathcal{W})$ satisfies
\begin{align*}
\|f\|_{L^2(\widetilde P)}\leq\varsigma\coloneqq\frac{20B_K^2}{\sqrt n}\qquad\text{and}\qquad\|f\|_{L^\infty(\widetilde P)}\leq \Pi\coloneqq24B_K^2.
\end{align*}
Moreover, by~\Cref{item 3 C14} of~\Cref{lem: app reference decomposition}, the image of an $\epsilon$-net of $\TT$ in $\|\cdot\|_\infty$ under $\T\mapsto\widetilde W(\T,\cdot)-\widetilde W(\Th,\cdot)$ is a $(10B_K\Lambda_\infty\epsilon/\sqrt n)$-net of $\mathcal{W}$ in $L^2(\widetilde P)$, and a $(20B_K\Lambda_\infty\epsilon)$-net of $\mathcal{W}$ in $L^\infty(\widetilde P)$. This means that, for every $\delta>0$,
\begin{align*}
\Hf\left(\delta,\mathcal{W},L^2(\widetilde P)\right)\leq\Hf\left(\omega_2(\delta),\TT,\|\cdot\|_\infty\right),\qquad\Hf\left(\delta,\mathcal{W},L^\infty(\widetilde P)\right)\leq\Hf\left(\omega_\infty(\delta),\TT,\|\cdot\|_\infty\right),
\end{align*}
where $\omega_2(\delta)\coloneqq \delta\sqrt n/(10B_K\Lambda_\infty)$ and $\omega_\infty(\delta)\coloneqq \delta/(20B_K\Lambda_\infty)$. Moreover, $\Hf(\delta,\TT,\|\cdot\|_\infty)=0$ for $\delta\geq R_\TT$, since any single element of $\TT$ is then a $\delta$-net. Thus, by~\Cref{entropy hypothesis} in~\Cref{assum: strengthened assum} and $\log(1+x)\leq\log(e+x)$, the log-covering numbers of $\mathcal{W}$ are bounded by
\begin{align*}
H_2(\delta)&\coloneqq C_\TT\,\omega_2(\delta)^{-\gamma}\log\left(e+\omega_2(\delta)^{-1}\right)\mathbf{1}\{\omega_2(\delta)<R_\TT\},\\
H_\infty(\delta)&\coloneqq C_\TT\,\omega_\infty(\delta)^{-\gamma}\log\left(e+\omega_\infty(\delta)^{-1}\right)\mathbf{1}\{\omega_\infty(\delta)<R_\TT\}.
\end{align*}
The covering numbers of $\mathcal{W}\cup(-\mathcal{W})$ are at most twice those of $\mathcal{W}$, which only adds $\log2$ to the entropies and is absorbed into the numerical constant $c$.

We can then apply~\Cref{prop: dudley general} with $\varepsilon=\tilde\varepsilon=0$, $L=0$ and $q=0$, as discussed in~\Cref{rem: on the dudley statement} to obtain
\begin{align*}
\sqrt m\,\mathbb{E}[\E_n]\lesssim\int_0^\varsigma\sqrt{H_2(\delta)}\,\dd \delta+\frac{1}{\sqrt m}\int_0^\Pi H_\infty(\delta)\,\dd \delta.
\end{align*}
We then upper bound the Dudley integrals. To do so, we perform the substitutions $\delta=10B_K\Lambda_\infty w/\sqrt n$ and $\delta=20B_K\Lambda_\infty w$ which map the ranges $[0,\varsigma]$ and $[0,\Pi]$ onto $[0,2B_K/\Lambda_\infty]$ and $[0,6B_K/(5\Lambda_\infty)]$, respectively.  
Since the integrands vanish for $w\geq R_\TT$, setting $\cc_1\coloneqq\min\{R_\TT,2B_K/\Lambda_\infty\}$ and $\cc_2\coloneqq\min\{R_\TT,6B_K/(5\Lambda_\infty)\}$, we can use the calculations of~\Cref{calc1} and~\Cref{calc2} in~\Cref{lem: app entropy integrals} to bound
\begin{align*}
\int_0^\Pi H_\infty(\delta)\,\dd \delta&=20B_K\Lambda_\infty\int_0^{\cc_2}C_\TT w^{-\gamma}\log(e+w^{-1})\,\dd w\\
&\leq\frac{80B_K\Lambda_\infty C_\TT\,\cc_2^{1-\gamma}\log(e+\cc_2^{-1})}{(1-\gamma)^2}=80\,\C_3.
\end{align*}
and 
\begin{align*}
\int_0^\varsigma\sqrt{H_2(\delta)}\,\dd \delta&=\frac{10B_K\Lambda_\infty}{\sqrt n}\int_0^{\cc_1}\sqrt{C_\TT w^{-\gamma}\log(e+w^{-1})}\,\dd w\\
&\leq\frac{60B_K\Lambda_\infty\sqrt{C_\TT}\,\cc_1^{1-\frac\gamma2}\sqrt{\log(e+\cc_1^{-1})}}{\sqrt n}=\frac{60\,\C_4}{\sqrt n}.
\end{align*}
Consequently, combining these two bounds:
\begin{align}\label{eq: expectation fluctuation bound}
\mathbb{E}[\E_n]\lesssim \frac{\C_3}{m}+\frac{\C_4}{\sqrt{mn}}.
\end{align}
Ultimately, the class $\mathcal{W}\cup(-\mathcal{W})$ is $\widetilde P$-centered by~\Cref{item 1 C14} of~\Cref{lem: app reference decomposition} and satisfies the assumptions of~\Cref{lem: bousquet} with $\varsigma=20B_K^2/\sqrt n$ and $\Pi=24B_K^2$. Thus, 
with probability at least $1-e^{-t}$,
\begin{align}\label{eq: high probability fluctuation bound}
\E_n\leq 2\,\mathbb{E}[\E_n]+\varsigma\sqrt{\frac{2t}{m}}+\frac{4\Pi t}{3m}=2\,\mathbb{E}[\E_n]+\frac{20\sqrt2\,B_K^2\sqrt t}{\sqrt{mn}}+\frac{32B_K^2t}{m}.
\end{align}
Thus, combining~\eqref{eq: expectation fluctuation bound} and~\eqref{eq: high probability fluctuation bound} concludes the proof.
\end{proof}

\subsection{Proof for the error bound of $\Th^{m,n}$}\label{subsec: proof of error bound with mn}
Finally, combining~\Cref{prop: app oracle inequality} for approximate minimizers with the results of~\Cref{subsec: reference sampling error control} yields the average conditional estimation error bound on the estimator $\Th^{m,n}$.

\begin{proof}[Proof of the second claim of~\Cref{thm: main error bound 2} for $\Th^{m,n}$] For every $\T\in\TT$, by the definition of $\delta_n$ in~\eqref{eq: reference sampling error defn}, adding and subtracting $\M^n(\nu^m,\Th^{m,n})$ and $\M^n(\nu^m,\T)$,
\begin{align}\label{eq: decoupling errors}
\M(\nu^m,\Th^{m,n})-\M(\nu^m,\T)=\left[\M^n(\nu^m,\Th^{m,n})-\M^n(\nu^m,\T)\right]-\left[\delta_n(\Th^{m,n})-\delta_n(\T)\right].
\end{align}
Note that by $\Th^{m,n}$ minimizing $\M^n(\nu^m,\cdot)$ over $\TT$, the first term in~\eqref{eq: decoupling errors} is non-positive. Applying the triangle inequality on the second term, we further bound
\begin{align*}
\left|\delta_n(\Th^{m,n})-\delta_n(\T)\right|\leq\left|\delta_n(\Th^{m,n})-\delta_n(\Th)\right|+\left|\delta_n(\T)-\delta_n(\Th)\right|\leq2\,\E_n.
\end{align*}
Taking the infimum over $\T\in\TT$ gives
\begin{align}\label{eq: app Thmn approximate minimizer}
\M(\nu^m,\Th^{m,n})\leq\inf_{\T\in\TT}\M(\nu^m,\T)+2\,\E_n.
\end{align}
Then, applying together~\Cref{prop: app oracle inequality} and~\Cref{prop: app reference error} with a union bound, with probability at least $1-2e^{-t}$, simultaneously,
\begin{align*}
\M(\nu^m,\Th^{m,n})&\leq\inf_{\T\in\TT}\M(\nu^m,\T)+\varrho,\qquad \varrho\coloneqq c\left(\frac{\C_4+B_K^2\sqrt t}{\sqrt{mn}}+\frac{\C_3+B_K^2t}{m}\right),\\
\DD(\nu,\Th^{m,n})&\leq c\left(\E^2_{\text{Bias}}+\varrho+(1+t)^2\left(\left(\frac{\C_1\log(e+m)}{m}\right)^{\frac{2}{2+\gamma}}+\frac{\C_2\log(e+m)}{m}\right)\right).
\end{align*}
Ultimately, simplifying by $\frac{B_K^2t}{m}\leq(1+t)^2\frac{\C_2\log(e+m)}{m}$, concludes the proof of~\Cref{thm: main error bound 2}.
\end{proof}

\section{Proofs of the results in~\Cref{sec: DD vs DG minimization}}\label{app: proofs sec5}

Throughout this section, we implicitly assume that $\mathcal{K}$ is separable and that $y\mapsto I_K(\rho(\cdot\mid y))$, $K$, $G$, and of all the integrands below are measurable. We also introduce the compact notation $\|G\|_\infty\coloneqq\sup_{y,y'\in\Y}|G(y,y')|<\infty$ and recall that we distinguish between the integral operator $A_G$ introduced in~\eqref{eq: kernel integral operator in y}, which acts on $\mathcal{K}$-valued functions in $L^2(\Y,\mathcal{K};\rho_\Y)$ and its analogue $A_G^{\RR}$ acting on scalar functions $g\in L^2(\Y,\RR;\rho_\Y)$. 

\subsection{Proof of~\Cref{sec: CD vs JMMD 1}, Energy representations of $\DD$ and $\DG$}\label{sec: proofs CD vs JMMD 1}

We start by providing the proofs yielding the energy representations of the $\MMD$ objectives considered in this paper.

\begin{proof}[Proof of~\Cref{prop: energy rep}]
Let $f,g\in L^2(\Y,\mathcal{K};\rho_\Y)$. By the Cauchy-Schwarz inequality,
\begin{align*}
\left\|A_G f(y)\right\|_{\mathcal{K}}\leq\|G\|_\infty\int_\Y\|f(y')\|_{\mathcal{K}}\,\rho_\Y(\dd y')\leq\|G\|_\infty\|f\|_{L^2(\Y, \mathcal{K};\rho_\Y)},\qquad\forall y\in\Y,
\end{align*}
so that $\|A_Gf\|_{L^2(\Y, \mathcal{K};\rho_\Y)}\leq\|G\|_\infty\|f\|_{L^2(\Y, \mathcal{K};\rho_\Y)}$ and $A_G$ is bounded on $L^2(\Y,\mathcal{K};\rho_\Y)$. Similarly, $(y,y')\to G(y,y')\langle f(y),g(y')\rangle_{\mathcal{K}}$ is dominated by $\|G\|_\infty\|f(y)\|_{\mathcal{K}}\|g(y')\|_{\mathcal{K}}\in L^1(\rho_\Y\otimes\rho_\Y)$. Hence, exchanging the inner product with the Bochner integral and invoking Fubini's theorem,
\begin{align}\label{eq: app bilinear form A_G}
\langle f,A_Gg\rangle_{\rho_\Y}=\iint_{\Y\times\Y}G(y,y')\left\langle f(y),g(y')\right\rangle_{\mathcal{K}}\rho_\Y(\dd y)\rho_\Y(\dd y').
\end{align}

\emph{Energy representation of $\DD$.} Since $K$ is bounded, $\PP_K(\U)=\PP(\U)$ and, by the definition in~\eqref{eq: MMD def}, $\MMD_K(\rho(\cdot\mid y),\rho'(\cdot\mid y))=\|\Delta_{\rho,\rho'}(y)\|_{\mathcal{K}}$ for $\rho_\Y$-a.e.\ $y\in\Y$. Since $\|\Delta_{\rho,\rho'}(y)\|_{\mathcal{K}}\leq 2B_K$, we have that $\Delta_{\rho,\rho'}\in L^2(\Y,\mathcal{K};\rho_\Y)$. Consequently, by~\eqref{ACD-MMD-2}, 
\begin{align}\label{eq: step energy DC}
\DD(\rho\pipe\rho')=\int_\Y\|\Delta_{\rho,\rho'}(y)\|_{\mathcal{K}}^2\rho_\Y(\dd y)=\|\Delta\|_{L^2(\Y,\mathcal{K};\rho_\Y)}^2.
\end{align}

\emph{Energy representation of $\DG$.} Let $F\coloneqq K\otimes G$ be the product kernel in~\eqref{eq: joint product kernel} and $\mathcal{F}$ be its induced RKHS. By the Schur product theorem, $F$ is PSD. Moreover, $|F|\leq B_K^2\|G\|_\infty$, so every $\mu\in\PP(\Y\times\U)$ admits the mean embedding $I_F(\mu)=\int F(z,\cdot)\mu(\dd z)\in\mathcal{F}$ and, by the reproducing property~\eqref{eq: reproducing property} in $\mathcal{F}$,
\begin{align*}
\left\langle I_F(\mu),I_F(\mu')\right\rangle_{\mathcal{F}}=\int\left(I_F(\mu)\right)(z')\,\mu'(\dd z')=\iint F(z,z')\,\mu(\dd z)\mu'(\dd z'),\qquad\forall\mu,\mu'\in\PP(\Y\times\U).
\end{align*}
Expanding the square in~\eqref{eq: Joint MMD squared}, we then obtain
\begin{align}\label{eq: app DG three terms}
\DG(\rho,\rho')=\iint F\,\dd\rho\,\dd\rho-2\iint F\,\dd\rho\,\dd\rho'+\iint F\,\dd\rho'\,\dd\rho'.
\end{align}
Now let $\mu,\mu'\in\{\rho,\rho'\}$, both with $\Y$-marginal $\rho_\Y$. Disintegrating each factor of $\mu\otimes\mu'$, using Fubini's theorem since $F$ is bounded, and applying the reproducing property~\eqref{eq: reproducing property} in $\mathcal{K}$ twice,
\begin{align*}
\iint F\,\dd\mu\,\dd\mu'&=\iint_{\Y\times\Y}G(y,y')\left(\int_\U\int_\U K(u,u')\,\mu(\dd u\mid y)\,\mu'(\dd u'\mid y')\right)\rho_\Y(\dd y)\rho_\Y(\dd y')\\
&=\iint_{\Y\times\Y}G(y,y')\left\langle I_K(\mu(\cdot\mid y)),I_K(\mu'(\cdot\mid y'))\right\rangle_{\mathcal{K}}\rho_\Y(\dd y)\rho_\Y(\dd y').
\end{align*}
Next, we write $a(y)\coloneqq I_K(\rho(\cdot\mid y))$ and $b(y)\coloneqq I_K(\rho'(\cdot\mid y))$, so that $\Delta_{\rho,\rho'}=a-b$. By the symmetry of $G$ and the exchange of the integration variables $y$ and $y'$,
\begin{align*}
\iint G(y,y')\langle a(y),b(y')\rangle_{\mathcal{K}}\rho_\Y(\dd y)\rho_\Y(\dd y')=\iint G(y,y')\langle b(y),a(y')\rangle_{\mathcal{K}}\rho_\Y(\dd y)\rho_\Y(\dd y').
\end{align*}
Therefore, splitting the cross term in~\eqref{eq: app DG three terms} into these two equal halves,
\begin{align*}
\DG(\rho,\rho')&=\iint G(y,y')\Big(\langle a(y),a(y')\rangle_{\mathcal{K}}-\langle a(y),b(y')\rangle_{\mathcal{K}}\\
&\qquad \qquad\qquad \qquad -\langle b(y),a(y')\rangle_{\mathcal{K}}+\langle b(y),b(y')\rangle_{\mathcal{K}}\Big)\rho_\Y(\dd y)\rho_\Y(\dd y')\\
&=\iint G(y,y')\left\langle \Delta_{\rho,\rho'}(y),\Delta_{\rho,\rho'}(y')\right\rangle_{\mathcal{K}}\rho_\Y(\dd y)\rho_\Y(\dd y')=\left\langle\Delta_{\rho,\rho'},A_G\Delta_{\rho,\rho'}\right\rangle_{\rho_\Y},
\end{align*}
where the last equality is~\eqref{eq: app bilinear form A_G} with $f=g=\Delta_{\rho,\rho'}$. Together with~\eqref{eq: step energy DC}, this proves~\eqref{eq: energy representaton DD and DG}.

It remains to establish the properties of $A_G$. Consider first the scalar operator $A_G^{\RR}$ acting on $L^2(\Y,\RR;\rho_\Y)$. Since $G$ is bounded, measurable and PSD with separable RKHS $\mathcal{G}$, we have $\int_\Y G(y,y)\,\rho_\Y(\dd y)\leq\|G\|_\infty<\infty$. Then, by~\cite[Thm.~4.27]{steinwart2008support}, $A_G^{\RR}=S_G^*S_G$, where $S_G:L^2(\Y,\RR;\rho_\Y)\to\mathcal{G}$ is the mapping $S(g)\coloneqq\int_\Y G(\cdot,y')g(y')\,\rho_\Y(\dd y')$, and $A_G^{\RR}$ is compact, positive, self-adjoint and Hilbert-Schmidt. By the spectral theorem~\cite[Thm.~A.5.13]{steinwart2008support}, there exist $N\in\mathbb{N}\cup\{\infty\}$, an orthonormal family $\{\phi_k\}_{k=1}^N\subset L^2(\Y,\RR;\rho_\Y)$, and eigenvalues $\lambda_1\geq\lambda_2\geq\dots>0$, with $\lambda_k\to0$ if $N=\infty$, such that
\begin{align*}
A_G^{\RR}g=\sum_{k=1}^N\lambda_k\langle g,\phi_k\rangle_{L^2(\Y,\RR;\rho_\Y)}\,\phi_k,\qquad\forall\,g\in L^2(\Y,\RR;\rho_\Y),
\end{align*}
where $\{\lambda_k\}_{k=1}^N$ is the set of nonzero eigenvalues of $A_G^{\RR}$, which are positive since $A_G^{\RR}$ is positive. In particular, $g\perp\phi_k$ for all $k$ implies $A_G^{\RR}g=0$, while each $\phi_k=\lambda_k^{-1}A_G^{\RR}\phi_k$ is orthogonal to $\ker A_G^{\RR}$ by self-adjointness. Thus, $\{\phi_k\}_{k=1}^N$ spans a dense subspace of $(\ker A_G^{\RR})^\perp$.

Next, let $\{e_j\}_{j\in J}$ be an orthonormal basis of the separable space $\mathcal{K}$, and, for an element $f\in L^2(\Y,\mathcal{K};\rho_\Y)$, write $f_j\coloneqq\langle f(\cdot),e_j\rangle_\mathcal{K}\in L^2(\Y,\RR;\rho_\Y)$, so that $\|f\|^2_{L^2(\Y,\mathcal{K};\rho_\Y)}=\sum_{j\in J}\|f_j\|^2_{L^2(\Y,\RR;\rho_\Y)}$ by Parseval's identity in $\mathcal{K}$ and Fubini-Tonelli's theorem to exchange integral and countable sum. Commuting the bounded functional $\langle\cdot,e_j\rangle_\mathcal{K}$ with the Bochner integral in~\eqref{eq: kernel integral operator in y} by~\cite[eq. A.32]{steinwart2008support} gives
\begin{align}\label{eq: app coordinates of A_G}
\left\langle A_Gf(y),e_j\right\rangle_\mathcal{K}=\left(A_G^{\RR}f_j\right)(y),\qquad j\in J,
\end{align}
that is, $A_G$ acts as $A_G^{\RR}$ on each coordinate. Consequently, for $f,g\in L^2(\Y,\mathcal{K};\rho_\Y)$, we get
\begin{align*}
\|A_Gf\|^2_{L^2(\Y,\mathcal{K};\rho_\Y)}&=\sum_{j\in J}\|A_G^{\RR}f_j\|_{L^2(\Y,\mathbb{R};\rho_\Y)}^2\leq\|A_G^{\RR}\|^2\|f\|^2_{L^2(\Y,\mathcal{K};\rho_\Y)},\\
\langle f,A_Gg\rangle_{L^2(\Y,\mathcal{K};\rho_\Y)}&=\sum_{j\in J}\langle f_j,A_G^{\RR}g_j\rangle_{L^2(\Y,\mathbb{R};\rho_\Y)},
\end{align*}
so that $A_G$ is bounded, self-adjoint and positive, as $A_G^{\RR}$ is. Moreover, $A_G(\phi_k\,e)=\lambda_k\phi_k\,e$ for every $e\in\mathcal{K}$. Hence, if $\dim\mathcal{K}=\infty$ and $A_G\neq0$, the orthonormal sequence $(\phi_1e_j)_{j\in J}$ is mapped to $(\lambda_1\phi_1e_j)_{j\in J}$, which has no convergent subsequence; in this case $A_G$ is not compact. So while $A_G^\mathbb{R}$ is compact $A_G$ not necessarily is.

If $G$ is integrally strictly positive definite with respect to $\rho_\Y$, then $\langle g,A_G^{\RR}g\rangle>0$ for every nonzero $g\in L^2(\Y,\RR;\rho_\Y)$, so that $\ker A_G^{\RR}=\{0\}$. Therefore, $\{\phi_k\}_{k=1}^N$ is an orthonormal basis of $L^2(\Y,\RR;\rho_\Y)$, and $N=\infty$ whenever $L^2(\Y,\RR;\rho_\Y)$ is infinite dimensional. Moreover, if $A_Gf=0$, then~\eqref{eq: app coordinates of A_G} gives $A_G^{\RR}f_j=0$, $f_j=0$ for every $j\in J$, so that $f=0$. Thus, $A_G$ is injective. Integral strict positive definiteness is defined in~\cite[Sec.~1.2]{sriperumbudur2010hilbert} with respect to all finite nonzero signed Borel measures, and thus implies the above condition with respect to $\rho_\Y$. The later considered Gaussian kernel $\Gsi$ (and $\Gsil$ too), is integrally strictly positive definite on $\RR^{d_\Y}$~\cite[Sec.~3.1]{sriperumbudur2010hilbert}; see also~\cite[Prop.~5]{sriperumbudur2011universality}.

Finally, for $f\in L^2(\Y,\mathcal{K};\rho_\Y)$, let $c_k(f)\coloneqq\int_\Y f(y)\phi_k(y)\,\rho_\Y(\dd y)\in\mathcal{K}$. Then, we have that $\langle f_j,\phi_k\rangle_{L^2(\Y,\mathbb{R};\rho_\Y)}=\langle c_k(f),e_j\rangle_\mathcal{K}$. By the above coordinate identities
\begin{align}\label{eq: spectral rep}
\langle f,A_Gf\rangle_{L^2(\Y,\mathcal{K};\rho_\Y)}=\sum_{j\in J}\sum_{k=1}^N\lambda_k\langle f_j,\phi_k\rangle_{L^2(\Y, \mathbb{R}; \rho_\Y)}^2=\sum_{k=1}^N\lambda_k\|c_k(f)\|_\mathcal{K}^2,
\end{align}
where the last step follows by Parseval's identity and Fubini-Tonelli theorem.
When $\{\phi_k\}_{k=1}^N$ is an orthonormal basis, Parseval's identity in $L^2(\Y,\RR;\rho_\Y)$ similarly gives $\|f\|^2_{L^2(\Y,\mathcal{K};\rho_\Y)}=\sum_{k=1}^N\|c_k(f)\|^2_\mathcal{K}$. Taking $f=\Delta_{\rho,\rho'}$ in~\eqref{eq: energy representaton DD and DG} yields the spectral form of $\DD$ and $\DG$ stated after~\Cref{prop: energy rep}.
\end{proof}

As an immediate result of this proposition, we can relate $\DD$ and $\DG$ via the spectrum of $A_G^\mathbb{R}$.

\begin{proof}[Proof of~\Cref{prop: comparison}]
Fix $\T\in\TT$. Since $P_\Y\circ\T_\t=P_\Y$ by~\eqref{eq:triangular-map-form}, we have $(\T_\t\#\eta)_\Y=\eta_\Y=\nu_\Y$ by~\Cref{assumption:independent-reference}. Hence the pair $\rho=\nu$, $\rho'=\T_\t\#\eta$ satisfies the assumptions of~\Cref{prop: energy rep} with $\rho_\Y=\nu_\Y$, and~\eqref{eq: energy representaton DD and DG} gives, with $\Delta_\T$ as in~\eqref{eq: compact notation CME difference with map},
\begin{align}\label{eq: app energies for maps}
\DD(\nu,\T)=\left\|\Delta_\T\right\|^2_{L^2(\Y,\mathcal{K};\nu_\Y)},\qquad \DG(\nu,\T)=\left\langle\Delta_\T,A_G\Delta_\T\right\rangle_{L^2(\Y,\mathcal{K};\nu_\Y)}.
\end{align}
If $\Delta_\T=0$, both quantities vanish and the two-sided bound holds trivially. Otherwise, $\Delta_\T\in\mathcal{D}\setminus\{0\}$, so by the definition of $\lambda_G^\pm$,
\begin{align*}
\lambda_G^-\leq\frac{\left\langle\Delta_\T,A_G\Delta_\T\right\rangle_{L^2(\Y,\mathcal{K};\nu_\Y)}}{\left\|\Delta_\T\right\|^2_{L^2(\Y,\mathcal{K};\nu_\Y)}}\leq\lambda_G^+ .
\end{align*}
Multiplying by $\|\Delta_\T\|^2_{L^2(\Y,\mathcal{K};\nu_\Y)}=\DD(\nu,\T)$ and using~\eqref{eq: app energies for maps} yields $\lambda_G^-\DD(\nu,\T)\leq\DG(\nu,\T)\leq\lambda_G^+\DD(\nu,\T)$. Note that $0\leq\lambda_G^-\leq\lambda_G^+\leq\|A_G\|\leq\|G\|_\infty<\infty$ by the proof of~\Cref{prop: energy rep}.

For~\eqref{eq: sandwitch bound}, the lower bound follows from $\Th_G\in\TT$ and $\E_{\text{Bias}}^2=\inf_{\T\in\TT}\DD(\nu,\T)$. For the upper bound, if $\lambda_G^-=0$ then $\kappa_G=\infty$ and there is nothing to prove. If $\lambda_G^->0$, then for any $\T\in\TT$, applying the two-sided bound to $\Th_G$ and to $\T$ and using that $\Th_G$ minimizes $\DG(\nu,\cdot)$ over $\TT$,
\begin{align*}
\lambda_G^-\,\DD(\nu,\Th_G)\leq\DG(\nu,\Th_G)\leq\DG(\nu,\T)\leq\lambda_G^+\,\DD(\nu,\T).
\end{align*}
Dividing by $\lambda_G^-$ and taking the infimum over $\T\in\TT$ gives $\DD(\nu,\Th_G)\leq\kappa_G\,\E_{\text{Bias}}^2$.

Finally, we consider the well-specified case without assuming $\kappa_G<\infty$. Suppose there exists $\T^\star\in\TT$ with $\DD(\nu,\T^\star)=0$ and that $A_G$ is injective. Then $\Delta_{\T^\star}=0$ by~\eqref{eq: app energies for maps}, hence $\DG(\nu,\T^\star)=0$ and, by minimality, $\DG(\nu,\Th_G)=0$. Since $A_G$ is bounded, self-adjoint and positive, it admits a square root, and $\DG(\nu,\Th_G)=\|A_G^{1/2}\Delta_{\Th_G}\|^2_{L^2(\Y,\mathcal{K};\nu_\Y)}=0$ implies $A_G\Delta_{\Th_G}=0$, hence $\Delta_{\Th_G}=0$ and $\DD(\nu,\Th_G)=0$. This justifies the consistency claim of~\Cref{rem: distorsion control}.
\end{proof}

When $\lambda_G^->0$, even if not very common to satisfy, we can then transfer any existing generalization bound of $\DG$ to obtain a bound on the average conditional estimation error of $\Th_G^m$.
\begin{proof}[Proof of~\Cref{prop: spectral bound joint estimator}]
Fix $t>0$ and work on the event of probability at least $1-e^{-t}$ on which the generalization bound of~\Cref{assum: generalization of joint MMD estimator} holds, i.e.,
\begin{align}\label{eq: app generalization G}
\sqrt{\DG\left(\nu,\Th^m_G\right)}\leq\sqrt{\DG\left(\nu,\Th_G\right)}+\R(m,t).
\end{align}
The rest of the argument is deterministic. By the upper bound in~\Cref{prop: comparison}, $\DG(\nu,\T)\leq\lambda_G^+\DD(\nu,\T)$ for every $\T\in\TT$. Since $\Th_G$ minimizes $\DG(\nu,\cdot)$ over $\TT$, we have that $\DG(\nu,\Th_G)\leq\DG(\nu,\T)\leq\lambda_G^+\DD(\nu,\T)$ for every $\T\in\TT$. Thus, taking the infimum over $\T\in\TT$ yields
\begin{align*}
\DG\left(\nu,\Th_G\right)\leq\lambda_G^+\,\E^2_{\text{Bias}}.
\end{align*}
Plugging this into~\eqref{eq: app generalization G} gives
\begin{align}\label{eq: first part of spectral bound}
\sqrt{\DG\left(\nu,\Th^m_G\right)}\leq\sqrt{\lambda_G^+}\,\E_{\text{Bias}}+\R(m,t).
\end{align}
Since $\Th^m_G\in\TT$, the lower bound in~\Cref{prop: comparison} gives $\lambda_G^-\DD(\nu,\Th^m_G)\leq\DG(\nu,\Th^m_G)$. Then, when we can divide by $\lambda_G^->0$~\eqref{eq: first part of spectral bound}, we obtain
\begin{align*}
\sqrt{\DD\left(\nu,\Th^m_G\right)}\leq\frac{1}{\sqrt{\lambda_G^-}}\sqrt{\DG\left(\nu,\Th^m_G\right)}\leq\sqrt{\frac{\lambda_G^+}{\lambda_G^-}}\,\E_{\text{Bias}}+\frac{\R(m,t)}{\sqrt{\lambda_G^-}}=\sqrt{\kappa_G}\,\E_{\text{Bias}}+\frac{\R(m,t)}{\sqrt{\lambda_G^-}}.
\end{align*}
Squaring both sides proves the claim. In the well-specified case $\E_{\text{Bias}}=0$, the bound reduces to $\DD(\nu,\Th^m_G)\leq\R(m,t)^2/\lambda_G^-$.
\end{proof}

We conclude this subsection by verifying~\Cref{assum: generalization of joint MMD estimator} in the parametric setting, i.e., the rate~\eqref{eq: parametric class joint MMD rate}.

\begin{lemma}[Generalization bound for parametric classes]\label{lem: app joint generalization}
Let $\TT=\TT_\Theta$ be as in~\eqref{eq: param class} with $\Theta\subseteq\RR^p$ compact, suppose that~\Cref{Boundedness kernel} and~\Cref{lipshitzness} in~\Cref{Assumptions} and~\Cref{inf embedding} in~\Cref{assum: strengthened assum} hold, let $G$ be a bounded measurable PSD kernel with $\sup_{y\in\Y}G(y,y)\leq B_G^2$, and assume $\|\T_\theta-\T_{\theta'}\|_\infty\leq L_\TT|\theta-\theta'|$ for all $\theta,\theta'\in\Theta$. Then~\Cref{assum: generalization of joint MMD estimator} holds with $\R(m,t)$ as in~\eqref{eq: parametric class joint MMD rate}.
\end{lemma}

\begin{proof}
For simplicity, we write compactly $\MMD\coloneqq\MMD_{K\otimes G}$, $P_\theta\coloneqq\T_{\theta\t}\#\eta$, $P^m_\theta\coloneqq\T_{\theta\t}\#\eta^m$, $k\coloneqq\sup_{(y,u)\in\Y\times\U}K\otimes G((y,u),(y,u))\leq B_K^2B_G^2$, and let $\widehat\theta^m_G$ and $\widehat\theta_G$ be such that $\Th^m_G=\T_{\widehat\theta^m_G}$ and $\Th_G=\T_{\widehat\theta_G}$. Next, we define $Z_m\coloneqq\sup_{\theta\in\Theta}\MMD(P_\theta,P^m_\theta)$. Since $\DG^m(\nu^m,\T_\theta)^{1/2}=\MMD(\nu^m,P^m_\theta)$,  by triangle inequality, the minimality of $\widehat\theta^m_G$, and triangle inequality again,
\begin{align*}
\sqrt{\DG(\nu,\Th^m_G)}&\leq\MMD(\nu,\nu^m)+\MMD\big(\nu^m,P^m_{\widehat\theta^m_G}\big)+\MMD\big(P^m_{\widehat\theta^m_G},P_{\widehat\theta^m_G}\big)\\
&\leq\MMD(\nu,\nu^m)+\MMD\big(\nu^m,P^m_{\widehat\theta_G}\big)+Z_m\\
&\leq 2\MMD(\nu,\nu^m)+\MMD\big(\nu,P_{\widehat\theta_G}\big)+2Z_m\\
&=\sqrt{\DG(\nu,\Th_G)}+2\MMD(\nu,\nu^m)+2Z_m,
\end{align*}
By~\cite[Lem.~1]{briol2019statistical}, whose proof in the finite dimensional setting holds verbatim on any measurable space (as pointed in~\cite[Rem. 2.8]{pandey2024diffeomorphic}), for any fixed probability measure $P$ and $\delta\in(0,1)$, the empirical measure $P^m$ of $m$ i.i.d.\ samples from $P$ satisfies, with probability at least $1-\delta$,
\begin{align}\label{eq: app briol concentration}
\MMD(P,P^m)\leq\sqrt{\frac{2k}{m}}\left(1+\sqrt{\log(1/\delta)}\right).
\end{align}
Then, we can apply~\eqref{eq: app briol concentration} to control $\MMD(\nu,\nu^m)$ choosing $\delta=e^{-t}/2$ and we just need to bound $Z_m$. To do so, first recall that $P^m_\theta$ is the empirical measure of $m$ i.i.d.\ samples from $P_\theta$. Moreover, since $\|K\otimes G((y,u),\cdot)-K\otimes G((y,u'),\cdot)\|=\sqrt{G(y,y)}\,\|K(u,\cdot)-K(u',\cdot)\|_\KK$, arguing as in~\eqref{eq: app lipschitz feature map} (and using Jensen's inequality for $P_\theta$) gives, almost surely,
\begin{align}\label{eq: Lipschitz type bound used}
\MMD(P_\theta,P_{\theta'})+\MMD(P^m_\theta,P^m_{\theta'})\leq 2\Lambda_\Theta|\theta-\theta'|,\qquad\Lambda_\Theta\coloneqq B_GL_KC_{\infty,\eta_\V}L_\TT.
\end{align}
By~\cite[Lem. 5.7]{wainwright2019high}, there exists a set $\mathcal N\subseteq\Theta$ such that it is an $\varepsilon$-net of $\Theta$ with $\varepsilon\coloneqq B_K/(L_KC_{\infty,\eta_\V}L_\TT\sqrt m)$ and $|\mathcal N|\leq(1+4\,\textnormal{diam}(\Theta)/\varepsilon)^p=(1+4L_\Theta\sqrt m)^p$. Consequently, by triangle inequality and~\eqref{eq: Lipschitz type bound used}, we can bound
\begin{align}\label{eq: app Zm net}
Z_m\leq\max_{\theta\in\mathcal N}\MMD(P_\theta,P^m_\theta)+2\Lambda_\Theta\varepsilon=\max_{\theta\in\mathcal N}\MMD(P_\theta,P^m_\theta)+\frac{2B_KB_G}{\sqrt m}.
\end{align}
Finally, using~\eqref{eq: app briol concentration} to each $\theta\in\mathcal N$ with $\delta=e^{-t}/(2|\mathcal N|)$, a union bound, and $\log|\mathcal N|\leq p\log(1+4L_\Theta\sqrt m)\leq 3p\log(e+L_\Theta m)$ yield, with probability at least $1-e^{-t}$,
\begin{align*}
\sqrt{\DG(\nu,\Th^m_G)}\leq\sqrt{\DG(\nu,\Th_G)}+4\sqrt{\frac{2k}{m}}\left(1+\sqrt{t+\log2+3p\log(e+L_\Theta m)}\right)+\frac{4B_KB_G}{\sqrt m}.
\end{align*}
Since $\sqrt k\leq B_KB_G$, this gives~\Cref{assum: generalization of joint MMD estimator} with $\R(m,t)$ as in~\eqref{eq: parametric class joint MMD rate}, which is decreasing in $m$, non-decreasing in $t$, and independent of $\sigma$ for $G=\Gsi$, since then $B_G=1$.
\end{proof}

\subsection{Proofs of~\Cref{sec: inconsistency}. Inconsistency of $\DG$ minimizer for the average conditional estimation error}\label{sec: proofs inconsistency}

\begin{proof}[Proof of~\Cref{thm: inconsistency}]
Fix $f\in L^2(\Y,\mathcal K; \nu_\Y)$. By~\Cref{regularity} in~\Cref{assum: inconsistency assumptions}, $\theta\mapsto f-\mu_\theta$ is $C^2$ from $\Omega$ into $L^2(\Y,\mathcal K; \nu_\Y)$. Hence $Q_f$ and $Q_{G,f}$ in~\eqref{eq: DG DD expression for general h} are $C^2$ on $\Omega$. Since $A_G$ is bounded and self-adjoint by~\Cref{prop: energy rep}, the chain rule gives, for $j=1,\dots,p$,
\begin{align}\label{eq: app gradients}
\partial_jQ_f(\theta)=-2\left\langle\partial_j\mu_\theta,f-\mu_\theta\right\rangle,\qquad \partial_jQ_{G,f}(\theta)=-2\left\langle A_G\partial_j\mu_\theta,f-\mu_\theta\right\rangle.
\end{align}
Consequently,
\begin{align}\label{eq: app stationarity}
\nabla Q_f(\theta)=0\iff f-\mu_\theta\perp E_\theta,\qquad \nabla Q_{G,f}(\theta)=0\iff f-\mu_\theta\perp A_GE_\theta,
\end{align}
so that $f\in\mathcal{B}$ if and only if $f-\mu_\theta\perp E_\theta+A_GE_\theta$ for some $\theta\in\Theta$. By~\Cref{regularity} and~\Cref{no Invariance} in~\Cref{assum: inconsistency assumptions}, $\dim(E_\theta+A_GE_\theta)\geq p+1$. Belonging to $\mathcal{B}$ therefore requires at least $p+1$ independent linear conditions on $f$, while only $p$ parameters are available to satisfy them. We make this count rigorous next.

\emph{$\mathcal{B}$ is closed, nowhere dense and Haar-null.} Let $f_n\in\mathcal{B}$ with $f_n\to f$ in $L^2(\Y,\mathcal K; \nu_\Y)$ with corresponding $\theta_n\in\Theta$ such that $\nabla Q_{f_n}(\theta_n)=0=\nabla Q_{G,f_n}(\theta_n)$. By compactness of $\Theta$, up to a subsequence, $\theta_n\to\theta\in\Theta$. Since $\theta\mapsto\mu_\theta$ and $\theta\mapsto\partial_j\mu_\theta$ are continuous on $\Omega$ and $A_G$ is bounded, the right-hand sides of~\eqref{eq: app gradients} are jointly continuous in $(\theta,f)\in\Omega\times L^2(\Y,\mathcal K; \nu_\Y)$. Passing to the limit gives $\nabla Q_f(\theta)=0=\nabla Q_{G,f}(\theta)$, i.e., $f\in\mathcal{B}$. The same argument shows that, for every compact $C\subseteq\Theta$, the set $\mathcal{B}_C\coloneqq\{f\in L^2(\Y,\mathcal K; \nu_\Y)\mid\exists\,\theta\in C\textnormal{ s.t. }\nabla Q_f(\theta)=0=\nabla Q_{G,f}(\theta)\}$ is also closed.

Next, fix $\theta_0\in\Theta$. By~\Cref{no Invariance} in~\Cref{assum: inconsistency assumptions}, there exists $j_0\in\{1,\dots,p\}$ with $A_G\partial_{j_0}\mu_{\theta_0}\notin E_{\theta_0}$. For $\theta\in\Omega$, set
\begin{align*}
v_j(\theta)\coloneqq\partial_j\mu_\theta\quad(j=1,\dots,p),\qquad v_{p+1}(\theta)\coloneqq A_G\partial_{j_0}\mu_\theta,\qquad L_\theta w\coloneqq\left(\langle v_a(\theta),w\rangle\right)_{a=1}^{p+1}\in\RR^{p+1}.
\end{align*}
By the injectivity of $J_{\theta_0}\mu_{\theta_0}$ and the choice of $j_0$, the vectors $v_1(\theta_0),\dots,v_{p+1}(\theta_0)$ are linearly independent. Moreover, if $f\in\mathcal{B}$ with corresponding $\theta$ such that $\nabla Q_{f}(\theta)=0=\nabla Q_{G,f}(\theta)$, then~\eqref{eq: app stationarity} gives $L_\theta(f-\mu_\theta)=0$. The first $p$ components vanish because $\nabla Q_f(\theta)=0$, and the last one is (up to the factor $-2$) the $j_0$-th component of $\nabla Q_{G,f}(\theta)$. Let $\iota:\RR^{p+1}\to L^2(\Y,\mathcal K; \nu_\Y)$, $\iota(x)\coloneqq\sum_{b=1}^{p+1}x_bv_b(\theta_0)$, and
\begin{align*}
\mathfrak{G}(\theta)\coloneqq\left(\langle v_a(\theta),v_b(\theta_0)\rangle\right)_{a,b=1}^{p+1}\in\RR^{(p+1)\times(p+1)}.
\end{align*}
Since $\theta\mapsto\partial_j\mu_\theta$ is $C^1$ on $\Omega$ and $A_G$ is bounded and linear, $\mathfrak{G}$ is $C^1$ on $\Omega$. Moreover, $\mathfrak{G}(\theta_0)$ is the Gram matrix of linearly independent vectors and hence invertible. Thus there is a closed ball $K_0\coloneqq\overline{B}(\theta_0,r_0)$ and an open set $U_0$ with $K_0\subset U_0\subseteq\Omega$ such that $\mathfrak{G}(\theta)$ is invertible for all $\theta\in U_0$.

Then, let $\mathcal{B}_0\coloneqq\mathcal{B}_{K_0\cap\Theta}$ (which is closed by what we showed at the beginning of the proof) and fix $f_0\in L^2(\Y,\mathcal K;\nu_\Y)$. For $\theta\in U_0$ and $x\in\RR^{p+1}$, we have $L_\theta(\iota(x))=\mathfrak{G}(\theta)x$. Hence, for $c(\theta)$ with entries $c_a(\theta)\coloneqq\left\langle v_a(\theta),\mu_\theta-f_0\right\rangle$,
\begin{align*}
L_\theta\big(f_0+\iota(x)-\mu_\theta\big)=0\iff \mathfrak{G}(\theta)x=c(\theta)\iff x=s(\theta)\coloneqq \mathfrak{G}(\theta)^{-1}c(\theta).
\end{align*}
This then implies
\begin{align*}
X_0(f_0)\coloneqq\left\{x\in\RR^{p+1}\mid f_0+\iota(x)\in\mathcal{B}_0\right\}\subseteq s(K_0\cap\Theta).
\end{align*}
The map $s$ is $C^1$ on $U_0$, hence Lipschitz on the convex compact set $K_0$. Since $K_0\cap\Theta\subseteq\RR^p$ has zero $(p+1)$-dimensional Hausdorff measure and Lipschitz maps do not increase Hausdorff measures beyond a multiplicative constant, $s(K_0\cap\Theta)$, and thus $X_0(f_0)$, is Lebesgue-null in $\RR^{p+1}$. Two consequences follow.
\begin{itemize}
\item[(a)] \emph{$\mathcal{B}_0$ is Haar-null.} Let $\mu_0\coloneqq\iota\#\textnormal{Unif}(\overline{B}_{\RR^{p+1}}(0,1))$, a compactly supported Borel probability measure on $L^2(\Y,\mathcal{K};\nu_\Y)$. For every $x\in L^2(\Y,\mathcal{K};\nu_\Y)$, taking $f_0=-x$ gives $\mu_0(\mathcal{B}_0+x)=\textnormal{Unif}\big(\{x'\mid -x+\iota(x')\in\mathcal{B}_0\}\big)=\textnormal{Unif}\big(X_0(-x)\big)=0$.
\item[(b)] \emph{$\mathcal{B}_0$ is nowhere dense.} Given arbitrary $f_0\in L^2(\Y,\mathcal K;\nu_\Y)$ and $r>0$, the set $\{x\mid\|\iota(x)\|<r\}$ is an open neighbourhood of $0$ in $\RR^{p+1}$. Hence, it has positive Lebesgue measure. It therefore contains some $x\notin X_0(f_0)$, i.e., $f_0+\iota(x)\in B(f_0,r)\setminus\mathcal{B}_0$. Thus the closed set $\mathcal{B}_0$ has empty interior, i.e., it is nowhere dense in its closure (which corresponds to itself since it is a closed set).
\end{itemize}

Ultimately, we extend the result to all of $\mathcal{B}$ by compactness of $\Theta$. Indeed, by this property, there exist finitely many $\theta_0^{(1)},\dots,\theta_0^{(N)}\in\Theta$ such that the interiors of the corresponding balls $K_0^{(i)}$ from the previous steps in the proof cover $\Theta$. Every $\theta\in\Theta$ for which $f\in\mathcal{B}$ then lies in some $K_0^{(i)}\cap\Theta$, so $\mathcal{B}=\bigcup_{i=1}^N\mathcal{B}_0^{(i)}$. A finite union of closed, nowhere dense sets is closed and nowhere dense, and a finite union of Haar-null sets is Haar-null. Hence $\mathcal{B}$ is closed, nowhere dense and Haar-null, yielding the first statement of the theorem. Next, we define 
\begin{align*}
f_\nu(y) = I_K(\nu(\cdot\mid y)),\qquad \forall y\in\Y,
\end{align*}
and proceed with the inconsistency results of the theorem assuming that $f_\nu\notin\mathcal{B}$.

\emph{Inconsistency statements.} Fix $\theta\in\Theta$. Then, $\Delta_{\T_\theta}=f_\nu-\mu_\theta$. By~\Cref{prop: energy rep}, we have that $\DD(\nu,\T_\theta)=Q_{f_\nu}(\theta)$ and $\DG(\nu,\T_\theta)=Q_{G,f_\nu}(\theta)$. As previously shown, both functions are continuous on the compact set $\Theta$. Thus, $\SS$ and $\SS_G$ are nonempty and compact, and $\E_{\text{Bias}}^2=\min_{\theta\in\Theta}Q_{f_\nu}(\theta)$.

Suppose $\theta^\ast\in\SS\cap\SS_G$. By~\Cref{interiority} in~\Cref{assum: inconsistency assumptions}, $\theta^\ast\in\textnormal{int}\,\Theta$, so there is an open ball around $\theta^\ast$ contained in $\Theta$ on which $\theta^\ast$ minimizes both $Q_{f_\nu}$ and $Q_{G,f_\nu}$. By Fermat's rule, $\nabla Q_{f_\nu}(\theta^\ast)=0=\nabla Q_{G,f_\nu}(\theta^\ast)$, i.e., $f_\nu\in\mathcal{B}$, leading to a contradiction. Hence, $\SS\cap\SS_G=\varnothing$.

Since $Q_{f_\nu}$ is continuous on the nonempty compact set $\SS_G$, it attains its minimum there at some $\theta^\circ$. By what shown, $\theta^\circ\notin\SS$. Consequently $Q_{f_\nu}(\theta^\circ)>\min_{\theta\in\Theta} Q_{f_\nu}=\E_{\text{Bias}}^2$ and $\mathfrak{d}_G=Q_{f_\nu}(\theta^\circ)-\E^2_{\text{Bias}}>0$, and $\DD(\nu,\T_{\theta_G})=Q_{f_\nu}(\theta_G)\geq Q_{f_\nu}(\theta^\circ)=\E^2_{\text{Bias}}+\mathfrak{d}_G$ for every $\theta_G\in\SS_G$.

Fix an outcome in the almost sure event on which $\sup_{\theta\in\Theta}|\DG^{m}(\nu^m,\T_\theta)-Q_{G,f_\nu}(\theta)|\to0$. Let $\theta_\infty$ be a limit point of the sequence $\{\widehat\theta^m_G\}_m$, say $\widehat\theta^{m_k}_G\to\theta_\infty\in\Theta$. For every $\theta\in\Theta$, by continuity of $Q_{G,f_\nu}$, uniform convergence, and minimality of $\widehat\theta^{m_k}_G$,
\begin{align*}
Q_{G,f_\nu}(\theta_\infty)=\lim_{k\to\infty}Q_{G,f_\nu}\big(\widehat\theta^{m_k}_G\big)=\lim_{k\to\infty}\DG^{m_k}\big(\nu^{m_k},\T_{\widehat\theta^{m_k}_G}\big)\leq\lim_{k\to\infty}\DG^{m_k}\big(\nu^{m_k},\T_\theta\big)=Q_{G,f_\nu}(\theta),
\end{align*}
and $\theta_\infty\in\SS_G$. Next, pick a subsequence for which $\DD(\nu,\T_{\widehat\theta^{m_k}_G})\to\liminf_{m\to\infty}\DD(\nu,\T_{\widehat\theta^{m}_G})$. By compactness of $\Theta$, it has a further subsequence converging to some $\theta_\infty$, which lies in $\SS_G$ by the previous argument. Then,
\begin{align*}
\liminf_{m\to\infty}\DD\big(\nu,\T_{\widehat\theta^m_G}\big)=Q_{f_\nu}(\theta_\infty)\geq\E^2_{\text{Bias}}+\mathfrak{d}_G.
\end{align*}
\end{proof}

\subsection{Proofs of~\Cref{sec: joint minimization conditional MMD rate}. Average conditional estimation error bound of $\Th_{\Gsi}^m$ under Gaussian smoothing
}\label{sec: proofs joint minimization conditional MMD rate}

We start with the result relating $\DDR$, $\DGsi$, and $\DGsil$ through Fourier representations.
\begin{proof}[Proof of~\Cref{lem: DDR vs DG}]
Note that $\|\Delta_{\rho,\rho'}(y)\|_\mathcal{K}\leq 2B_K$. Thus, $\|f_{\rho,\rho'}(y)\|_\mathcal{K}\leq 2B_Kp_{\rho_\Y}(y)$ and $\int\|f_{\rho,\rho'}(y)\|_\mathcal{K}\,\dd y\leq 2B_K\int p_{\rho_\Y}(y)\,\dd y=2B_K$. Consequently,
\begin{align*}
\int\|f_{\rho,\rho'}(y)\|^2_\mathcal{K}\,\dd y\leq 4B_K^2\int p_{\rho_\Y}^2(y)\,\dd y\leq 4B_K^2\|p_{\rho_\Y}\|_\infty\int p_{\rho_\Y}(y)\,\dd y=4B_K^2\|p_{\rho_\Y}\|_\infty,
\end{align*}
meaning that $f_{\rho,\rho'}\in L^1\cap L^2(\RR^{d_\Y},\mathcal{K};\dd y)$.

The kernel $\Gsil(y,y')=\phi_\sigma(y-y')$ is bounded, measurable and PSD. Hence~\Cref{prop: energy rep} applies with $G=\Gsil$, and~\eqref{eq: app bilinear form A_G} with $f=g=\Delta_{\rho,\rho'}$ gives
\begin{align*}
\DGsil(\rho,\rho')&=\iint\phi_\sigma(y-y')\left\langle\Delta_{\rho,\rho'}(y),\Delta_{\rho,\rho'}(y')\right\rangle_\mathcal{K}\rho_\Y(\dd y)\rho_\Y(\dd y')\\
&=\iint\phi_\sigma(y-y')\left\langle f_{\rho,\rho'}(y),f_{\rho,\rho'}(y')\right\rangle_\mathcal{K}\dd y\,\dd y'=\left\langle f_{\rho,\rho'},\phi_\sigma*f_{\rho,\rho'}\right\rangle_{L^2(\mathbb{R}^{d_\Y}, \mathcal{K};\dd y)},
\end{align*}
where the last equality follows from Fubini's theorem, since $\phi_\sigma$ is bounded and $f_{\rho,\rho'}\in L^1(\mathbb{R}^{d_\Y}, \mathcal{K};\dd y)$. Since $\|\phi_\sigma\|_{L^1(\mathbb{R}^{d_\Y}, \mathbb{R};\dd y)}=1$ it follows that $\phi_\sigma*f\in L^2(\RR^{d_\Y},\mathcal{K};\dd y)$ by Young's convolution inequality. By Plancherel's theorem, $\DDR(\rho \pipe \rho') = \int \|\widehat f_{\rho,\rho'}(\xi)\|_\mathcal{K}^2\dd\xi$. Analogously, using the fact that $\widehat{\phi_\sigma * f_{\rho,\rho'}}(\xi) = (2\pi)^{d_\Y/2}\widehat{\phi_\sigma}\widehat{f_{\rho,\rho'}}(\xi) =e^{-\sigma^2\|\xi\|^2/2}\widehat{f_{\rho,\rho'}}(\xi)$,
\begin{align}\label{eq: Fourier thing}
\DGsil(\rho,\rho')=\left\langle\widehat f_{\rho,\rho'},e^{-\sigma^2\|\cdot\|^2/2}\,\widehat f_{\rho,\rho'}\right\rangle_{\dd \xi}=\int e^{-\sigma^2\|\xi\|^2/2}\left\|\widehat f_{\rho,\rho'}(\xi)\right\|_\mathcal{K}^2\dd\xi.
\end{align}
Since $e^{-\sigma^2\|\xi\|^2/2}\in(0,1]$,~\eqref{eq: Fourier thing} gives $0\leq\DGsil(\rho,\rho')\leq\DDR(\rho\pipe\rho')$. Moreover, $e^{-\sigma^2\|\xi\|^2/2}$ increases pointwise to $1$ as $\sigma\downarrow0$. Thus, the monotone convergence theorem yields $\DGsil(\rho,\rho')\uparrow\DDR(\rho\pipe\rho')$.

Next, note that $\iint\phi_\sigma(z)\|f_{\rho,\rho'}(y)\|_\mathcal{K}\|f_{\rho,\rho'}(y-z)\|_\mathcal{K}\,\dd y\,\dd z\leq\|f_{\rho,\rho'}\|_{L^2(\mathbb{R}^{d_\Y}, \mathcal{K};\dd y)}^2$ by applying the Cauchy-Schwarz inequality. Thus, by Fubini's theorem again and noting that $\phi_\sigma$ is even,
\begin{align*}
\left\langle f_{\rho,\rho'},\phi_\sigma*f_{\rho,\rho'}\right\rangle_{L^2(\mathbb{R}^{d_\Y}, \mathcal{K};\dd y)}&=\int\phi_\sigma(z)\left\langle f_{\rho,\rho'},f_{\rho,\rho'}(\cdot-z)\right\rangle_{L^2(\mathbb{R}^{d_\Y}, \mathcal{K};\dd y)}\dd z\\
&=\int\phi_\sigma(z)\left\langle f_{\rho,\rho'},f_{\rho,\rho'}(\cdot+z)\right\rangle_{L^2(\mathbb{R}^{d_\Y}, \mathcal{K};\dd y)}\dd z.
\end{align*}
By translation invariance of the Lebesgue measure, 
\begin{align*}
\|f_{\rho,\rho'}\|_{L^2(\mathbb{R}^{d_\Y}, \mathcal{K};\dd y)}^2-\langle f_{\rho,\rho'},f_{\rho,\rho'}(\cdot+z)\rangle_{L^2(\mathbb{R}^{d_\Y}, \mathcal{K};\dd y)}=\frac12\|f_{\rho,\rho'}(\cdot+z)-f_{\rho,\rho'}\|_{L^2(\mathbb{R}^{d_\Y}, \mathcal{K};\dd y)}^2.
\end{align*}
Consequently, we obtain the increment equality in~\eqref{eq: increment identity}:
\begin{align*}
\DDR(\rho\pipe\rho')-\DGsil(\rho,\rho')=\frac12\int\phi_\sigma(z)\left\|f(\cdot+z)-f\right\|^2_{L^2(\mathbb{R}^{d_\Y},\mathcal{K};\dd y)}\dd z.
\end{align*}
\end{proof}

Before jumping into the proof of~\Cref{Error bound of joint minimizer}, we begin by presenting some preliminary lemmas. We start by controlling the conversion gap between $\DD(\nu,\T)$ and $\DDR(\nu,\T)$. To this end, we introduce
\begin{align}\label{eq: DRAC min}
\ThR \in\argmin_{\T\in\TT}\DDR(\nu,\T).
\end{align}

\begin{lemma}[Conversion gap bound]\label{lem: conversion gap}
Let $K$ be bounded as in~\Cref{Boundedness kernel} of~\Cref{Assumptions}, $\nu_\Y$ have density $p_{\nu_\Y}$ with $\|p_{\nu_\Y}\|_\infty<\infty$, and suppose that~\Cref{margin condition} in~\Cref{assum: smoothing and conversion gap} holds. Then:
\begin{enumerate}
\item For every $\T\in\TT$,
\begin{align}\label{eq: app conversion bound}
\DD(\nu,\T)\leq 2\C_\beta\,\DDR(\nu,\T)^{\cc_\beta}.
\end{align}
\item If $\T^*\in\TT$ satisfies $\DDR(\nu,\T^*)\leq\inf_{\T\in\TT}\DDR(\nu,\T)+\varepsilon$ for some $\varepsilon\geq0$, then
\begin{align*}
\DD(\nu,\T^*)\leq 2\C_\beta\left(\|p_{\nu_\Y}\|_\infty\E^2_{\text{Bias}}+\varepsilon\right)^{\cc_\beta}.
\end{align*}
In particular, $\DD(\nu,\ThR)\leq 2\C_\beta(\|p_{\nu_\Y}\|_\infty\E^2_{\text{Bias}})^{\cc_\beta}$ for $\ThR$ in~\eqref{eq: DRAC min} and
\begin{align}\label{eq: conversion bound with smoothing gap}
\DD(\nu,\Th_{\Gsi})\leq 2\C_\beta\left(\|p_{\nu_\Y}\|_\infty\E^2_{\text{Bias}}+\Xi_\sigma\right)^{\cc_\beta},
\end{align}
with $\Xi_\sigma$ the uniform smoothing gap in~\eqref{eq: uniform smoothing gap}.
\end{enumerate}
\end{lemma}

\begin{proof}
Fix $\T\in\TT$ and set $\psi(y)\coloneqq\|\Delta_\T(y)\|_\mathcal{K}^2$. Note that $0\leq\psi\leq 4B_K^2$ since $\|\Delta_\T(y)\|_\mathcal{K}\leq 2B_K$. Let $I\coloneqq\DDR(\nu,\T)=\int_\Y\psi(y) p_{\nu_\Y}(y)\nu_\Y(\dd y)$, and note that $\DD(\nu,\T)=\int_\Y\psi(y)\nu_\Y(\dd y)$ by~\Cref{prop: energy rep}. Since $\nu_\Y(\{p_{\nu_\Y}=0\})=\int_{\{p_{\nu_\Y}(y)=0\}}p_{\nu_\Y}(y)\dd y=0$, splitting $\{p_{\nu_\Y}(y)>0\}$ into $\{p_{\nu_\Y}(y)\geq\lambda\}$ and $\{0<p_{\nu_\Y}(y)<\lambda\}$ gives, for every $\lambda>0$,
\begin{align}\label{eq: app splitting}
\begin{split}
\DD(\nu,\T)&=\int_{\{p_{\nu_\Y}(y)\geq\lambda\}}\psi(y)\nu_\Y(\dd y)+\int_{\{0<p_{\nu_\Y}(y)<\lambda\}}\psi(y) \nu_\Y(\dd y)\\
&\leq\frac{1}{\lambda}\int_{\{p_{\nu_\Y}(y)\geq\lambda\}}\psi(y) p_{\nu_\Y}(y)\nu_\Y(\dd y)+4B_K^2\,\m(\lambda)\leq\frac{I}{\lambda}+4B_K^2\,\m(\lambda).
\end{split}
\end{align}
Next, we distinguish the cases where $\beta$ is either infinity or not.

\emph{Case $\beta=\infty$.} Taking $\lambda=p_{\nu_\Y}^{\min}$ in~\eqref{eq: app splitting} and using $\m(p_{\nu_\Y}^{\min})=0$ trivially yields $\DD(\nu,\T)\leq(p_{\nu_\Y}^{\min})^{-1}I=2\C_\infty I^{\cc_\infty}$, since $\cc_\infty=1$.

\emph{Case $\beta<\infty$.} By the margin condition,~\eqref{eq: app splitting} gives $\DD(\nu,\T)\leq h(\lambda)\coloneqq I/\lambda+\Phi\lambda^\beta$ for every $\lambda>0$, where $\Phi\coloneqq 4B_K^2C_\beta$. If $I=0$, letting $\lambda\downarrow0$ gives $\DD(\nu,\T)=0$, and~\eqref{eq: app conversion bound} holds trivially. If $I>0$, choosing $\lambda^\ast\coloneqq(I/(\beta \Phi))^{1/(1+\beta)}$, so that $\Phi(\lambda^\ast)^{1+\beta}=I/\beta$, gives
\begin{align*}
\DD(\nu,\T)\leq h(\lambda^\ast)=\frac{I}{\lambda^\ast}\left(1+\frac1\beta\right)=c_\beta\,\Phi^{\frac{1}{1+\beta}}\,I^{\frac{\beta}{1+\beta}},\qquad c_\beta\coloneqq(1+\beta)\,\beta^{-\frac{\beta}{1+\beta}}.
\end{align*}
Ultimately, $\ell(\beta)\coloneqq\log c_\beta=\log(1+\beta)-\frac{\beta}{1+\beta}\log\beta$ satisfies $\ell'(\beta)=-\log\beta/(1+\beta)^2$. Hence $\ell$ is maximized at $\beta=1$, where $c_1=2$, so that $c_\beta\leq 2$ for all $\beta\in(0,\infty)$. Since $\Phi^{1/(1+\beta)}=(4B_K^2C_\beta)^{1/(1+\beta)}=\C_\beta$ and $\beta/(1+\beta)=\cc_\beta$, this concludes the proof of~\eqref{eq: app conversion bound}.

Next, consider $\T^*\in\TT$ satisfying 
\begin{align}\label{eq: condition almost optimal}
\DDR(\nu,\T^*)\leq\inf_{\T\in\TT}\DDR(\nu,\T)+\varepsilon
\end{align}
for some $\varepsilon\geq0$. Then, recalling that $\DDR(\nu,\T)\leq \|p_{\nu_\Y}\|_\infty \DD(\nu, \T)$ for any $\T\in\TT$, taking the infimum over $\TT$ yields
\begin{align*}
\DDR(\nu, \T^*)\leq \|p_{\nu_\Y}\|_\infty \DD(\nu, \Th) + \varepsilon = \|p_{\nu_\Y}\|_\infty \E_{\text{Bias}}^2 + \varepsilon.
\end{align*}
Thus, by~\eqref{eq: app conversion bound} and monotonicity of $x\to(x)^{\cc_\beta}$, it follows that
\begin{align}\label{eq: epsilon thing to hold}
\DD(\nu,\T^*)\leq 2\C_\beta\left(\|p_{\nu_\Y}\|_\infty\E^2_{\text{Bias}}+\varepsilon\right)^{\cc_\beta}.
\end{align}
Finally, by the definition~\eqref{eq: uniform smoothing gap} of $\Xi_\sigma$, the minimality of $\Th_{\Gsi}$, and $\DGsil\leq\DDR$ shown in~\Cref{lem: DDR vs DG},
\begin{align*}
\DDR(\nu,\Th_{\Gsi})\leq\DGsil(\nu,\Th_{\Gsi})+\Xi_\sigma\leq\DGsil(\nu,\T)+\Xi_\sigma\leq\DDR(\nu,\T)+\Xi_\sigma,
\end{align*}
for every $\T\in\TT$. Thus $\Th_{\Gsi}$ satisfies~\eqref{eq: condition almost optimal} with $\varepsilon=\Xi_\sigma$ so that~\eqref{eq: epsilon thing to hold} yields~\eqref{eq: conversion bound with smoothing gap}.
\end{proof}
Next, we move on to bound the uniform smoothing gap $\Xi_{\sigma}$ in~\eqref{eq: uniform smoothing gap}

\begin{lemma}[Smoothing gap bound]\label{lem: smoothing gap}
Let $K$ be bounded as in~\Cref{Boundedness kernel} of~\Cref{Assumptions}, $\nu_\Y$ have Lebesgue density $p_{\nu_\Y}$ with $\|p_{\nu_\Y}\|_\infty<\infty$, and suppose that~\Cref{L2-modulus assumption} in~\Cref{assum: smoothing and conversion gap} holds. Then, for every $\sigma>0$, the uniform smoothing gap~\eqref{eq: uniform smoothing gap} satisfies
\begin{align}\label{eq: bounds on Xi}
0\leq\Xi_\sigma\leq\frac12\,\mathcal{M}_\zeta^2\,\mathbb{E}\left[\|Z\|^{2\zeta}\right]\sigma^{2\zeta}\leq\C_\Xi\,\sigma^{2\zeta},\qquad Z\sim N(0,\textnormal{Id}_{d_\Y}),
\end{align}
with $\C_\Xi$ as in~\eqref{eq: smoothing gap bound and constant}.
\end{lemma}

\begin{proof}
First, recall that, by~\Cref{lem: DDR vs DG}, $\DGsil(\nu,\T)\leq\DDR(\nu,\T)$ for every $\T\in\TT$. Thus, $\Xi_\sigma\geq0$. Next, fix $\T\in\TT$. Combining the increment identity in~\eqref{eq: increment identity} with~\Cref{L2-modulus assumption} in~\Cref{assum: smoothing and conversion gap},
\begin{align*}
\DDR(\nu,\T)-\DGsil(\nu,\T)&=\frac12\int\phi_\sigma(z)\left\|f_\T(\cdot+z)-f_\T\right\|^2_{L^2(\mathbb{R}^{d_\Y},\mathcal{K};\dd y)}\dd z\\
&\leq\frac12\mathcal{M}_\zeta^2\int\phi_\sigma(z)\|z\|^{2\zeta}\dd z=\frac12\mathcal{M}_\zeta^2\,\sigma^{2\zeta}\,\mathbb{E}\left[\|Z\|^{2\zeta}\right],
\end{align*}
where $Z\sim N(0, \textnormal{Id}_{d_\Y})$ since $\phi_\sigma$ is the density of a standard normal distribution rescaled by $\sigma$. Ultimately, by the concavity of $x\mapsto x^\zeta$ on $[0,\infty)$ for $\zeta\in(0,1]$ and Jensen's inequality, $\mathbb{E}[\|Z\|^{2\zeta}]=\mathbb{E}[(\|Z\|^2)^\zeta]\leq(\mathbb{E}\|Z\|^2)^\zeta=d_\Y^\zeta$, completing the last inequality in~\eqref{eq: bounds on Xi}.
\end{proof}

We can now proceed with the proof of the main average conditional estimation error bound on $\Th_{\Gsi}^m$.

\begin{proof}[Proof of~\Cref{Error bound of joint minimizer}]
$\C_\Xi=0$ immediately implies that $\mathcal{M}_\zeta=0$. In that case every $f_\T$ is then invariant under translations and hence vanishes. Since $p_{\nu_\Y}>0$ $\nu_\Y$-a.e., this gives $\Delta_\T=0$ $\nu_\Y$-a.e., i.e., $\DD(\nu,\T)=0$ for every $\T\in\TT$, and all claims hold trivially. We therefore assume $\C_\Xi>0$ throughout the rest of the proof.

\emph{Error bound of~\Cref{Fixed length-scale} for fixed length-scale.}Fix $t>0$ and general $\sigma>0$, and consider the event of probability at least $1-e^{-t}$ on which~\Cref{assum: generalization of joint MMD estimator} holds for $\Gsi$. Squaring the generalization bound gives
\begin{align}\label{eq: square gen bound}
\DGsi\left(\nu,\Th^m_{\Gsi}\right)\leq\DGsi\left(\nu,\Th_{\Gsi}\right)+2\sqrt{\DGsi\left(\nu,\Th_{\Gsi}\right)}\,\R(m,t)+\R(m,t)^2.
\end{align}
Next, using the minimality of $\Th_{\Gsi}$, the inequality $\DGsil\leq\DDR$ from~\Cref{lem: DDR vs DG}, and the upper bound $\DDR(\nu,\T)=\int\|\Delta_\T(y)\|^2_\mathcal{K}p_{\nu_\Y}(y)\nu_\Y(\dd y)\leq\|p\|_\infty\DD(\nu,\T)$, we get that
\begin{align*}
c_\sigma\DGsi\left(\nu,\Th_{\Gsi}\right)=\DGsil\left(\nu,\Th_{\Gsi}\right)\leq\DGsil(\nu,\T)\leq\DDR(\nu,\T)\leq\|p\|_\infty\DD(\nu,\T),
\end{align*}
for any $\T\in\TT$. Thus, taking the infimum over $\TT$,
\begin{align}\label{eq: app population joint term}
\DGsil\left(\nu,\Th_{\Gsi}\right)=c_\sigma\DGsi\left(\nu,\Th_{\Gsi}\right)\leq\|p\|_\infty\E^2_{\text{Bias}}.
\end{align}
Multiplying~\eqref{eq: square gen bound} by $c_\sigma$ and combining it with~\eqref{eq: app population joint term},
\begin{align}\label{eq: important step proof TGm}
\begin{split}
\DGsil\left(\nu,\Th^m_{\Gsi}\right)&\leq\DGsil\left(\nu,\Th_{\Gsi}\right)+2\sqrt{c_\sigma}\sqrt{c_\sigma\DGsi\left(\nu,\Th_{\Gsi}\right)}\,\R(m,t)+c_\sigma\R(m,t)^2\\
&\leq\DGsil\left(\nu,\Th_{\Gsi}\right)+\widetilde\C_1\E_{\text{Bias}}\R(m,t)\,\sigma^{-d_\Y/2}+\widetilde\C_2\R(m,t)^2\sigma^{-d_\Y} \\
&\leq
\|p\|_\infty\E^2_{\text{Bias}}+\widetilde\C_1\E_{\text{Bias}}\R(m,t)\,\sigma^{-d_\Y/2}+\widetilde\C_2\R(m,t)^2\sigma^{-d_\Y}
\end{split}
\end{align}
where we used $2\sqrt{c_\sigma}\sqrt{\|p\|_\infty}=2(2\pi)^{-d_\Y/4}\sqrt{\|p\|_\infty}\,\sigma^{-d_\Y/2}=\widetilde\C_1\sigma^{-d_\Y/2}$ and $c_\sigma=\widetilde\C_2\sigma^{-d_\Y}$.
Then, using the definition~\eqref{eq: uniform smoothing gap} of $\Xi_\sigma$ followed by~\Cref{lem: smoothing gap} give
\begin{align*}
\DDR\left(\nu,\Th^m_{\Gsi}\right)&\leq\DGsil\left(\nu,\Th^m_{\Gsi}\right)+\Xi_\sigma\\
&\leq\|p\|_\infty\E^2_{\text{Bias}}+
\widetilde\C_1\E_{\text{Bias}}\R(m,t)\,\sigma^{-d_\Y/2}+\widetilde\C_2\R(m,t)^2\sigma^{-d_\Y}+\C_\Xi\sigma^{2\zeta}\\
&:=\E_\sigma(m,t).
\end{align*}
By~\Cref{lem: conversion gap} applied to $\T=\Th^m_{\Gsi}$, the previous inequality, and the monotonicity of $x\to x^{\cc_\beta}$ on $[0,\infty)$,
\begin{align*}
\DD\left(\nu,\Th^m_{\Gsi}\right)\leq 2\C_\beta\,\DDR\left(\nu,\Th^m_{\Gsi}\right)^{\cc_\beta}\leq 2\C_\beta\,\E_\sigma(m,t)^{\cc_\beta}.
\end{align*}
Distributing the sub-additive $\cc_\beta$ power among the four addends of $\E_{\sigma}(m,t)$ then yields the final bound of~\Cref{Fixed length-scale}. Next, we move to the two specific $\sigma$-selections outlined in~\Cref{Bias-agnostic length-scale} and~\Cref{Oracle length-scale}.

\emph{Bias-agnostic $\sigma$-selection of~\Cref{Bias-agnostic length-scale}.} Abbreviate $\sigma_2\coloneqq\sigma_2(m,t)$, $\R\coloneqq\R(m,t)$ and $\widetilde\R_2\coloneqq\widetilde\R_2(m,t)$. By~\eqref{eq: sigma 2 and B},
\begin{align*}
\C_\Xi\sigma_2^{2\zeta}=\C_\Xi^{1-\frac{2\zeta}{2\zeta+d_\Y}}\left(\widetilde\C_2\R^2\right)^{\frac{2\zeta}{2\zeta+d_\Y}}=\widetilde\R_2,\qquad \widetilde\C_2\R^2\sigma_2^{-d_\Y}=\C_\Xi^{\frac{d_\Y}{2\zeta+d_\Y}}\left(\widetilde\C_2\R^2\right)^{1-\frac{d_\Y}{2\zeta+d_\Y}}=\widetilde\R_2.
\end{align*}
Moreover, the mixed sampling term at $\sigma_2$ satisfies
\begin{align*}
\widetilde\C_1\E_{\text{Bias}}\R\,\sigma_2^{-d_\Y/2}=\frac{\widetilde\C_1}{\sqrt{\widetilde\C_2}}\,\E_{\text{Bias}}\sqrt{\widetilde\C_2\R^2\sigma_2^{-d_\Y}}=2\sqrt{\|p\|_\infty}\,\E_{\text{Bias}}\sqrt{\widetilde\R_2}.
\end{align*}
Hence, using $a^2+2ab+2b^2\leq(a+\sqrt2 b)^2$ for $a,b\geq0$,
\begin{align*}
\E_{\sigma_2}(m,t)=\|p\|_\infty\E^2_{\text{Bias}}+2\sqrt{\|p\|_\infty}\,\E_{\text{Bias}}\sqrt{\widetilde\R_2}+2\widetilde\R_2\leq\left(\sqrt{\|p\|_\infty}\,\E_{\text{Bias}}+\sqrt{2\widetilde\R_2}\right)^2.
\end{align*}
Since $\sigma_2$ is deterministic, the first bound of~\Cref{Fixed length-scale} with $\sigma=\sigma_2$ proves~\Cref{Bias-agnostic length-scale}.

\emph{Oracle $\sigma$-selection of~\Cref{Oracle length-scale}.} Let $E(\sigma)\coloneqq\C_\Xi\sigma^{2\zeta}+A\sigma^{-d_\Y/2}+B\sigma^{-d_\Y}$ be the $\sigma$-dependent part of $\E_\sigma(m,t)$, with $A\coloneqq\widetilde\C_1\E_{\text{Bias}}\R(m,t)$ and $B\coloneqq\widetilde\C_2\R(m,t)^2$. If $\E_{\text{Bias}}>0$, then $\sigma_1\coloneqq\sigma_1(m,t)>0$ solves $\C_\Xi\sigma^{2\zeta}=A\sigma^{-d_\Y/2}$, and by~\eqref{eq: sigma 1 and A},
\begin{align*}
\C_\Xi\sigma_1^{2\zeta}=A\sigma_1^{-d_\Y/2}=\C_\Xi^{1-\frac{4\zeta}{4\zeta+d_\Y}}A^{\frac{4\zeta}{4\zeta+d_\Y}}=\widetilde\R_1(m,t).
\end{align*}
If $\E_{\text{Bias}}=0$, then $A=0$ and $\sigma_1=\widetilde\R_1(m,t)=0$. Likewise, we have previously shown that, by~\eqref{eq: sigma 2 and B}, $\C_\Xi\sigma_2^{2\zeta}=B\sigma_2^{-d_\Y}=\widetilde\R_2(m,t)$. Since $\sigma_{\textnormal{opt}}\geq\sigma_1,\sigma_2$ and $\sigma\to A\sigma^{-d_\Y/2}$ and $\sigma\to B\sigma^{-d_\Y}$ are nonincreasing while $\sigma\to\C_\Xi\sigma^{2\zeta}$ is increasing,
\begin{align*}
A\sigma_{\textnormal{opt}}^{-d_\Y/2}\leq\widetilde\R_1(m,t),\qquad B\sigma_{\textnormal{opt}}^{-d_\Y}\leq\widetilde\R_2(m,t),\qquad \C_\Xi\sigma_{\textnormal{opt}}^{2\zeta}=\max\left\{\widetilde\R_1(m,t),\widetilde\R_2(m,t)\right\}.
\end{align*}
Hence $E(\sigma_{\textnormal{opt}})\leq 2(\widetilde\R_1(m,t)+\widetilde\R_2(m,t))$ and $\E_{\sigma_{\textnormal{opt}}}(m,t)\leq\|p\|_\infty\E^2_{\text{Bias}}+2\widetilde\R_1(m,t)+2\widetilde\R_2(m,t)$. Applying~\Cref{Fixed length-scale} with the deterministic $\sigma=\sigma_{\textnormal{opt}}$, distributing the sub-additive $\cc_\beta$ power among the summands of $\E_{\sigma}(m,t)$, and bounding $2^{\cc_\beta}\leq2$, then yields
\begin{align*}
\DD\left(\nu,\Th^m_{G_{\sigma_{\textnormal{opt}}}}\right)&\leq 2\C_\beta\left(\left(\|p\|_\infty\E^2_{\text{Bias}}\right)^{\cc_\beta}+2^{\cc_\beta}\widetilde\R_1(m,t)^{\cc_\beta}+2^{\cc_\beta}\widetilde\R_2(m,t)^{\cc_\beta}\right)\\
&\leq 4\C_\beta\left(\left(\|p\|_\infty\E^2_{\text{Bias}}\right)^{\cc_\beta}+\widetilde\R_1(m,t)^{\cc_\beta}+\widetilde\R_2(m,t)^{\cc_\beta}\right).
\end{align*}
concluding the proof.
\end{proof}

\end{document}